\documentclass{article}
\usepackage{ijcai25}

\usepackage{times}
\usepackage{soul}
\usepackage{url}
\usepackage[hidelinks]{hyperref}
\usepackage[utf8]{inputenc}
\usepackage[small]{caption}
\usepackage{graphicx}
\usepackage{amsmath}
\usepackage{amsthm}
\usepackage{booktabs}
\usepackage{algorithm}
\usepackage{algorithmic}
\usepackage[switch]{lineno}

\usepackage{cleveref}
\usepackage{multirow}
\usepackage{amsmath}
\usepackage{amsfonts}
\usepackage{mathtools}
\usepackage[table,xcdraw]{xcolor}
\usepackage{hhline}
\usepackage{bigstrut}
\usepackage{makecell}
\usepackage{bm}
\usepackage{booktabs}
\usepackage{enumitem}
\usepackage{subfigure}
\usepackage{graphicx}
\usepackage{boxedminipage}

\setlist[itemize]{leftmargin=*}

\title{Recalling The Forgotten Class Memberships:\\
Unlearned Models Can Be Noisy Labelers to Leak Privacy}

\author{
	Zhihao Sui$^1$\and
	Liang Hu$^2$\thanks{Corresponding author}\and
    Jian Cao$^{1*}$\and
    Dora D. Liu$^2$\\
	Usman Naseem$^{3}$\and
    Zhongyuan Lai$^{4}$\and
	Qi Zhang$^2$\\
	\affiliations
	$^1$Shanghai Jiao Tong University\\
	$^2$Tongji University\\
	$^3$Macquarie University\\
	$^4$Shanghai Ballsnow Intelligent Technology Co. Ltd\\
	\emails
	fancyboy@sjtu.edu.cn,
	lianghu@tongji.edu.cn,
    cao-jian@sjtu.edu.cn,
    liudongmei\_0506@163.com\\
    usman.naseem@mq.edu.au,
	abrikosoff@yahoo.com,
    zhangqi\_cs@tongji.edu.cn
}

\begin{document}

\maketitle


\Crefname{equation}{Eq}{Eqs}

\begin{abstract}
Machine Unlearning (MU) technology facilitates the removal of the influence of specific data instances from trained models on request. Despite rapid advancements in MU technology, its vulnerabilities are still underexplored, posing potential risks of privacy breaches through leaks of ostensibly unlearned information. Current limited research on MU attacks requires access to original models containing privacy data, which violates the critical privacy-preserving objective of MU. To address this gap, we initiate the innovative study on recalling the forgotten class memberships from unlearned models (ULMs) without requiring access to the original one. Specifically, we implement a Membership Recall Attack (MRA) framework with a teacher-student knowledge distillation architecture, where ULMs serve as noisy labelers to transfer knowledge to student models. Then, it is translated into a Learning with Noisy Labels (LNL) problem for inferring correct labels of the forgetting instances. Extensive experiments on state-of-the-art MU methods with multiple real datasets demonstrate that the proposed MRA strategy exhibits high efficacy in recovering class memberships of unlearned instances. As a result, our study and evaluation have established a benchmark for future research on MU vulnerabilities.
\end{abstract}

\section{Introduction}

Data privacy is a core concept in the era of big data and extensive interconnectivity~\cite{Wu0000024,abs-2312-13508}. If a machine learning model has been trained on sensitive and private data, it can lead to significant security risks.
In this context, the emergence of machine unlearning (MU) has been driven by stringent data privacy regulations such as GDPR ~\cite{Hoofnagle2019TheEU} and CCPA ~\cite{Itakura2018TheSA}, which require the removal of specific sensitive data upon request. MU is designed to forget particular data points from the learned models ~\cite{Cao2015TowardsMS}. As concerns about the increasing data misuse and privacy breaches, MU has gained more attention as a critical component in building safe machine learning systems. 

\begin{figure}
    \centering
    \includegraphics[width=0.9\linewidth]{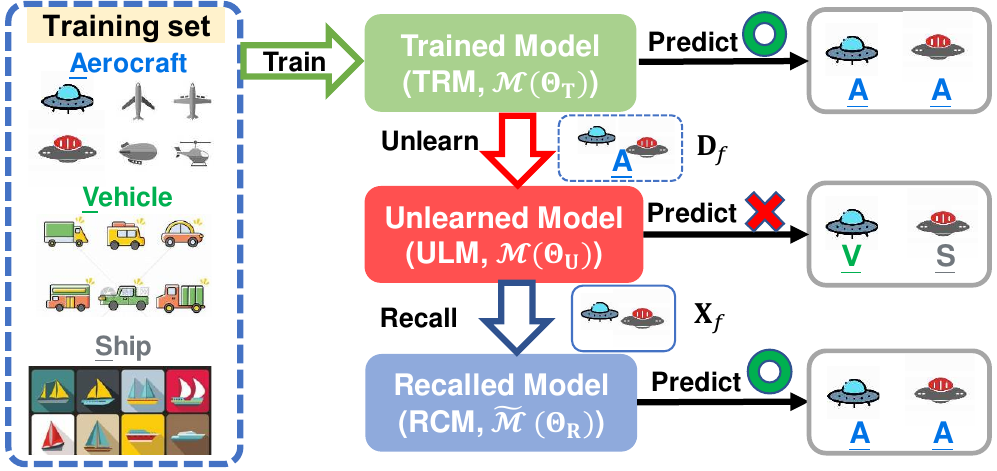}
    \vspace{-2mm}
    \caption{The demonstration of recall attack on the unlearned model (ULM) to recover class memberships having been unlearned via MU. This study is critical to the vulnerability of current MU models.}
    \label{fig:research-problem}
    \vspace{-2mm}
\end{figure}

Rapid advancements in MU research have increasingly posed a potential risk of privacy breaches by recovering the unlearned information about private data, highlighting the limited research on the full scope of MU vulnerabilities.
In fact, the most relevant research \cite{hu2024learn} that investigates inversion attacks against MU models was published recently.
However, this work is based on an impractical assumption that unlearning inversion attacks require access to both \textbf{Trained Model (TRM, $\mathcal{M}(\Theta_T)$)}, and \textbf{Unlearned Model (ULM, $\mathcal{M}(\Theta_U)$)}, as shown in \Cref{fig:research-problem}. In general, only ULM is accessible to users, where sensitive privacy has been removed from TRM.


The membership inference attack (MIA) ~\cite{shokri2017membership} is originally used to detect data samples used to train a machine learning model. Recently, MIA has been used to assess whether the influence of a forgetting dataset $\mathcal{D}_f=\{\mathcal{X}_f,\mathcal{Y}_f\}$ has been successfully erased after MU \cite{chen2021when}.
To go one step further, we formulate the attack on recovering the forgotten class memberships by only accessing ULM. As shown in \Cref{fig:research-problem}, our objective is to learn a \textbf{Recalled Model (RCM, $\Tilde{\mathcal{M}}(\Theta_R)$)} from the ULM $\mathcal{M}(\Theta_U)$ to correctly infer the labels of the input data $\mathcal{X}_p$, where $\mathcal{X}_p$ contains a subset of instances in $\mathcal{X}_f$ in which the class memberships, $\mathcal{Y}_f$, have been forgotten in the ULM. 
As a result, we propose a Membership Recall Attack (MRA) framework to learn RCM in this paper. As most MU models restrict the scope of the investigation to the area of image classification tasks \cite{bourtoule2021machine,fan2024salun,Jia2023ModelSC,chen2024bu,ijcai2024p40}, we correspondingly study the proposed MRA on these MU models in this area.

To implement the MRA framework, we designed a teacher-student architecture to distill the knowledge from the ULM $\mathcal{M}(\Theta_{U})$ (as a teacher) to a \textbf{Student Model (STM, $\Tilde{\mathcal{M}}({\Theta_S})$)}.
More specifically, given an input image set $\mathcal{X}_p$, the ULM $\mathcal{M}(\Theta_{U})$ outputs the prediction labels $\mathcal{M}(\mathcal{X}_p;\Theta_U)\mapsto\mathcal{Y}_p$, where the prediction labels $\mathcal{Y}_p$ may be noisy, especially when $\mathcal{X}_p$ contains many instances of forgetting data $\mathcal{X}_f$. 
Therefore, we use the ULM $\mathcal{M}(\Theta_{U})$ to serve as a \textbf{noisy labeling teacher} for knowledge distillation, that is, using $\{\mathcal{X}_p,\mathcal{Y}_p\}$ to train the STM $\Tilde{\mathcal{M}}({\Theta_S})$.
In this context, MRA can be further translated into a Learning with Noisy Labels (LNL) problem \cite{ALGAN2021106771} over $\{\mathcal{X}_p,\mathcal{Y}_p\}$ that aims to infer correct labels from the noisy ones. 
Consequently, we selected samples with high confidence agreement between the teacher model $\mathcal{M}(\Theta_{U})$ and the STM $\Tilde{\mathcal{M}}({\Theta_S})$ for LNL.
In particular, we discuss two cases for MRA, one is the closed-source case where the parameters of ULM $\Theta_{U}$ are not accessible, and the other is the open-source case where the parameters are open for use. Moreover, we design a unified learning scheme of MRA to train the RCMs for these two cases. We summarize our contributions as follows.
\begin{itemize}
    \item To our knowledge, this is the \emph{first attempt} study of the recall attack of class membership, which can effectively assess the risk of data privacy breaches and promote the robustness of the MU study.
    \item We propose MRA, a model-agnostic attack framework, to effectively recover class memberships of forgotten instances from unlearned ULMs via various MU methods.
    \item We implement MRA with a teacher-student architecture where the ULM serves as a noisy labeling teacher to distill the knowledge to train the STM with noisy labels.
    \item We conducted extensive experiments in four widely used datasets in MU research, demonstrating both the theoretical and practical efficacy of our MRA approach against various SOTA MU methods. 
\end{itemize}

\section{Related Work}
\subsection{Machine Unlearning}
\label{bg:ul}

\textbf{Exact Unlearning.}
Retraining the model from scratch after removing specific data can intuitively and effectively achieve exact unlearning. In addition, \cite{bourtoule2021machine} proposed SISA (Sharded, Isolated, Sliced, Aggregated) training, which trains isolated models on data shards for efficient unlearning by retraining only affected shards. Although effective, these unlearning approaches are computationally expensive and impractical for large-scale models and datasets.

\noindent \textbf{Approximate Unlearning.}
The idea of modestly sacrificing the accuracy of forgetting in exchange for significant improvements in unlearning efficiency has spurred the exploration of approximate unlearning techniques. 
Gradient ascent (\textbf{GA})~\cite{graves2021amnesiac,golatkar2020eternal,thudi2022unrolling,abs-2407-19183} reverses the training of the model by adding gradients, thus moving the model towards greater loss for the data points targeted for removal. 
Random labeling (\textbf{RL}) \cite{golatkar2020eternal} that involves finetuning the original model on the forgetting dataset using random labels to enforce unlearning.
Several methods estimate the impact of forgetting samples on the model parameters and conduct forgetting through the fisher information matrix {(\textbf{FF})} ~\cite{becker2022evaluating} or influence function {(\textbf{IU})} \cite{koh2017understanding,izzo2021approximate}. 
$\ell1$-sparse (\textbf{L1-SP}) \cite{Jia2023ModelSC} infuses weight sparsity into unlearning. 

Moreover, most MU methods may degrade model performance. and lead to ``\textit{over-unlearning}''. Some recent work has explored more precise unlearning on target forget instances. Boundary unlearning (\textbf{BU}) \cite{chen2024bu} shifts the decision boundary of the original model to imitate the decision behavior of the model retrained from scratch.
\textbf{SalUn} \cite{fan2024salun} introduces the concept of `weight saliency' to narrow the performance gap with exact unlearning.
To avoid over-unlearning, \textbf{UNSC} \cite{ijcai2024p40}
constrains the unlearning process within a null space tailored to the remaining samples to ensure that unlearning does not negatively impact the model performance.

\subsection{Attacks on Machine Unlearning}

Despite advances in MU techniques, the study of their vulnerabilities remains underexplored. To date, very limited MU attack methods have been proposed to affect efficiency ~\cite{marchant2022hard} or fidelity ~\cite{di2022hidden,hu2023duty}. Studying attacks on MU is crucial to developing robust and secure MU methods.

\noindent \textbf{Membership Inference Attack (MIA).} MIA is originally used to infer if data samples are used to train a machine learning model \cite{shokri2017membership}. With the development of MU, MIA has been widely used to check if the influence of forgetting data had been removed from the original model. However, \cite{chen2021when} show that MU can jeopardize privacy in terms of MIA. The goals of MIA and the proposed MRA are different. MIA aims to detect data samples if used for training, whereas MRA aims at recalling and inferring the class memberships of forgetting samples from ULMs.

\noindent \textbf{Model Inversion Attack.} It aims to reconstruct the original input data from the model outputs. \cite{fredrikson2015model} introduced model inversion attacks using the confidence scores output by a model to reconstruct input images.
\cite{hu2024learn} proposed the first inversion attack against unlearning. It extracts features and labels of forgetting samples, which most closely match the objectives of our study. Although the attack demonstrates notable effectiveness, it requires \emph{access to the original TRM} before unlearning, which is impractical in real scenarios. 
In contrast, the proposed MRA only needs to access ULMs and supports more versatile MU methods. To our knowledge, we are the first to explore the attack \textbf{only using ULMs} to recall the class memberships of forgetting samples, without comparable prior work.

\section{Preliminaries}
We first introduce the datasets and models used in our study, followed by a formal definition of the problem.

\subsection{Involved Datasets}
\noindent\textbf{Training dataset} $\mathcal{D}_{tr}: \{\mathcal{X}_{tr},\mathcal{Y}_{tr}\}$ is all data used to initially train machine learning models, where $\mathcal{X}_{tr}$ denotes the image set and $\mathcal{Y}_{tr}$ denotes the corresponding label set.

\noindent\textbf{Forgetting dataset} $\mathcal{D}_{f}:\{ \mathcal{X}_{f},\mathcal{Y}_{f}\}$ is a subset of $\mathcal{D}_{tr}$, that is, $\mathcal{D}_{f}\subset\mathcal{D}_{tr}$. In MU, $\mathcal{D}_{f}$ is a set of sensitive data that should be unlearned from the trained model, that is, the ULM cannot tell the true labels when $\mathcal{X}_{f}$ is input.

\noindent\textbf{Remaining dataset} $\mathcal{D}_{r}: \{\mathcal{X}_{r},\mathcal{Y}_{r}\}$ is the remaining data of $\mathcal{D}_{tr}$, that is, $\mathcal{D}_{r}=\mathcal{D}_{tr}\setminus\mathcal{D}_{f}$, which should not be forgotten.

\noindent\textbf{Prediction dataset} $\mathcal{D}_{p}: \{\mathcal{X}_{p},\mathcal{Y}_{p}\}$ is the dataset for prediction, where $\mathcal{X}_{p}=\mathcal{X}_{ts}\cup\mathcal{X}_{u}$ can be decomposed into two parts. $\mathcal{X}_u\subseteq\mathcal{X}_{f}$ is the subset of the forgotten instances while $\mathcal{X}_{ts}$ is the unseen dataset for testing.

\subsection{Involved Models}
\noindent\textbf{Trained Model (TRM)} $\mathcal{M}(\Theta_T)$ is the model that has been trained on the training dataset $\mathcal{D}_{tr}$.

\noindent \textbf{Unlearned Model (ULM)} $\mathcal{M}(\Theta_U)$ is the model that has unlearned the forgetting dataset $\mathcal{D}_{f}$ based on $\mathcal{M}(\Theta_U)$. It will serve as a noisy labeling teacher to distill knowledge.

 \noindent\textbf{Student Model (STM)} $\Tilde{\mathcal{M}}(\Theta_S)$ is the model that receives the knowledge distilled from $\mathcal{M}(\Theta_U)$.
 
\noindent\textbf{Recalled Model (RCM)} $\Tilde{\mathcal{M}}(\Theta_R)$ is the model that has recalled forgotten class memberships based on $\mathcal{M}(\Theta_U)$.

\subsection{Problem Formulation}
MU models remove the influence of forgetting the dataset $\mathcal{D}_{f}$ from TRM $\mathcal{M}({\theta_T})$, and release a ULM $\mathcal{M}(\Theta_U)$ for public use. This paper aims to implement the MRA framework to recall forgotten class memberships given $\mathcal{M}(\Theta_U)$. 
In particular, we employ $\mathcal{M}(\Theta_U)$ as a noisy labeler (i.e., teacher model) to distill the knowledge inferred from the prediction dataset $\mathcal{X}_p$ to the STM $\Tilde{\mathcal{M}}(\Theta_S)$.

Moreover, we discuss two common cases that lead to the final RCM $\Tilde{\mathcal{M}}(\Theta_R)$. In the first case, ULM $\mathcal{M}(\Theta_U)$ is usable but not trainable, e.g. $\mathcal{M}(\Theta_U)$ is only accessible as a black-box service (\textbf{closed-source case}), and the STM $\Tilde{\mathcal{M}}(\Theta_S)$ will finally serve as RCM $\Tilde{\mathcal{M}}(\Theta_R)$.
In the second case, $\mathcal{M}(\Theta_U)$ is trainable, for example, $\mathcal{M}\textbf{}(\Theta_U)$ is released with its ULM parameters $\Theta_U$ (\textbf{open-source case}), and the ULM $\mathcal{M}(\Theta_U)$ is recovered to serve as the RCM $\Tilde{\mathcal{M}}(\Theta_R)$.

\section{Proposed Method}

\begin{figure}[t]
    \centering
    \includegraphics[width=0.95\linewidth]{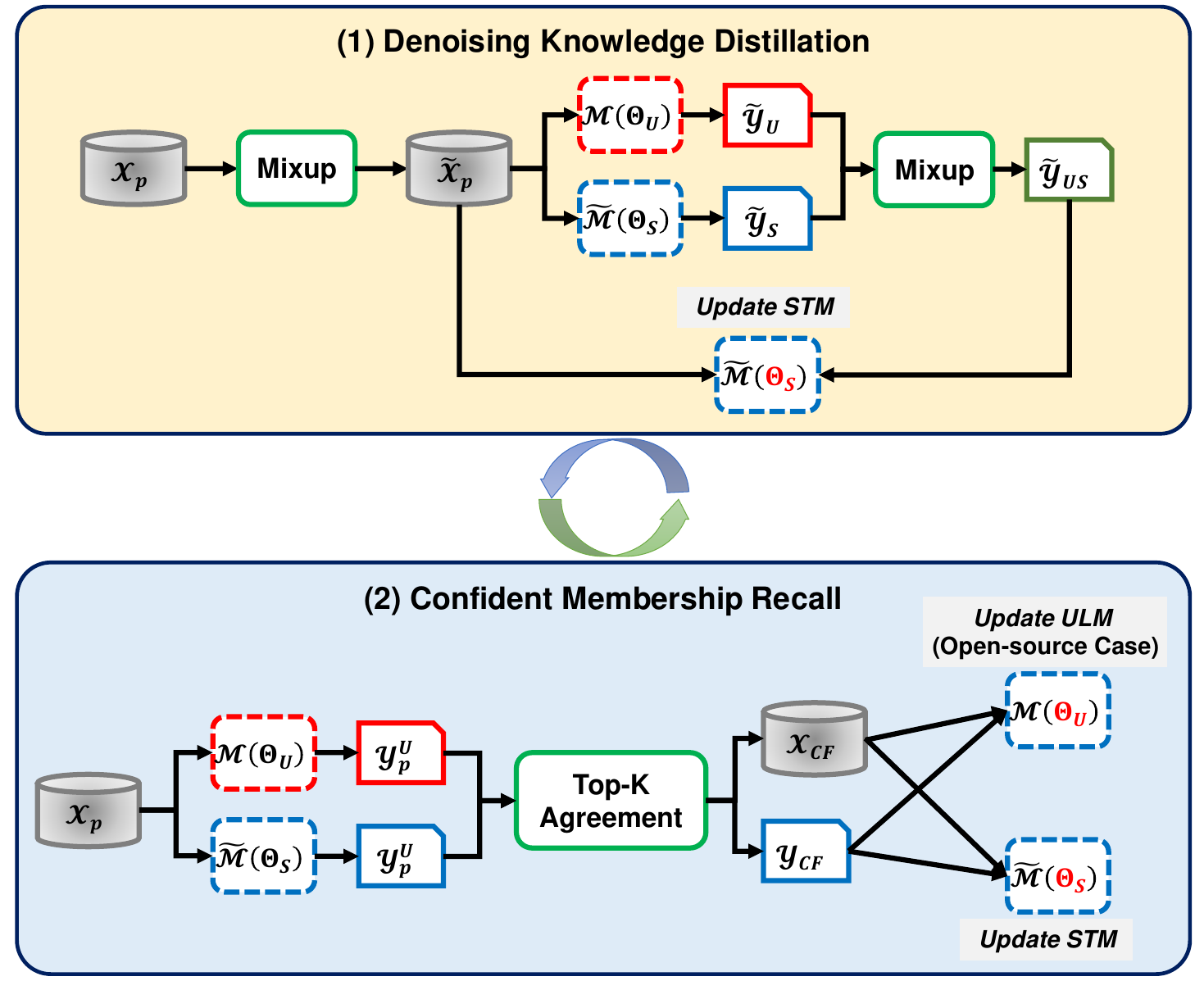}
    \caption{The workflow of proposed MRA framework. It consists of two alternative learning steps to obtain RCM: (1) Denosing Knowledge Distillation; (2) Confident Membership Recall.}
    \label{fig:framework}
\end{figure}

\subsection{Overview}
Firstly, we can easily obtain ULMs by applying various MU methods on a TRM. Given a ULM, \Cref{fig:framework} demonstrates the workflow to implement the proposed MRA framework, which consists of two alternative learning steps to obtain the RCM $\Tilde{\mathcal{M}}(\Theta_R)$.

\noindent\textbf{(1) Denosing Knowledge Distillation}: The ULM $\mathcal{M}(\Theta_U)$ serves as a noisy labeler on the prediction dataset $\mathcal{D}_{p}$ with some augmentation strategy, and the inferred pseudo labels with the augmented images are used to train STM $\Tilde{\mathcal{M}}(\Theta_S)$.

\noindent\textbf{(2) Confident Membership Recall}: Both ULM $\mathcal{M}(\Theta_U)$ and STM $\Tilde{\mathcal{M}}(\Theta_S)$ are to generate the prediction on $\mathcal{D}_{p}$, and the top-\textit{K} data samples with the highest probability of joint prediction, that is, the most confidently agreed pseudo labels for each class are selected to train STM $\Tilde{\mathcal{M}}(\Theta_S)$ and ULM $\mathcal{M}(\Theta_U)$ (optional for the open source case).

\subsection{Model Training and Unlearning}\label{sec:train_unlearn}
Given the training dataset $\mathcal{D}_{tr}=\mathcal{D}_f\cup\mathcal{D}_r$, we use $\mathcal{D}_{tr}$ to train the model $\mathcal{M}$, which results in the TRM $\mathcal{M}(\Theta_T)$ with the parameters $\Theta_T$. Then, we apply various SOTA MU methods on TRM $\mathcal{M}(\Theta_T)$, which leads to ULMs $\mathcal{M}(\Theta_U)$.







\subsection{Implementation of MRA Framework}\label{sec:mra_scheme}
We implement the MRA framework with the following two alternative learning steps to obtain RCM $\Tilde{\mathcal{M}}(\Theta_R)$.

\subsubsection{(1) Denoising Knowledge Distillation}
A sophisticated MU method should only unlearn the influence of forgetting dataset $\mathcal{D}_f$ but retain the classification capability that is learned from the remaining dataset $\mathcal{D}_r$. As a result, the ULM $\mathcal{M}(\Theta_U)$ can serve as a noisy labeler to distill knowledge to the STM $\Tilde{\mathcal{M}}(\Theta_S)$ given the set of prediction images $\mathcal{X}_{p}$.
$\mathcal{M}(\Theta_U)$ is prone to mislabeling on $\mathcal{X}_{p}$ if it contains forgotten instances in $\mathcal{D}_f$. 
To avoid the input of original images that have been forgotten, we create augmented images by mixup \cite{zhang2018mixup} which is an effective regularization technique to deal with label noise \cite{carratino2022mixup}. Given an image $x_1\in\mathcal{X}_{p}$, we randomly sample another image $x_2$ from $\mathcal{X}_{p}$, and mix them as follows:
\begin{equation}
    \Tilde{x}=\beta_x \cdot x_1 + (1-\beta_x) \cdot  x_2  
    \label{eq:mixup_img}
\end{equation}
where $\beta_x \sim Beta(\alpha_x, \alpha_x)$ ($\alpha_x=0.2$ in this paper). Then, we take $\Tilde{x}$ as input to retrieve the soft pseudo label from ULM $\mathcal{M}(\Theta_U)$ and STM $\Tilde{\mathcal{M}}(\Theta_S)$:
\begin{gather}
    \Tilde{\mathbf{y}}_U=\mathcal{M}(\Tilde{x};\Theta_U),\quad
    \Tilde{\mathbf{y}}_S=\Tilde{\mathcal{M}}(\Tilde{x};\Theta_S)
    \label{eq:y_S}
\end{gather}
Then, we can obtain the mixed soft pseudo label:
\begin{equation}
    \Tilde{\mathbf{y}}_{US}=\beta_y \cdot \Tilde{\mathbf{y}}_U + (1-\beta_y) \cdot \Tilde{\mathbf{y}}_S  
    \label{eq:mixup_label}
\end{equation}
where $\beta_y=1$ is applied in the first warmup epoch (i.e. completely accept the knowledge from the teacher model) and $\beta_y \sim Beta(\alpha_y, \alpha_y)$ ($\alpha_y=0.75$ in this paper) for label denoising after the warmup stage. 
Following \Cref{eq:mixup_img,eq:y_S,eq:mixup_label}, we can obtain $\Tilde{\mathcal{Y}}_S=\{\Tilde{\mathbf{y}}_S\}$ and $\Tilde{\mathcal{Y}}_{US}=\{\Tilde{\mathbf{y}}_{US}\}$ over the augmented set $\Tilde{x}\in\Tilde{\mathcal{X}}_{p}$.
As a result, the parameters of STM $\Tilde{\mathcal{M}}(\Theta_S)$ can be updated by decreasing the mini-batch gradient in terms of minimizing cross-entropy (CE) loss.
\begin{equation}
   \Theta_S = \arg\min_{\Theta_S}{\text{CE}(\Tilde{\mathcal{Y}}_S,\Tilde{\mathcal{Y}}_{US})}
    \label{eq:theta_US}
\end{equation}

\subsubsection{(2) Confident Membership Recall}
After the above step, STM $\Tilde{\mathcal{M}}(\Theta_S)$ has been trained on the augmented dataset based on $\mathcal{X}_{p}$ with pseudo labels from the noisy labeler $\mathcal{M}(\Theta_U)$. Inspired by the LNL methods \cite{ALGAN2021106771}, we design a balanced class membership recall strategy based on the highest confidence agreements between $\mathcal{M}(\Theta_U)$ and $\Tilde{\mathcal{M}}(\Theta_S)$. More specifically, we first input $\mathcal{X}_{p}$ into both ULM $\mathcal{M}(\Theta_U)$ and STM $\Tilde{\mathcal{M}}(\Theta_S)$ to predict soft labels (that is, probability over each class):
\begin{gather}
	\mathbf{Y}_{p}^U=\mathcal{M}(\mathcal{X}_{p};\Theta_U),\quad
    \mathbf{Y}_{p}^S=\Tilde{\mathcal{M}}(\mathcal{X}_{p};\Theta_S)
	\label{eq:Y_p}
\end{gather}
where $\mathbf{Y}_{p}^U, \mathbf{Y}_{p}^U \in \mathbb{R}^{N \times C}$, $N=|\mathcal{X}_{p}|$ denotes the number of samples in $\mathcal{X}_{p}$ and $C$ denotes the number of classes. For each instance $x_i\in\mathcal{X}_{p}$, we apply Laplace smoothing on its soft label $\mathbf{y}_i^U=\mathbf{Y}_{p}^U[i,:]$ to avoid zero probability:
\begin{equation}
	\Tilde{\mathbf{y}}_i^U = \frac{\mathbf{y}_i^U + \gamma_l\cdot\mathbf{1}}{1+C\cdot\gamma_l}
    \label{eq:lap_smooth}
\end{equation}
As a result, we obtain the smoothed probability matrices $\Tilde{\mathbf{Y}}_{p}^U$ over $\mathcal{X}_{p}$. $\Tilde{\mathbf{Y}}_{p}^S$ can be obtained in the same way. Then, we have the joint probability $\Tilde{\mathbf{y}}_i$ on $x_i$:
\begin{equation}
\Tilde{\mathbf{y}}_i=\Tilde{\mathbf{y}}_i^U\odot\Tilde{\mathbf{y}}_i^S \quad
	 \text{for}~\Tilde{\mathbf{y}}_i^U \in \Tilde{\mathbf{Y}}_{p}^U, \Tilde{\mathbf{y}}_i^S \in \Tilde{\mathbf{Y}}_{p}^S
     \label{eq:joint_probs}
\end{equation}
For all $x_i\in\mathcal{X}_{p}$, we have $\Tilde{\mathbf{Y}}_{p} = \Tilde{\mathbf{Y}}_{p}^U \odot \Tilde{\mathbf{Y}}_{p}^S$ in the matrix form, where $\odot$ is the element-wise product. 
Given a class $c$, we have the joint probabilities $\mathbf{y}_c=\Tilde{\mathbf{Y}}_{p}[:,c]$ of all instances. A higher joint probability $y_i \in \mathbf{y}_c$ implies greater confidence in the teacher and student models that the instance $x_i$ should have the membership of the class $c$. 
Consequently, we apply a balanced strategy to select top-$K$ instances with the maximum joint probability for each class.
\begin{gather}\label{eq:D_conf}
    \mathcal{D}_{CF}=\{\mathcal{X}_{CF}, \mathcal{Y}_{CF}\}=\{(x_i, \Tilde{\mathbf{y}}_c)|y_i \in \text{top-}K(\mathbf{y}_c)\\
    \text{for}~ c \in \{1,\cdots,C\}\nonumber
\end{gather}
where $K=\lceil\tau\cdot N/C\rceil$, and $\Tilde{\mathbf{y}}_c = (1-\gamma_s)\cdot\mathbf{y}_c + \frac{\gamma_s}{K}\cdot\mathbf{1}$ denotes the smoothing of the label versus the hard label $c$ ($\mathbf{y}_c$ stands for the one-hot encoding of $c$) which can effectively mitigate label noise \cite{lukasik2020does}. Then, the STM $\Tilde{\mathcal{M}}(\Theta_S)$ is updated on $\mathcal{D}_{CF}$ to learn the confident memberships.
\begin{gather}
	\mathcal{Y}_S=\Tilde{\mathcal{M}}(\mathcal{X}_{CF};\Theta_S)\\
	\Theta_S = \arg\min_{\Theta_S}{\text{CE}(\mathcal{Y}_S, \mathcal{Y}_{CF})}
	\label{eq:theta_S}
\end{gather}
In the \textbf{open-source case}, the parameters of ULM $\mathcal{M}(\Theta_U)$ are also accessible, so we will construct $\mathcal{D}_{CF}$ to refine $\Theta_U$ by \Cref{eq:Y_p,eq:lap_smooth,eq:joint_probs,eq:D_conf} with the above updated STM $\Tilde{\mathcal{M}}(\Theta_S)$.
\begin{gather}
	\mathcal{Y}_U=\mathcal{M}(\mathcal{X}_{CF};\Theta_U)\\
	\Theta_U = \arg\min_{\Theta_U}{\text{CE}(\mathcal{Y}_U, \mathcal{Y}_{CF})}
	\label{eq:theta_U}
\end{gather}
As a result, it leads to an alternative improvement co-training process between the teacher $\mathcal{M}(\Theta_U)$ and the student $\Tilde{\mathcal{M}}(\Theta_S)$, which can more effectively recall class memberships thanks to the knowledge retained by $\mathcal{M}(\Theta_U)$.

\begin{algorithm}[t]
\caption{The Scheme of MRA}
\label{alg:mra}
\textbf{Input}: Prediction Set: $\mathcal{X}_p$, Number of Epochs: $K_E,K_D,K_R$
\textbf{Output}: Recalled Labels: $\hat{\mathcal{Y}}_p$
\begin{algorithmic}[1]
\FOR{e in \{1, $\cdots$, $K_E$\}}
    \STATE \textit{(1) Denoising Knowledge Distillation}
    \FOR{i in \{1, $\cdots$, $K_D$\}}
        \STATE $\Tilde{\mathcal{X}}_p,\Tilde{\mathcal{Y}}_{US}$ \COMMENT{Mixup augmentation by \Cref{eq:mixup_img,eq:y_S,eq:mixup_label}}
        \STATE $\Theta_S = \arg\min_{\Theta_S}{\text{CE}(\Tilde{\mathcal{Y}}_S,\Tilde{\mathcal{Y}}_{US})}$ \COMMENT{Update STM}
    \ENDFOR
    \STATE \textit{(2) Confident Membership Recall}
    \FOR{i in \{1, $\cdots$, $K_R$\}}
        \STATE $\mathcal{D}_{CF}$ \COMMENT{Confident agreements by \Cref{eq:Y_p,eq:lap_smooth,eq:joint_probs,eq:D_conf}}
        \STATE $\Theta_S = \arg\min_{\Theta_S}{\text{CE}(\mathcal{Y}_S, \mathcal{Y}_{CF})}$ \COMMENT{Update STM}
        \IF{$\mathcal{M}(\Theta_U)$ is trainable (\textbf{open-source case})}
            \STATE $\mathcal{D}_{CF}$  \COMMENT{Confident agreements by \Cref{eq:Y_p,eq:lap_smooth,eq:joint_probs,eq:D_conf}}
             \STATE $\Theta_U = \arg\min_{\Theta_U}{\text{CE}(\mathcal{Y}_U,\mathcal{Y}_{CF})}$\COMMENT{Update ULM}
        \ENDIF
    \ENDFOR
\ENDFOR
\RETURN $\hat{\mathcal{Y}}_p=\Tilde{\mathcal{M}}(\mathcal{X}_{p};\Theta_S)$  \COMMENT{Closed-source case}
\STATE \qquad$~~~~~\hat{\mathcal{Y}}_p=\mathcal{M}(\mathcal{X}_{p};\Theta_U)$  \COMMENT{Open-source case}
\end{algorithmic}
\end{algorithm}

In \Cref{alg:mra}, we concisely summarize the above MRA scheme. After MRA, the updated STM $\Tilde{\mathcal{M}}(\Theta_S)$ (in closed-source case) or ULM $\mathcal{M}(\Theta_U)$ (in open-source case) will serve as the RCM to predict the labels $\hat{\mathcal{Y}}_p$ of $\mathcal{X}_p$.

\begin{table}[b]
  \centering
  \scalebox{0.85}{
    \begin{tabular}{l|c|ccc}
        \hline
        \multicolumn{1}{c|}{\textbf{Dataset}} &{\# Classes} & $\mathcal{D}_{tr}$ & $\mathcal{D}_{ts}$ & $\mathcal{D}_f$ \\
        \hline
        CIFAR-10 & 10 & 50,000 & 10,000 & 2500$\times$5 \\
        CIFAR-100 & 100 & 50,000 & 10,000 & 250$\times$5 \\
        Pet-37 & 37 & 3,680 & 3,669 & 50$\times$5 \\
        Flower-102 & 102 & 3,074 & 1,020 & 224 \\
        \hline
    \end{tabular}%
    }
    \caption{Statistic summary of the datasets and their splits} 
  \label{tab:dataset_details}%
\end{table}%

\section{Experiments}
\subsection{Experiment Setup}

\begin{table*}[tbh!]
  \centering
  \scalebox{0.75}{
    \begin{tabular}{c|c|c|c|c|c|c|c|c|c|c|c}
    \toprule
          &       & \textbf{TRM}   &       & \textbf{FF} & \textbf{RL}    & \textbf{GA}    & \textbf{IU}    & \textbf{BU}    & \textbf{L1-SP} & \textbf{SalUn} & \textbf{UNSC} \\
    \midrule
    \midrule
    \multicolumn{1}{c|}{\multirow{6}[12]{*}{\shortstack{CIFAR-10\\T: EFN\\S: EFN}}} & \multirow{3}[6]{*}{$\mathcal{D}_{ts}$} & \multirow{3}[6]{*}{0.833} & ULM   & 0.164  & 0.388  & 0.505  & 0.105  & 0.462  & 0.451  & 0.486  & 0.193  \\
\cmidrule{4-12}          &       &       & RCM   & 0.252  & 0.479  & 0.528  & 0.115  & 0.651  & 0.456  & 0.697  & 0.568  \\
\cmidrule{4-12}          &       &       & $\Delta Acc$ & \textbf{0.088} & \textbf{0.091 } & \textbf{0.023 } & \textbf{\textcolor{blue}{0.010} } & \textbf{0.188 } & \textbf{\textcolor{blue}{0.005} } & \textbf{\textcolor{red}{0.211} } & \textbf{\textcolor{red}{0.375} } \\
\cmidrule{2-12}          & \multirow{3}[6]{*}{$\mathcal{D}_{f}$} & \multirow{3}[6]{*}{1.000 } & ULM   & 0.142  & 0.153  & 0.267  & 0.090  & 0.082  & 0.154  & 0.117  & 0.031  \\
\cmidrule{4-12}          &       &       & RCM   & 0.195  & 0.328  & 0.358  & 0.096  & 0.507  & 0.231  & 0.615  & 0.513  \\
\cmidrule{4-12}          &       &       & $\Delta Acc$ & \textbf{0.053 } & \textbf{0.175 } & \textbf{0.091 } & \textbf{\textcolor{blue}{0.005} } & \textbf{0.424 } & \textbf{\textcolor{blue}{0.077} } & \textbf{\textcolor{red}{0.498} } & \textbf{\textcolor{red}{0.482} } \\
    \midrule
    \midrule
    \multicolumn{1}{c|}{\multirow{6}[12]{*}{\shortstack{CIFAR-100\\T: EFN\\S: EFN}}} & \multirow{3}[6]{*}{$\mathcal{D}_{ts}$} & \multirow{3}[6]{*}{0.643 } & ULM   & 0.159  & 0.593  & 0.531  & 0.140  & 0.529  & 0.537  & 0.410  & 0.275  \\
\cmidrule{4-12}          &       &       & RCM   & 0.295  & 0.607  & 0.551  & 0.297  & 0.566  & 0.550  & 0.522  & 0.480  \\
\cmidrule{4-12}          &       &       & $\Delta Acc$ & \textbf{0.137 } & \textbf{\textcolor{blue}{0.014} } & \textbf{0.020 } & \textbf{\textcolor{red}{0.157} } & \textbf{0.037 } & \textbf{\textcolor{blue}{0.013} } & \textbf{0.112 } & \textbf{\textcolor{red}{0.206 }} \\
\cmidrule{2-12}          & \multirow{3}[6]{*}{$\mathcal{D}_{f}$} & \multirow{3}[6]{*}{1.000 } & ULM   & 0.094  & 0.086  & 0.201  & 0.201  & 0.254  & 0.199  & 0.227  & 0.209  \\
\cmidrule{4-12}          &       &       & RCM   & 0.214  & 0.537  & 0.270  & 0.256  & 0.423  & 0.258  & 0.358  & 0.372  \\
\cmidrule{4-12}          &       &       & $\Delta Acc$ & \textbf{0.119 } & \textbf{\textcolor{red}{0.451} } & \textbf{0.070 } & \textbf{\textcolor{blue}{0.055} } & \textbf{0.169 } & \textbf{\textcolor{blue}{0.058} } & \textbf{0.130 } & \textbf{\textcolor{red}{0.163 }} \\
    \midrule
    \midrule
    \multicolumn{1}{c|}{\multirow{6}[12]{*}{\shortstack{Pet-37\\T: ResNet\\S: ResNet}}} & \multirow{3}[6]{*}{$\mathcal{D}_{ts}$} & \multirow{3}[6]{*}{0.895 } & ULM   & 0.257  & 0.754  & 0.740  & 0.622  & 0.646  & 0.740  & 0.706  & 0.769  \\
\cmidrule{4-12}          &       &       & RCM   & 0.486  & 0.816  & 0.757  & 0.682  & 0.785  & 0.758  & 0.778  & 0.799  \\
\cmidrule{4-12}          &       &       & $\Delta Acc$ & \textbf{\textcolor{red}{0.229} } & \textbf{0.062 } & \textbf{\textcolor{blue}{0.018} } & \textbf{0.060 } & \textbf{\textcolor{red}{0.140} } & \textbf{\textcolor{blue}{0.018} } & \textbf{0.072 } & \textbf{0.030 } \\
\cmidrule{2-12}          & \multirow{3}[6]{*}{$\mathcal{D}_{f}$} & \multirow{3}[6]{*}{1.000 } & ULM   & 0.224  & 0.332  & 0.224  & 0.200  & 0.084  & 0.196  & 0.128  & 0.304  \\
\cmidrule{4-12}          &       &       & RCM   & 0.388  & 0.884  & 0.520  & 0.448  & 0.856  & 0.440  & 0.712  & 0.660  \\
\cmidrule{4-12}          &       &       & $\Delta Acc$ & \textbf{\textcolor{blue}{0.164} } & \textbf{0.552 } & \textbf{0.296 } & \textbf{0.248 } & \textbf{\textcolor{red}{0.772} } & \textbf{\textcolor{blue}{0.244} } & \textbf{\textcolor{red}{0.584} } & \textbf{0.356 } \\
    \midrule
    \midrule
    \multicolumn{1}{c|}{\multirow{6}[11]{*}{\shortstack{Flower-102\\T: Swin-T\\S: ResNet}}} & \multirow{3}[6]{*}{$\mathcal{D}_{ts}$} & \multirow{3}[6]{*}{0.939 } & ULM   & 0.294  & 0.756  & 0.314  & 0.512  & 0.706  & 0.530  & 0.496  & NA \\
\cmidrule{4-12}          &       &       & RCM   & 0.376  & 0.831  & 0.510  & 0.589  & 0.758  & 0.655  & 0.599  & NA \\
\cmidrule{4-12}          &       &       & $\Delta Acc$ & \textbf{0.082 } & \textbf{\textcolor{blue}{0.075} } & \textbf{\textcolor{red}{0.196} } & \textbf{0.077 } & \textbf{\textcolor{blue}{0.052} } & \textbf{\textcolor{red}{0.125} } & \textbf{0.103 } & NA \\
\cmidrule{2-12}          & \multirow{3}[5]{*}{$\mathcal{D}_{f}$} & \multirow{3}[5]{*}{1.000 } & ULM   & 0.235  & 0.150  & 0.239  & 0.291  & 0.340  & 0.324  & 0.267  & NA \\
\cmidrule{4-12}          &       &       & RCM   & 0.312  & 0.567  & 0.352  & 0.364  & 0.478  & 0.538  & 0.429  & NA \\
\cmidrule{4-12}          &       &       & $\Delta Acc$ & \textbf{\textcolor{blue}{0.077} } & \textbf{\textcolor{red}{0.417} } & \textbf{0.113 } & \textbf{\textcolor{blue}{0.073} } & \textbf{0.138 } & \textbf{0.215 } & \textbf{0.162 } & NA \\
\bottomrule
    \end{tabular}%
    }
\captionof{table}{Performance comparison of MRA (closed-source case) on various SOTA MU methods, where ULM indicates the \textit{Acc} after MU while RCM indicates the \textit{Acc} after MRA, and $\Delta Acc$ shows the improvement (the Top-2 $\Delta Acc$ are marked in \textcolor{red}{red} while the Lowest-2 are marked in \textcolor{blue}{blue}). TRM illustrates the \textit{Acc} of original model.}
  \label{tab:main_close_source}%
\end{table*}

\begin{figure*}
    \begin{minipage}[c]{\linewidth}
    	\includegraphics[width=0.163\linewidth]{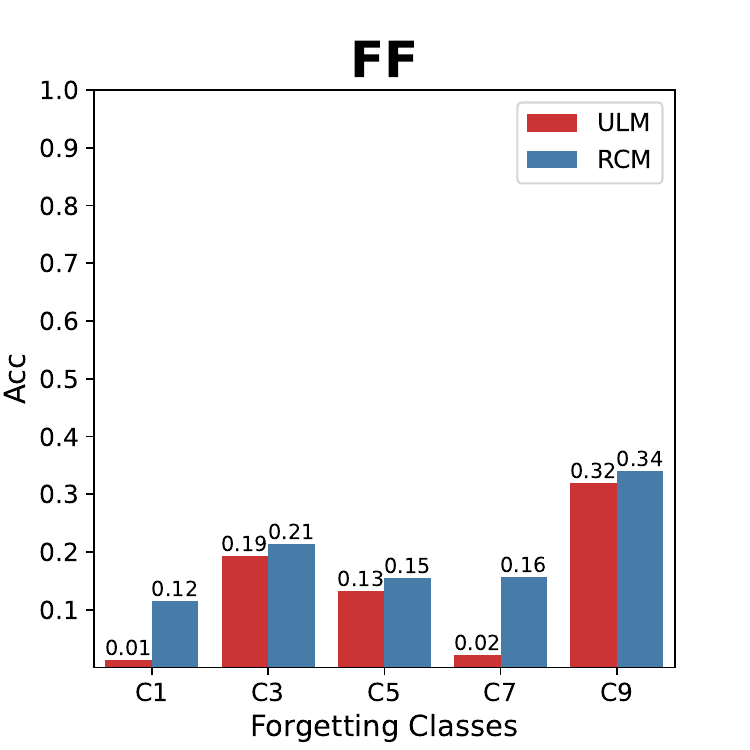}
        \includegraphics[width=0.163\linewidth]{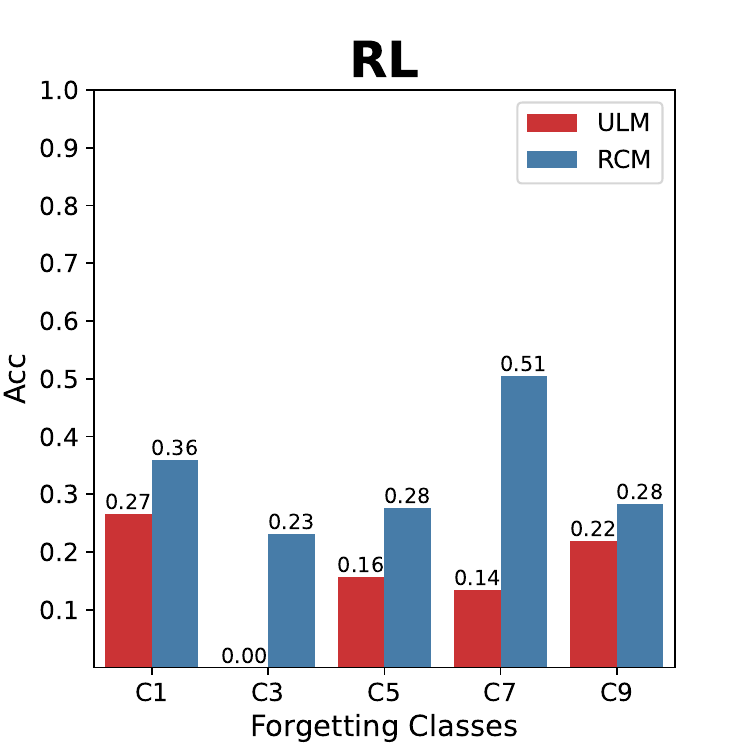}
        \includegraphics[width=0.163\linewidth]{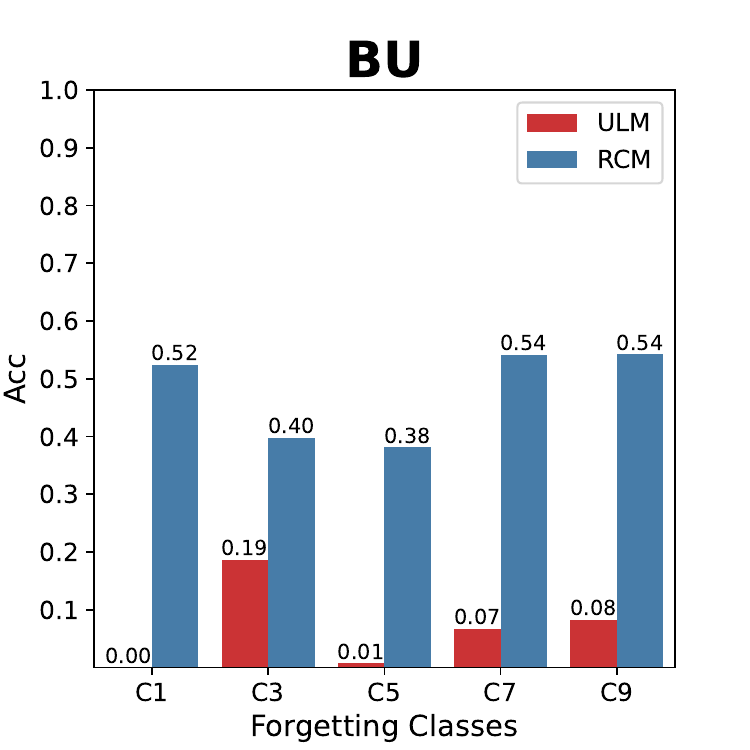}
        \includegraphics[width=0.163\linewidth]{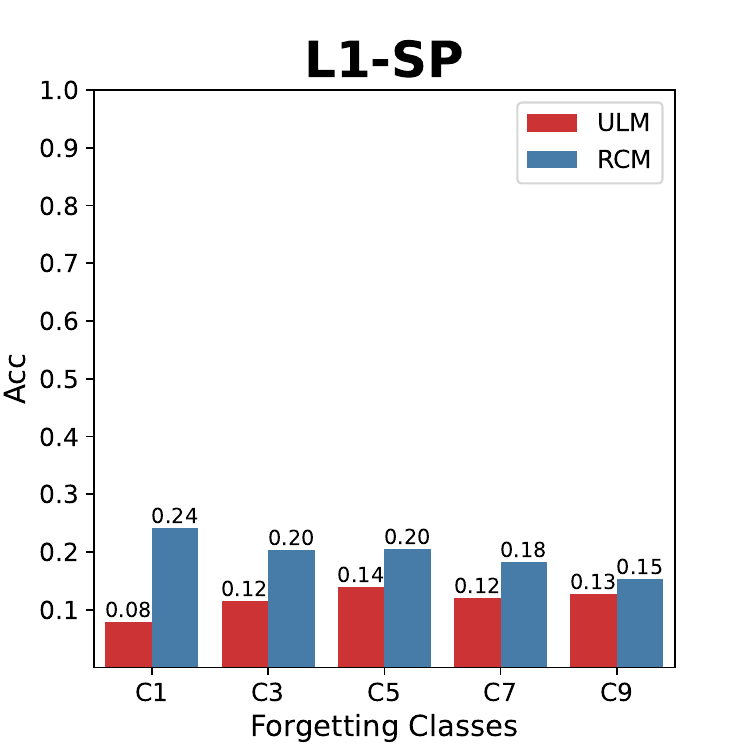}
         \includegraphics[width=0.163\linewidth]{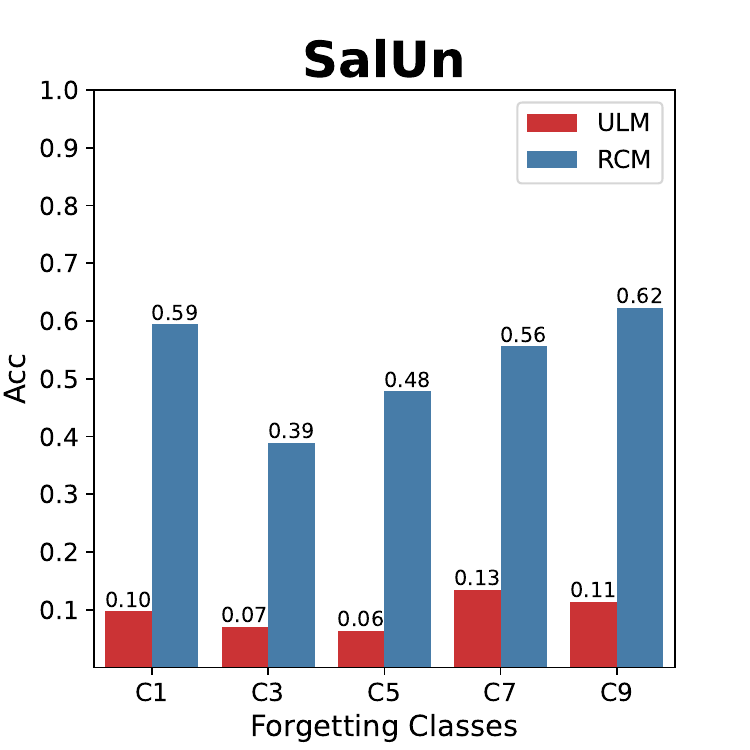}
         \includegraphics[width=0.163\linewidth]{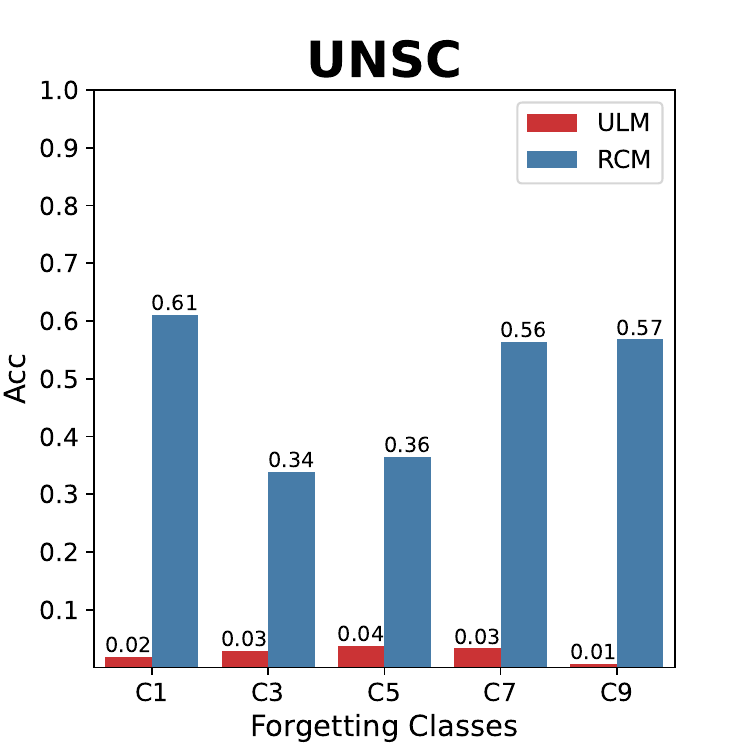}
        \caption{Comparison of the \textit{Acc} between ULM and RCM (\textbf{closed-source case}) after MRA w.r.t. each forgetting class on CIFAR-10 dataset. Due to space limit, \textbf{LARGER figures} can be found in the online extended version.}
        \label{fig:close_source_cifar10}
    \end{minipage}
     \begin{minipage}[c]{\linewidth}
    	\includegraphics[width=0.163\linewidth]{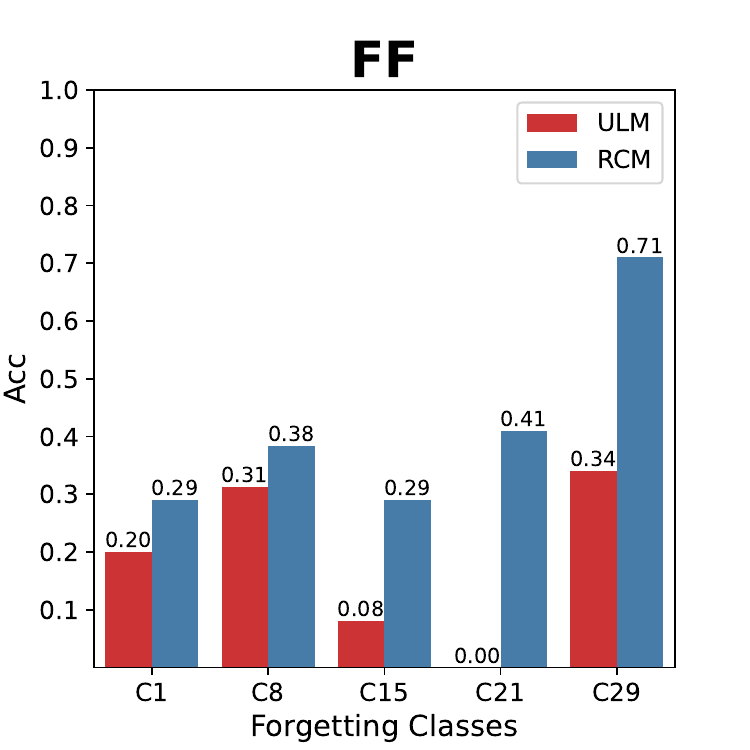}
        \includegraphics[width=0.163\linewidth]{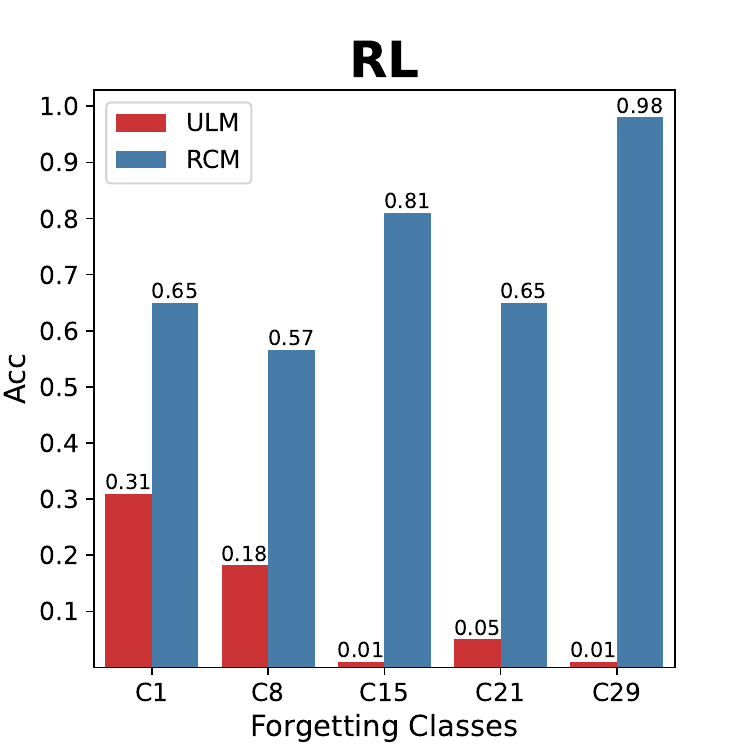}
        \includegraphics[width=0.163\linewidth]{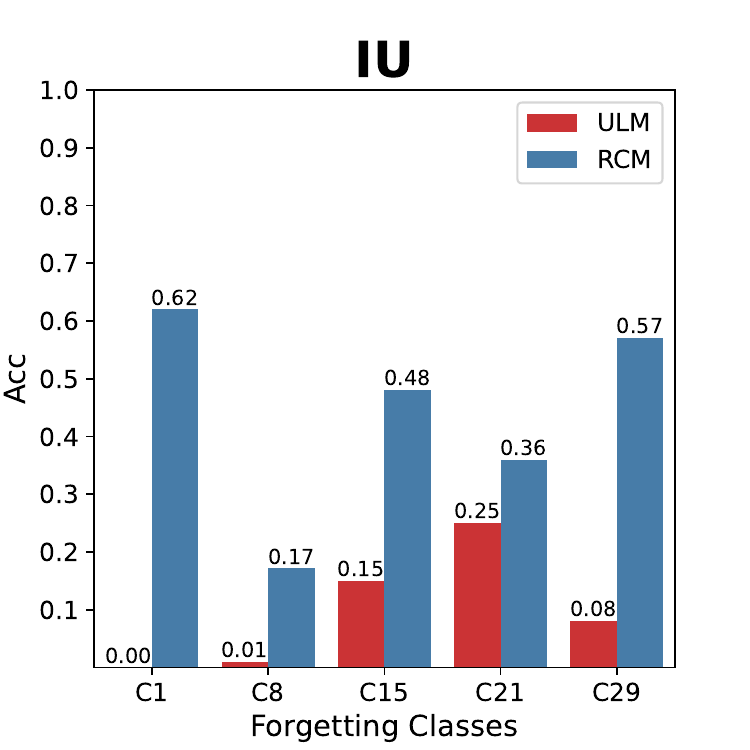}
        \includegraphics[width=0.163\linewidth]{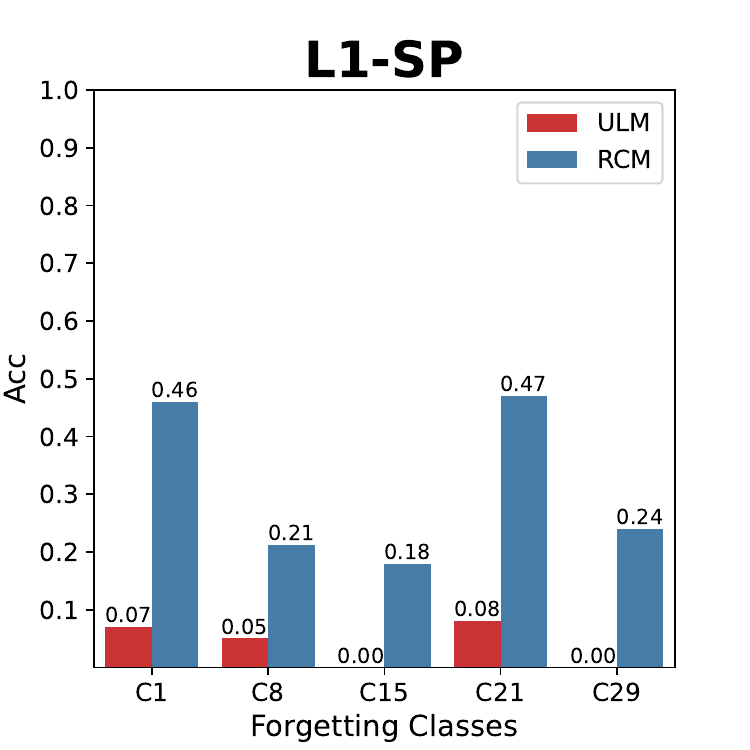}
         \includegraphics[width=0.163\linewidth]{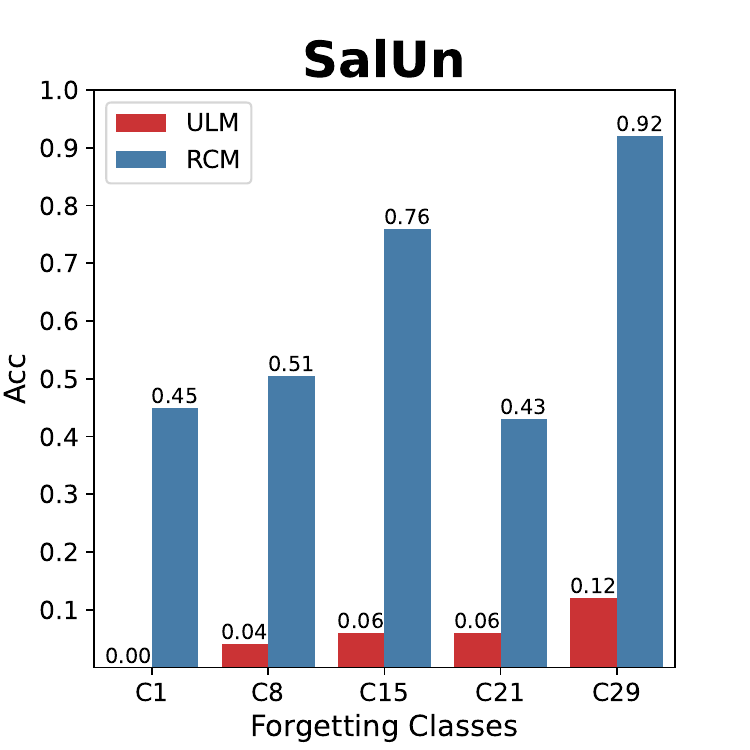}
         \includegraphics[width=0.163\linewidth]{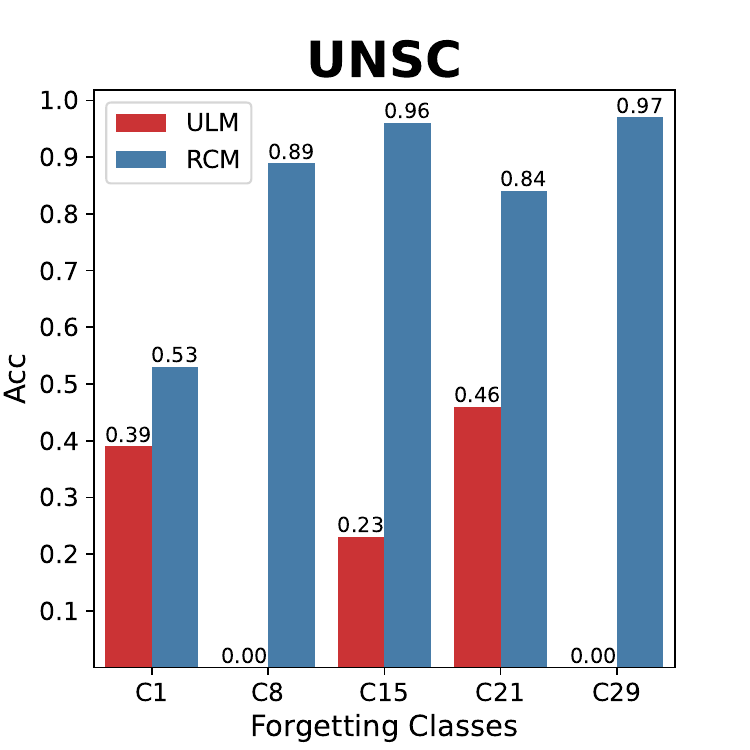}
        \caption{Comparison of the \textit{Acc} between ULM and RCM (\textbf{closed-source case}) w.r.t. each forgetting class on Pet-37 dataset.}
        \label{fig:close_source_pet37}
    \end{minipage}
\end{figure*}

\subsubsection{Data Preparation}
In our experiments, four real datasets are used, which cover both low- and high-resolution data, providing a comprehensive evaluation. CIFAR-10 and CIFAR-100 \cite{cifar} consist of 60,000 low-resolution images classified into 10 and 100 classes, respectively. Oxford-IIIT Pet (Pet-37) \cite{pet-37} contains 7,349 high-resolution images of cats and dogs in 37 classes. In particular, Oxford 102 Flower (Flower-102) ~\cite{flower-102} has 8,189 high-resolution images in 102 classes, where each class consists of between 40 and 258 images.

As shown in \Cref{tab:dataset_details}, we use the official splits of the training dataset $\mathcal{D}_{tr}$ and the testing dataset $\mathcal{D}_{ts}$ provided in the dataset package. For each dataset, five classes are selected to construct the forgetting dataset ${\mathcal{D}_f}$ by randomly sampling 50\% of data from $\mathcal{D}_{tr}$, and the prediction dataset used for evaluation is constructed by mixing $\mathcal{D}_{ts}$ and ${\mathcal{D}_f}$, i.e., $\mathcal{D}_{p}=\mathcal{D}_{ts}\cup\mathcal{D}_{f}$ to jointly assess the prediction and recovery capability of RCM.

\subsubsection{Model Configuration}
The proposed MRA framework is model-agnostic, so we use EfficientNet (EFN) \cite{efficientnet} on CIFAR datasets, ResNet \cite{resnet18} on Pet-37, and Swin-Transformer
(Swin-T) \cite{liu2022swin} on Flower-102 for a comprehensive study.
Moreover, in our framework, the STM does not necessarily have the same architecture as the ULM (Teacher Model). Therefore, we also evaluate the case of heterogeneous architectures on Flower-102, where the ULM is based on Swin-T while the STM is based on ResNet, as shown in \Cref{tab:main_close_source,tab:main_open_source}.

We use SGD optimizer for MU methods with a momentum of 0.9, a weight decay of 0.005, and AdamW for our MRA scheme. Other more detailed settings can be found in the online extended version. For each comparison model, we carefully tuned their hyperparameters to achieve optimal performance.

\subsubsection{MU Methods for MRA Evaluation} 
In the experiments, a set of SOTA MU methods, including \textbf{GA} \cite{graves2021amnesiac}, \textbf{RL} \cite{golatkar2020eternal}, \textbf{FF} \cite{becker2022evaluating}, \textbf{IU} \cite{koh2017understanding}, \textbf{BU} \cite{chen2024bu},\textbf{ L1-SP} \cite{Jia2023ModelSC}, \textbf{UNSC} \cite{ijcai2024p40} and \textbf{SalUn} \cite{fan2024salun}, are involved to comprehensively evaluate the recall capability of proposed MRA. In particular, \textbf{UNSC} does not support the Swin-T architecture, so the evaluation results on Flower-102 are not available.

\subsection{MRA in The Closed-source Case}
First, we evaluated the performance of MRA in the closed-source case, i.e., ULMs are used as a black-box service where the model parameters are not accessible.

\subsubsection{Overall MRA Efficacy Analysis}  \Cref{tab:main_close_source} show the accuracy (\textit{Acc}) of prediction dataset $\mathcal{D}_{p}$ in terms of its subsets $\mathcal{D}_{ts}$ and $\mathcal{D}_{ts}$ respectively. We compared the \textit{Acc} of ULM and RCM over four datasets and diverse configurations of the MU model. 
From the point of view of this table, all of the \textit{Acc} on $\mathcal{D}_f$ of TRM are 1.000 after training, while they drop to low \textit{Acc} after MU, illustrating that all selected MU methods can successfully mitigate the influence of $\mathcal{D}_f$ from well-trained models.
Furthermore, according to all the results of improvement ($\Delta Acc$) on both $\mathcal{D}_{f}$ and $\mathcal{D}_{ts}$, we find that RCMs can unexceptionally improve the prediction accuracy on all datasets for all MU models, which overall proves that the proposed MRA is an effective and versatile model-agnostic framework to recover the class memberships of forgotten instances from ULMs. 

More specifically, the improvement ($\Delta Acc$) of \textbf{L1-SP} is overall smaller than that of other MU methods on both $\mathcal{D}_{f}$ and $\mathcal{D}_{ts}$. This can be attributed to the weight pruning on TRM, which makes the parameters of ULM significantly different from those of TRM. As a result, the knowledge distillation from ULM is prone to having higher label noise. In contrast, the improvement ($\Delta Acc$) of \textbf{SalUn} and \textbf{UNSC} is overall larger than that of other MU methods. This is because most MU methods degrade the model performance after unlearning, known as ``over-unlearning''. In comparison, \textbf{SalUn} and \textbf{UNSC}, can precisely unlearn
target forgetting samples without over-unlearning. As a result, \textbf{SalUn} and \textbf{UNSC} can provide less noisy pseudo labels for knowledge distillation, leading to better recall from ULMs.

According to the observation and analysis above, it reveals a phenomenon that \emph{``The MU methods in precise unlearning may lead to high success rate to recall the forgotten class memberships via MRA''}. As a result, MRA can serve as a valuable tool to assess the potential risk of privacy leakage for MU methods, thus facilitating the development of more robust MU models.

\subsubsection{Demonstration of Class-specific Recovery Efficacy}
To intuitively demonstrate the capacity of MRA to recall the forgotten class memberships on the prediction images $\mathcal{X}_f$, we further conducted detailed evaluations with respect to each forgetting class. 
\Cref{fig:close_source_cifar10,fig:close_source_pet37} demonstrate the comparison of the \textit{Acc} between ULM and RCM after MRA on CIFAR-10 and Pet-37 for each forgetting class. By checking the improvement of \textit{Acc} for each class, we can observe a similar phenomenon as shown in \Cref{tab:main_close_source}. For example, the improvement of \textit{Acc} for each class on \textbf{L1-SP} is relatively small due to over-unlearning. In comparison, the ULM via \textbf{SalUn} can precisely unlearn the target forgetting images, that is, $\mathcal{X}_f$, which leads to very low \textit{Acc} for each forgetting class, whereas the improvement for each class after MRA is the most significant. That is, \emph{``precise forgetting, easy recalling''}.

\begin{table*}[tbh!]
  \centering
  \scalebox{0.75}{
    \begin{tabular}{c|c|c|c|c|c|c|c|c|c|c|c}
    \toprule
          &       & \textbf{TRM}   &       & \textbf{FF} & \textbf{RL}    & \textbf{GA}    & \textbf{IU}    & \textbf{BU}    & \textbf{L1-SP} & \textbf{SalUn} & \textbf{UNSC} \\
    \midrule
    \midrule
    \multicolumn{1}{c|}{\multirow{6}[12]{*}{\shortstack{CIFAR-10\\T: EFN\\S: EFN}}} & \multirow{3}[6]{*}{$\mathcal{D}_{ts}$} & \multirow{3}[6]{*}{0.833 } & ULM   & 0.164  & 0.388  & 0.505  & 0.105  & 0.462  & 0.451  & 0.486  & 0.193  \\
\cmidrule{4-12}          &       &       & RCM   & 0.708  & 0.471  & 0.582  & 0.689  & 0.783  & 0.479  & 0.829  & 0.774  \\
\cmidrule{4-12}          &       &       & $\Delta Acc$ & \textbf{0.544 } & \textbf{\textcolor{blue}{0.083} } & \textbf{0.077 } & \textbf{\textcolor{red}{0.584} } & \textbf{0.321 } & \textbf{\textcolor{blue}{0.027} } & \textbf{0.343 } & \textbf{\textcolor{red}{0.581} } \\
\cmidrule{2-12}          & \multirow{3}[6]{*}{$\mathcal{D}_{f}$} & \multirow{3}[6]{*}{1.000 } & ULM   & 0.142  & 0.153  & 0.267  & 0.090  & 0.082  & 0.154  & 0.117  & 0.031  \\
\cmidrule{4-12}          &       &       & RCM   & 0.726  & 0.351  & 0.509  & 0.759  & 0.877  & 0.338  & 0.997  & 0.921  \\
\cmidrule{4-12}          &       &       & $\Delta Acc$ & \textbf{0.584 } & \textbf{\textcolor{blue}{0.197} } & \textbf{0.242 } & \textbf{0.668 } & \textbf{0.795 } & \textbf{\textcolor{blue}{0.184} } & \textbf{\textcolor{red}{0.880} } & \textbf{\textcolor{red}{0.890} } \\
    \midrule
    \midrule
    \multicolumn{1}{c|}{\multirow{6}[12]{*}{\shortstack{CIFAR-100\\T: EFN\\S: EFN}}} & \multirow{3}[6]{*}{$\mathcal{D}_{ts}$} & \multirow{3}[6]{*}{0.643 } & ULM   & 0.159  & 0.593  & 0.531  & 0.140  & 0.529  & 0.537  & 0.410  & 0.275  \\
\cmidrule{4-12}          &       &       & RCM   & 0.521  & 0.599  & 0.589  & 0.596  & 0.587  & 0.598  & 0.564  & 0.598  \\
\cmidrule{4-12}          &       &       & $\Delta Acc$ & \textbf{\textcolor{red}{0.363} } & \textbf{\textcolor{blue}{0.006} } & \textbf{0.058 } & \textbf{\textcolor{red}{0.456} } & \textbf{0.058 } & \textbf{\textcolor{blue}{0.061} } & \textbf{0.154 } & \textbf{0.324 } \\
\cmidrule{2-12}          & \multirow{3}[6]{*}{$\mathcal{D}_{f}$} & \multirow{3}[6]{*}{1.000 } & ULM   & 0.094  & 0.086  & 0.201  & 0.201  & 0.254  & 0.199  & 0.227  & 0.209  \\
\cmidrule{4-12}          &       &       & RCM   & 0.686  & 0.898  & 0.682  & 0.967  & 0.970  & 0.764  & 0.945  & 0.985  \\
\cmidrule{4-12}          &       &       & $\Delta Acc$ & \textbf{0.592 } & \textbf{\textcolor{red}{0.813} } & \textbf{\textcolor{blue}{0.481} } & \textbf{0.766 } & \textbf{0.715 } & \textbf{\textcolor{blue}{0.565} } & \textbf{0.718 } & \textbf{\textcolor{red}{0.776} } \\
    \midrule
    \midrule
    \multicolumn{1}{c|}{\multirow{6}[12]{*}{\shortstack{Pet-37\\T: ResNet\\S: ResNet}}} & \multirow{3}[6]{*}{$\mathcal{D}_{ts}$} & \multirow{3}[6]{*}{0.895 } & ULM   & 0.257  & 0.754  & 0.740  & 0.622  & 0.646  & 0.740  & 0.706  & 0.769  \\
\cmidrule{4-12}          &       &       & RCM   & 0.782  & 0.850  & 0.843  & 0.838  & 0.846  & 0.841  & 0.849  & 0.856  \\
\cmidrule{4-12}          &       &       & $\Delta Acc$ & \textbf{\textcolor{red}{0.525} } & \textbf{\textcolor{blue}{0.096} } & \textbf{0.103 } & \textbf{\textcolor{red}{0.216} } & \textbf{0.200 } & \textbf{0.102 } & \textbf{0.143 } & \textbf{\textcolor{blue}{0.087} } \\
\cmidrule{2-12}          & \multirow{3}[6]{*}{$\mathcal{D}_{f}$} & \multirow{3}[6]{*}{1.000 } & ULM   & 0.224  & 0.332  & 0.224  & 0.200  & 0.084  & 0.196  & 0.128  & 0.304  \\
\cmidrule{4-12}          &       &       & RCM   & 0.904  & 0.960  & 0.952  & 0.936  & 0.960  & 0.920  & 0.948  & 0.976  \\
\cmidrule{4-12}          &       &       & $\Delta Acc$ & \textbf{0.680 } & \textbf{\textcolor{blue}{0.628} } & \textbf{0.728 } & \textbf{0.736 } & \textbf{\textcolor{red}{0.876} } & \textbf{0.724 } & \textbf{\textcolor{red}{0.820} } & \textbf{\textcolor{blue}{0.672} } \\
    \midrule
    \midrule
    \multicolumn{1}{c|}{\multirow{6}[11]{*}{\shortstack{Flower102\\T: Swin-T\\S: ResNet}}} & \multirow{3}[6]{*}{$\mathcal{D}_{ts}$} & \multirow{3}[6]{*}{0.939 } & ULM   & 0.294  & 0.756  & 0.314  & 0.512  & 0.706  & 0.530  & 0.496  & NA \\
\cmidrule{4-12}          &       &       & RCM   & 0.607  & 0.915  & 0.772  & 0.829  & 0.889  & 0.856  & 0.857  & NA \\
\cmidrule{4-12}          &       &       & $\Delta Acc$ & \textbf{0.313 } & \textbf{\textcolor{blue}{0.159} } & \textbf{\textcolor{red}{0.458} } & \textbf{0.318 } & \textbf{\textcolor{blue}{0.183} } & \textbf{0.325 } & \textbf{\textcolor{red}{0.361} } & NA \\
\cmidrule{2-12}          & \multirow{3}[5]{*}{$\mathcal{D}_{f}$} & \multirow{3}[5]{*}{1.000 } & ULM   & 0.235  & 0.150  & 0.239  & 0.291  & 0.340  & 0.324  & 0.267  & NA \\
\cmidrule{4-12}          &       &       & RCM   & 0.482  & 0.988  & 0.709  & 0.725  & 0.972  & 0.935  & 0.915  & NA \\
\cmidrule{4-12}          &       &       & $\Delta Acc$ & \textbf{\textcolor{blue}{0.247} } & \textbf{\textcolor{red}{0.838} } & \textbf{0.470 } & \textbf{\textcolor{blue}{0.433} } & \textbf{0.632 } & \textbf{0.611 } & \textbf{\textcolor{red}{0.648} } & NA \\
\bottomrule
\end{tabular}%
}
\captionof{table}{Performance comparison of MRA (\textbf{open-source case}) on various SOTA MU methods, where ULM indicates the \textit{Acc} after MU while RCM indicates the \textit{Acc} after MRA, and $\Delta Acc$ shows the improvement (the Top-2 $\Delta Acc$ are marked in \textcolor{red}{red} while the Lowest-2 are marked in \textcolor{blue}{blue}). TRM illustrates the \textit{Acc} of original model.}
\label{tab:main_open_source}%
\end{table*}

\begin{figure*}
\begin{minipage}[c]{\textwidth}
    \centering
    \begin{minipage}[c]{\linewidth}
    	\includegraphics[width=0.163\linewidth]{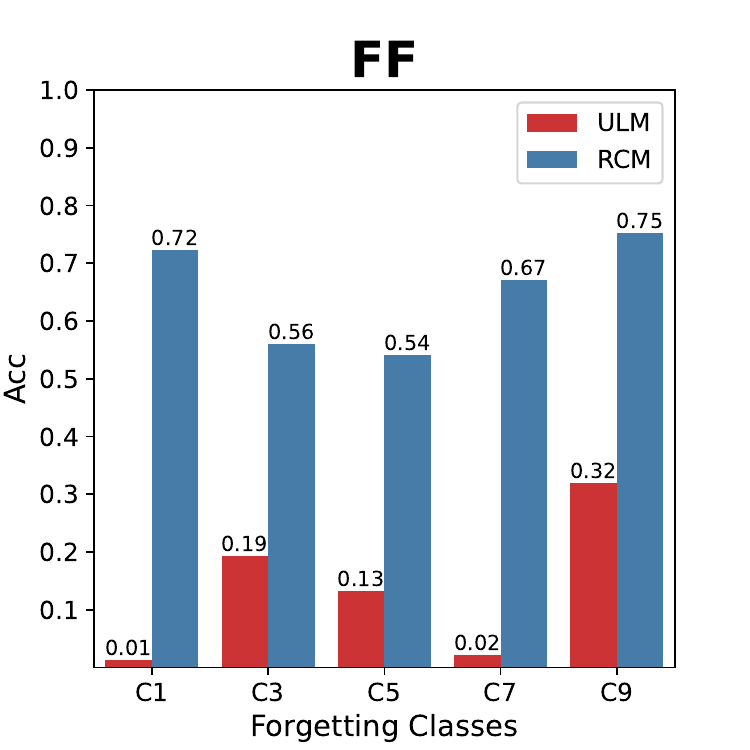}
        \includegraphics[width=0.163\linewidth]{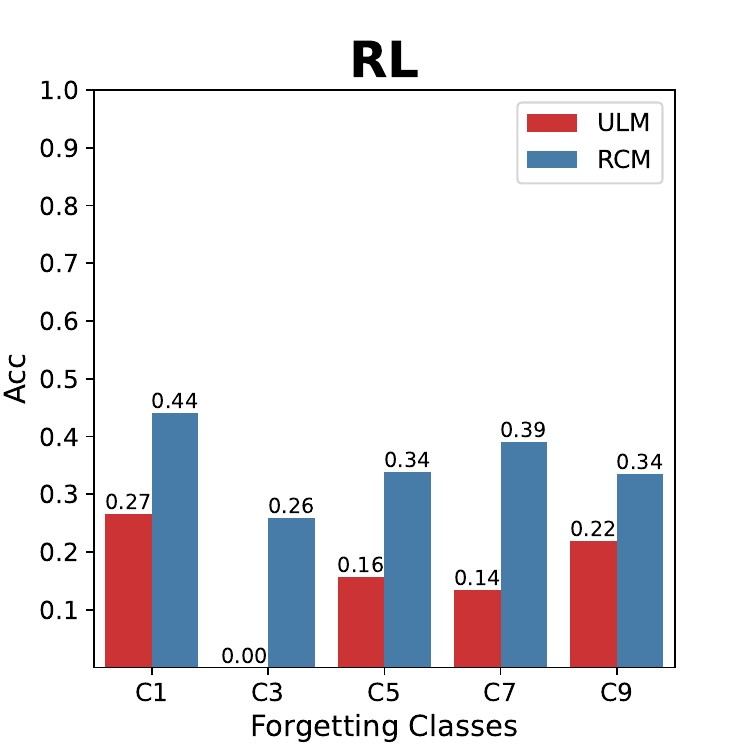}
        \includegraphics[width=0.163\linewidth]{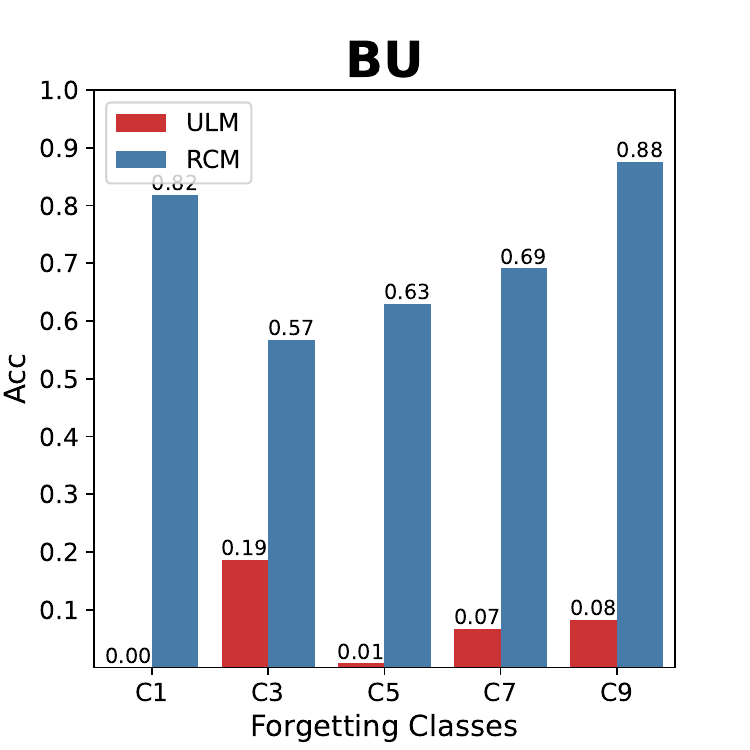}
        \includegraphics[width=0.163\linewidth]{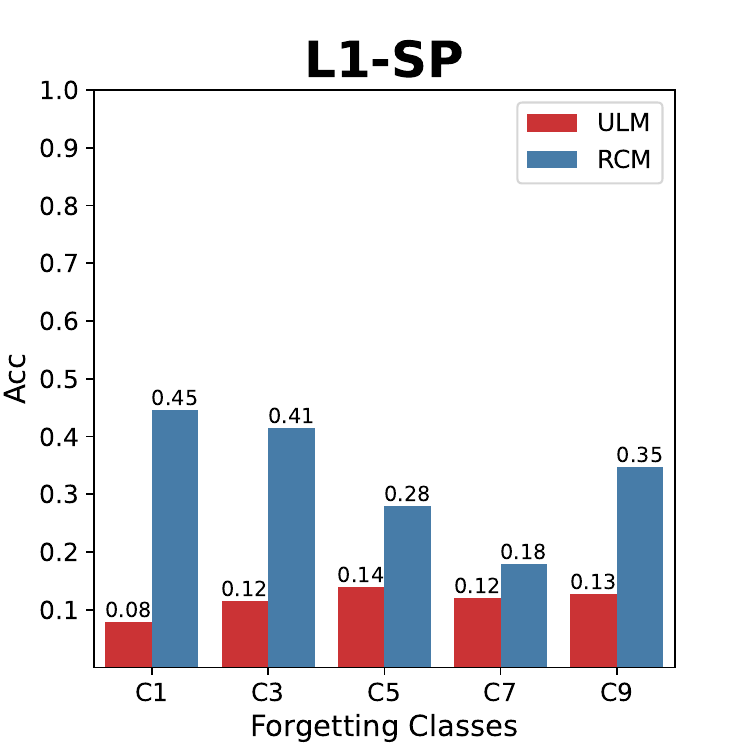}
         \includegraphics[width=0.163\linewidth]{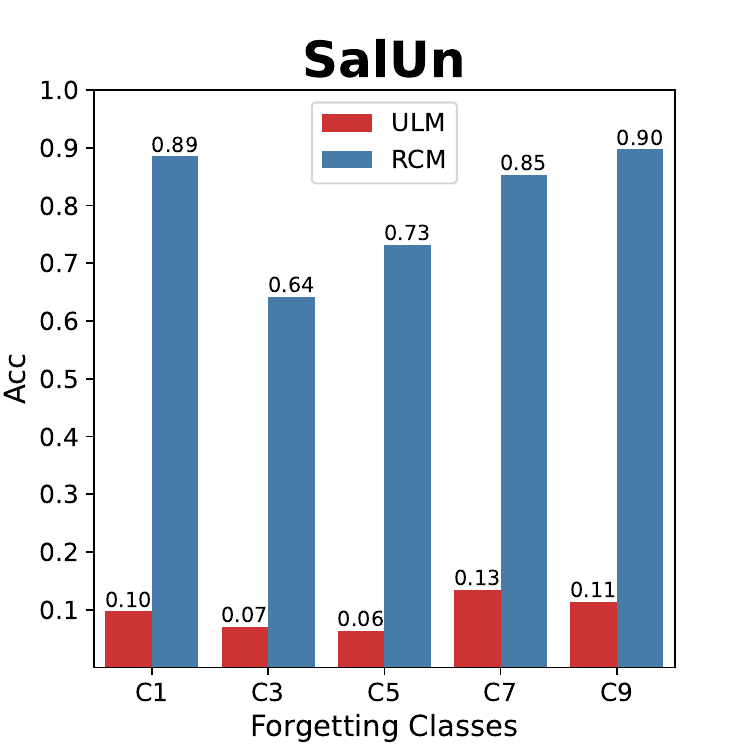}
         \includegraphics[width=0.163\linewidth]{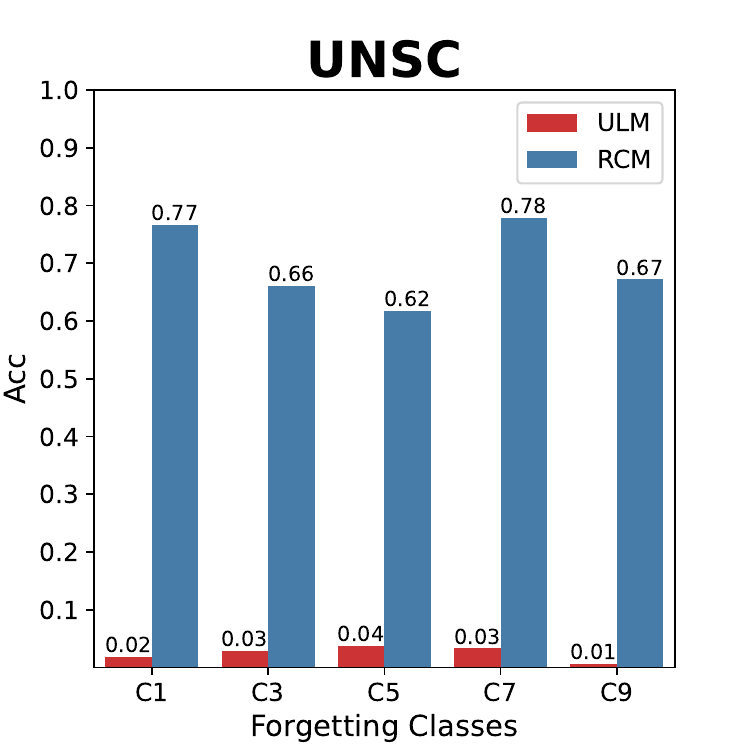}
        \caption{Comparison of the \textit{Acc} between ULM and RCM (\textbf{open-source case}) after MRA w.r.t. each forgetting class on CIFAR-10 dataset. Due to space limit, \textbf{LARGER figures} can be found in the online extended version.}
        \label{fig:open_source_cifar10}
    \end{minipage}
     \begin{minipage}[c]{\linewidth}
    	\includegraphics[width=0.163\linewidth]{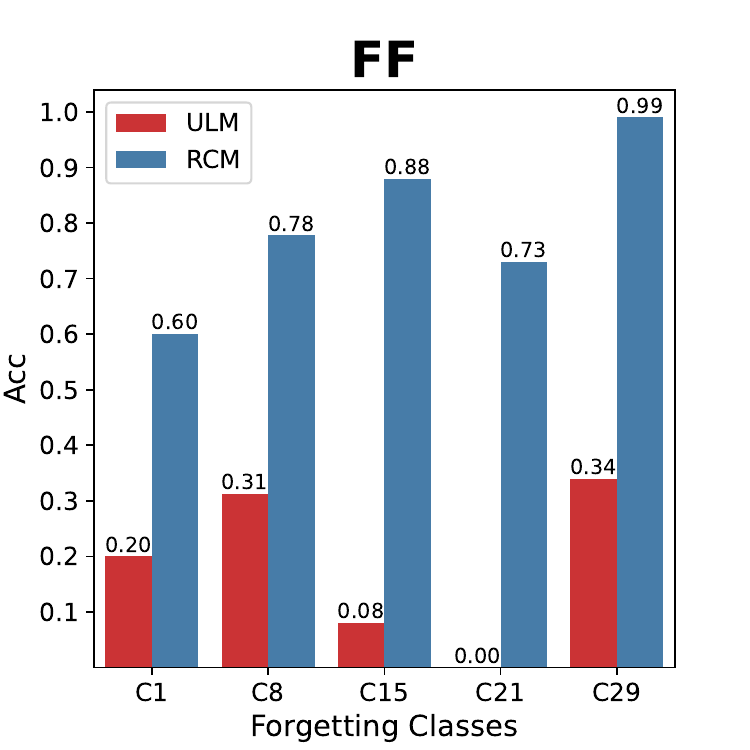}
        \includegraphics[width=0.163\linewidth]{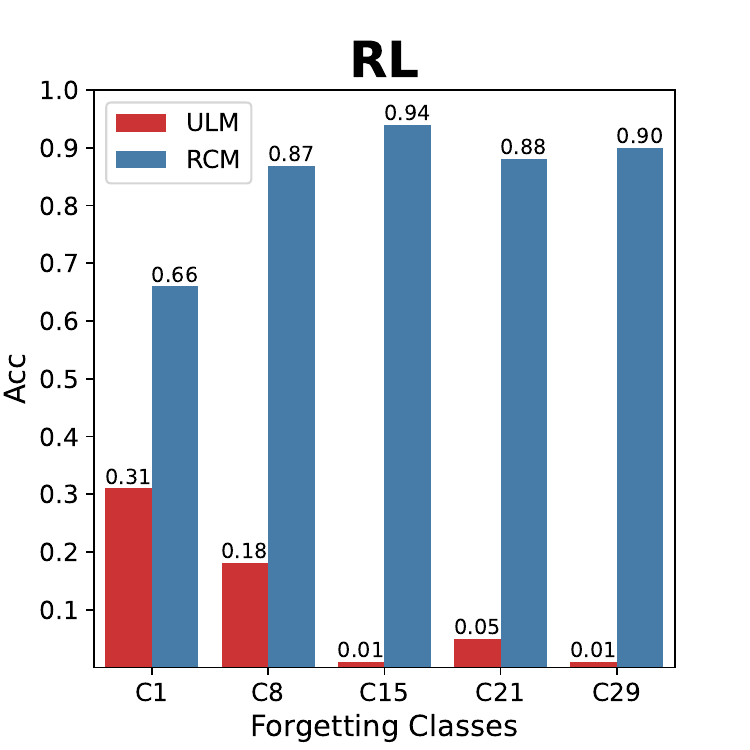}
        \includegraphics[width=0.163\linewidth]{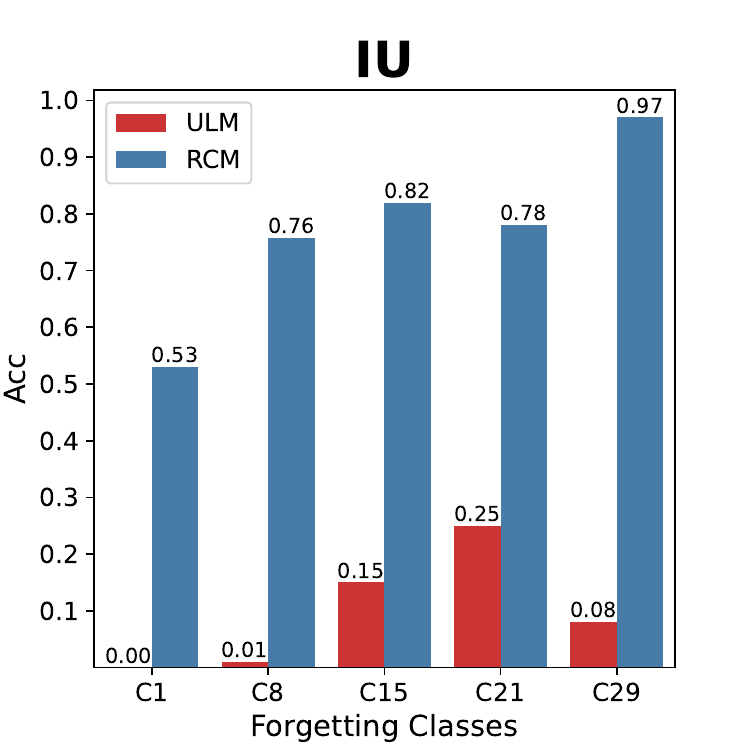}
        \includegraphics[width=0.163\linewidth]{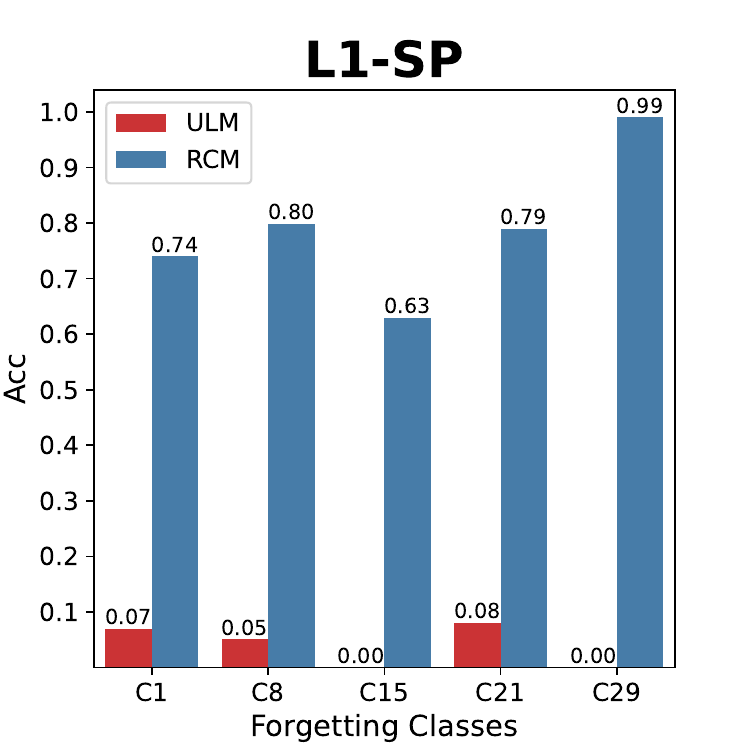}
         \includegraphics[width=0.163\linewidth]{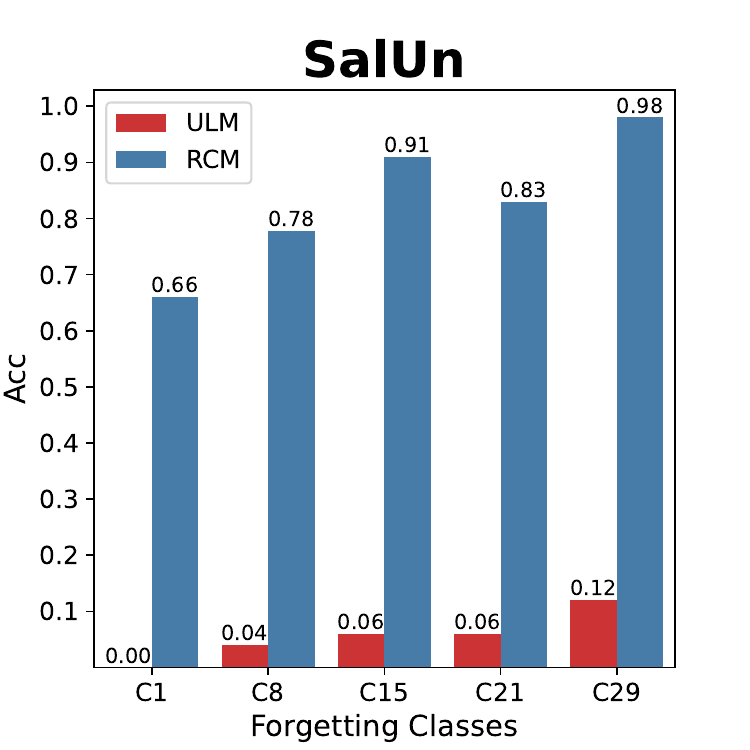}
         \includegraphics[width=0.163\linewidth]{figures/results/pet-37/fr_0.5_ijcai/UNSC_resnet18_restore_bar.pdf}
        \caption{Comparison of the \textit{Acc} between ULM and RCM (\textbf{open-source case}) w.r.t. each forgetting class on Pet-37 dataset.}
        \label{fig:open_source_pet37}
    \end{minipage}
\end{minipage}
\end{figure*}

\subsection{MRA in The Open-source Case}\label{sec:exp_mra_open_source}
In comparison to the closed-source case, the ULMs are released with their parameters in the open-source case. As a result, the ULMs can be updated during the MRA process.

\subsubsection{Overall MRA Efficacy Analysis}
For the open-source case, we can find that the improvement ($\Delta Acc$) of \textbf{SalUn} and \textbf{UNSC} is significant again due to the same reason presented above.
Comparing \Cref{tab:main_open_source} with \Cref{tab:main_close_source}, it is easy to find that the improvement of \textit{Acc} in the open-source case method is significantly greater than that in the closed-source case.
Especially, we find \textbf{IU} achieves the Lowest-2 improvement three times in the closed-source case but it achieves the Top-2 improvement three times in the open-source case.
This is because the Confident Membership Recall step (cf. \Cref{alg:mra}) can effectively recall the knowledge retained by the ULM (as a noisy labeler) using confident pseudo-label samples. Then, the improved teacher model can distill less noisy knowledge to the STM. This alternative optimization process results in significant improvement.

\subsubsection{Demonstration of Class-specific Recovery Efficacy}
The analogy to the closed-source case, \Cref{fig:open_source_cifar10,fig:open_source_pet37} demonstrate the comparison of the \textit{Acc} between ULM and RCM after MRA on CIFAR-10 and Pet-37 for each forgetting class in the open-source case. Compared to \Cref{fig:close_source_cifar10,fig:close_source_pet37}, we can find that the gaps between different MU methods are much smaller in \Cref{fig:open_source_cifar10,fig:open_source_pet37}.
Especially, we can find the improvement of \textit{Acc} on different MU methods after the MRA is close for each forgetting class in \Cref{fig:open_source_pet37}. Even the over-unlearning models, the ULMs via \textbf{IU} and \textbf{L1-SP} are effectively recalled their forgotten instances through the MRA process in terms of the balanced class membership recall strategy (cf. \Cref{sec:mra_scheme}).

\subsection{Ablation Study}
In this section, we discuss the effectiveness of each key component in the implementation of the MRA framework. Since the MRA framework consists of two alternative learning steps, we will evaluate the following components.

\noindent \textbf{DST:} This component is the \textit{Denosing Knowledge Distillation} step presented in \Cref{sec:mra_scheme}, which aims to distill knowledge from noisy labeling teacher $\mathcal{M}(\Theta_U)$ to STM $\Tilde{\mathcal{M}}(\Theta_S)$.

\noindent \textbf{STU:} The component serves as the class membership recall process for STM, as presented in the \textit{Confident Membership Recall} step. That is, \textbf{DST+STU} is equivalent to the closed-source case of MRA.

\noindent \textbf{TCH:} The component serves as the class membership recall process for teacher models, as presented in the \textit{Confident Membership Recall} step. That is, \textbf{DST+STU+TCH} is equivalent to the open-source case of MRA.


\begin{table}[htbp]
  \centering
  \scalebox{0.625}{
    \begin{tabular}{c|c|c||c|c|c|c|c|c|c|c}
    \toprule
    \multicolumn{3}{c||}{\textbf{Component}} & \multicolumn{2}{c|}{\textbf{FF}} & \multicolumn{2}{c|}{\textbf{BU}} & \multicolumn{2}{c|}{\textbf{SalUn}} & \multicolumn{2}{c}{\textbf{UNSC}} \\
    \midrule
    \textbf{DST} & \textbf{STU} & \textbf{TCH} &  $\mathcal{D}_{ts}$ & $\mathcal{D}_{f}$  &  $\mathcal{D}_{ts}$ & $\mathcal{D}_{f}$  & $\mathcal{D}_{ts}$ & $\mathcal{D}_{f}$  &  $\mathcal{D}_{ts}$ & $\mathcal{D}_{f}$ \\
    \midrule
    \checkmark &    &    & 0.223  & 0.168  & 0.659  & 0.128  & 0.698  & 0.108  & 0.767  & 0.320    
    \\
    \midrule
    \checkmark & \checkmark &       & 0.486  & 0.388  & 0.785  & 0.856  & 0.778  & 0.712  & 0.799  & 0.660  \\
    \midrule
    \checkmark & \checkmark & \checkmark & 0.782  & 0.904  & 0.846  & 0.960  & 0.849  & 0.948  & 0.856  & 0.976  \\
    \bottomrule
    \end{tabular}%
    }
  \caption{Ablation results (\textit{Acc}) of MRA on Pet-37 dataset}
  \label{tab:ablation}%
\end{table}%

\begin{figure*}[t!]
    \subfigure[\textbf{ULM}]{
        \begin{minipage}[t]{0.50\linewidth}
            \includegraphics[width=0.243\linewidth]{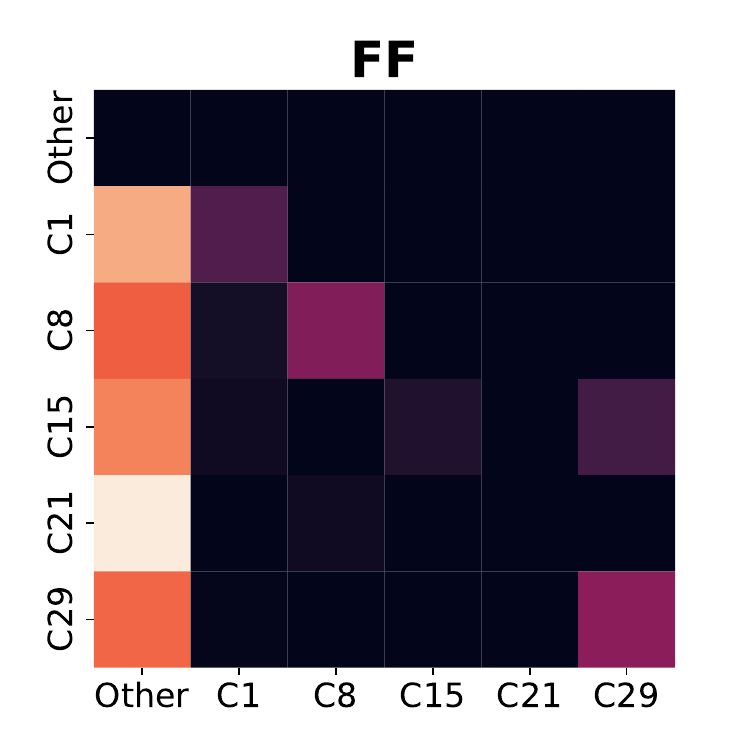}
            \includegraphics[width=0.243\linewidth]{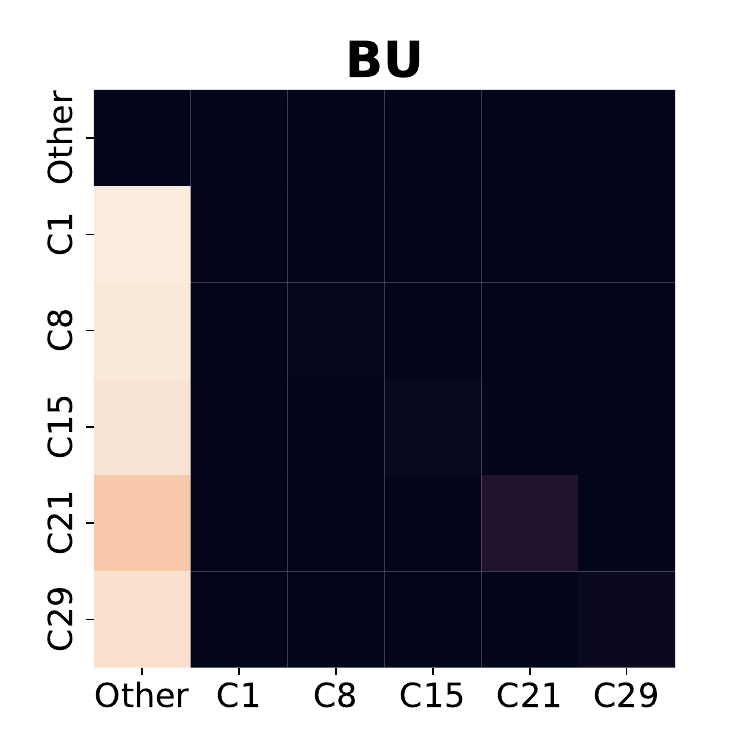}
            \includegraphics[width=0.243\linewidth]{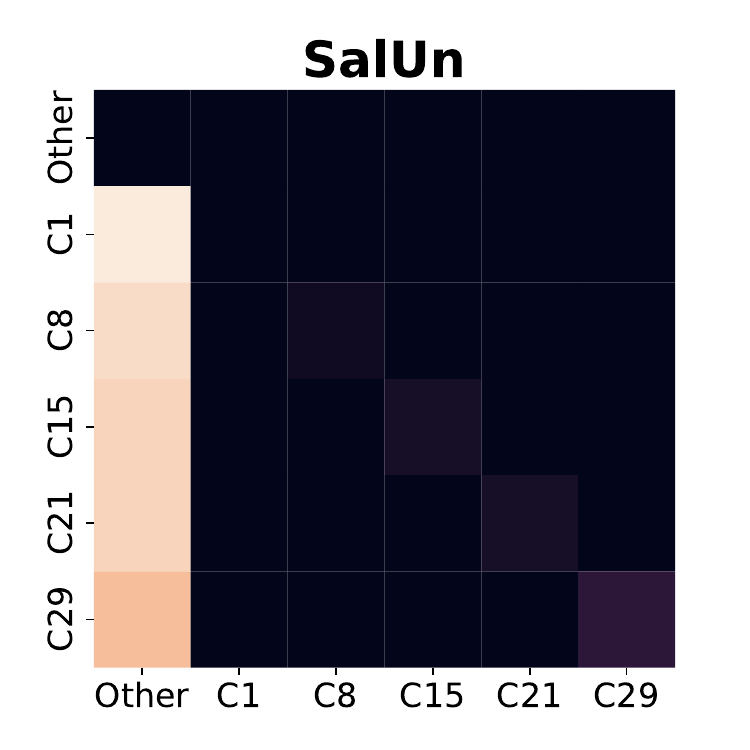}
            \includegraphics[width=0.243\linewidth]{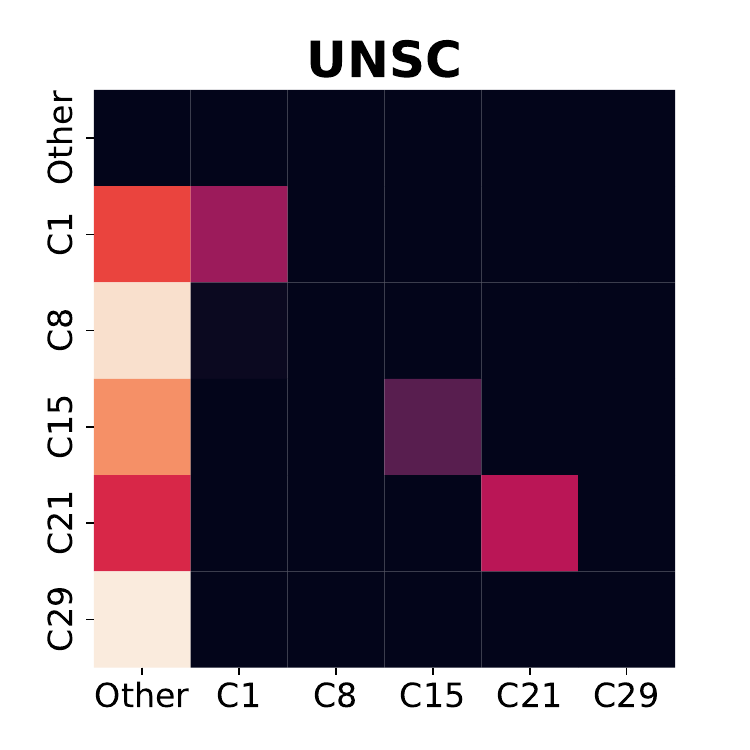}
        \end{minipage}
        }
     \subfigure[\textbf{DST}]{
        \begin{minipage}[t]{0.50\linewidth}
            \includegraphics[width=0.243\linewidth]{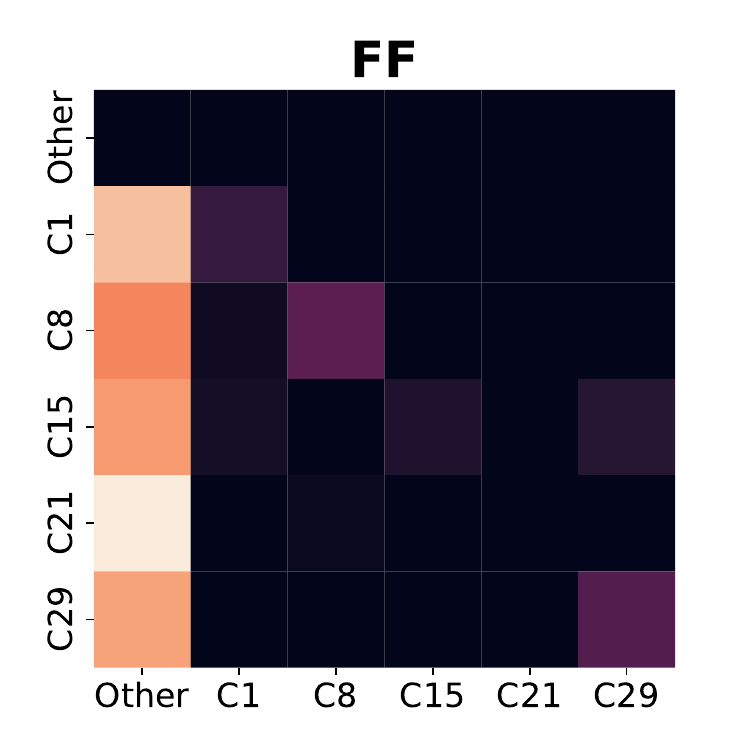}
            \includegraphics[width=0.243\linewidth]{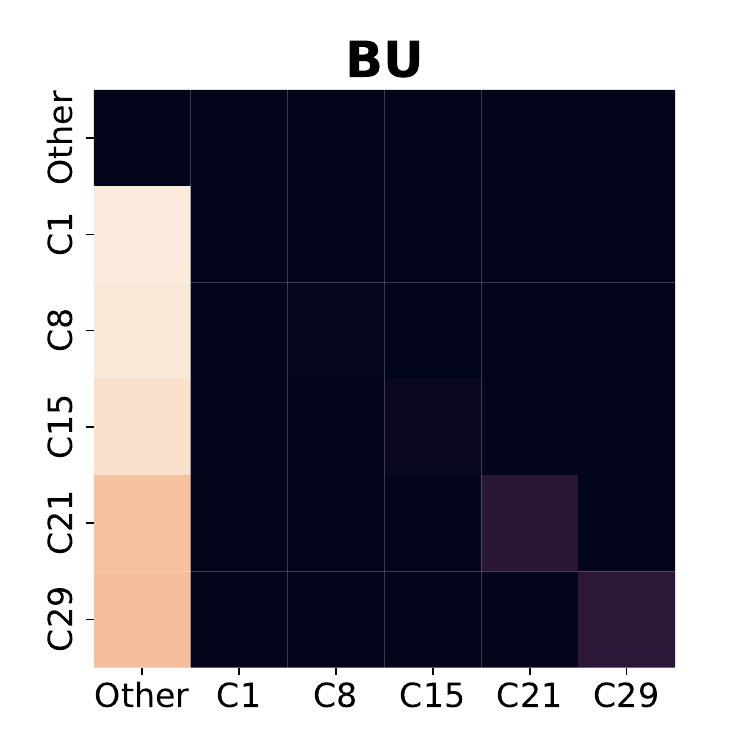}
            \includegraphics[width=0.243\linewidth]{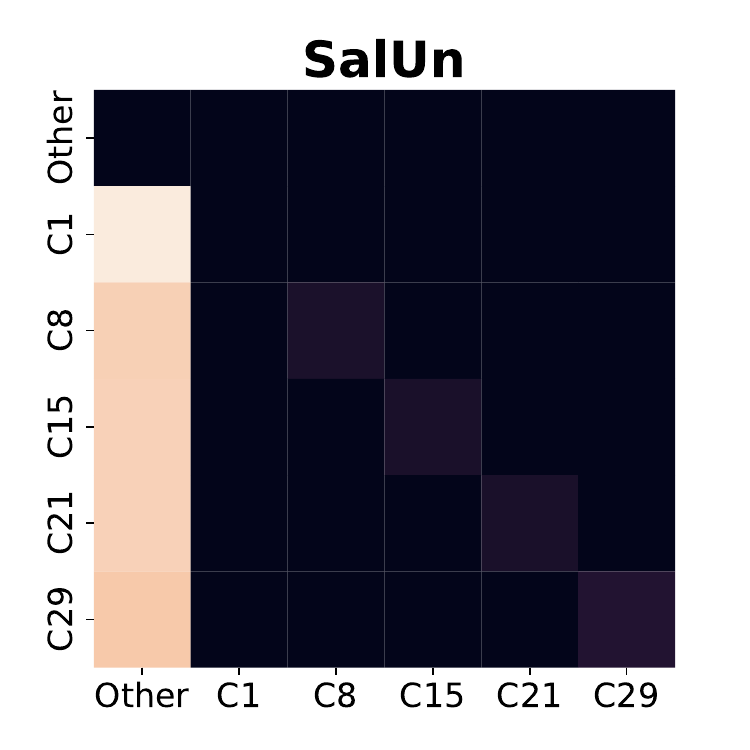}
            \includegraphics[width=0.243\linewidth]{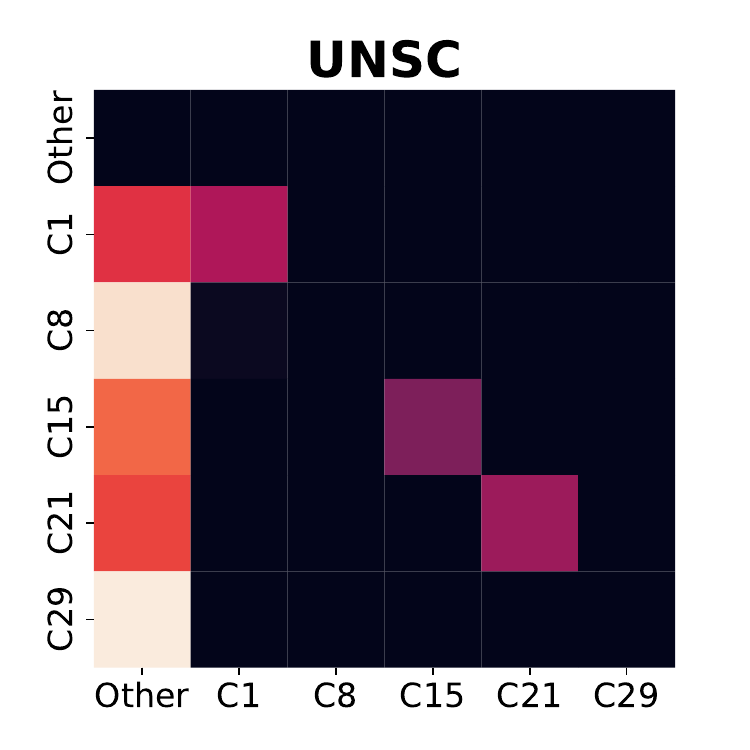}
        \end{minipage}
        }
     \subfigure[\textbf{DST+STU}]{
        \begin{minipage}[t]{0.50\linewidth}
            \includegraphics[width=0.243\linewidth]{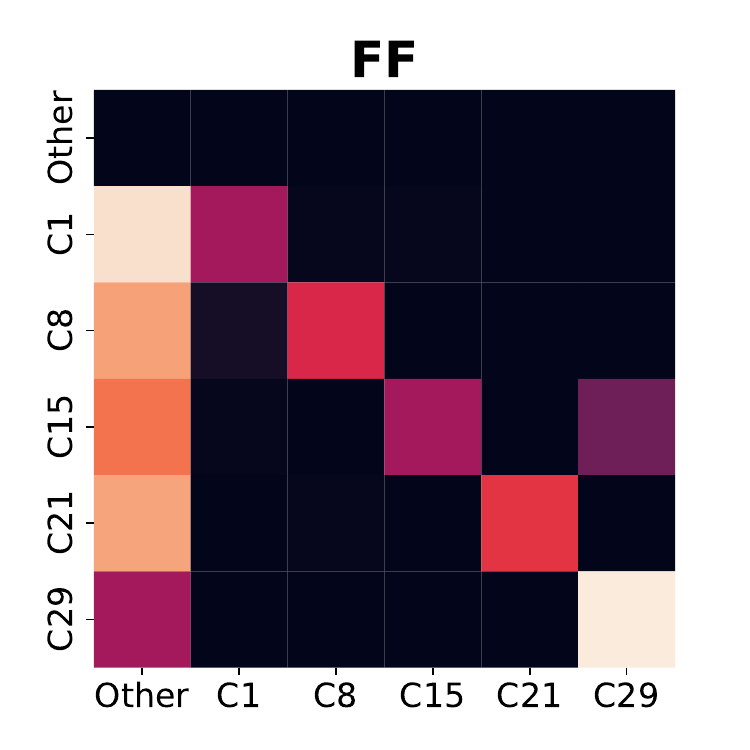}
            \includegraphics[width=0.243\linewidth]{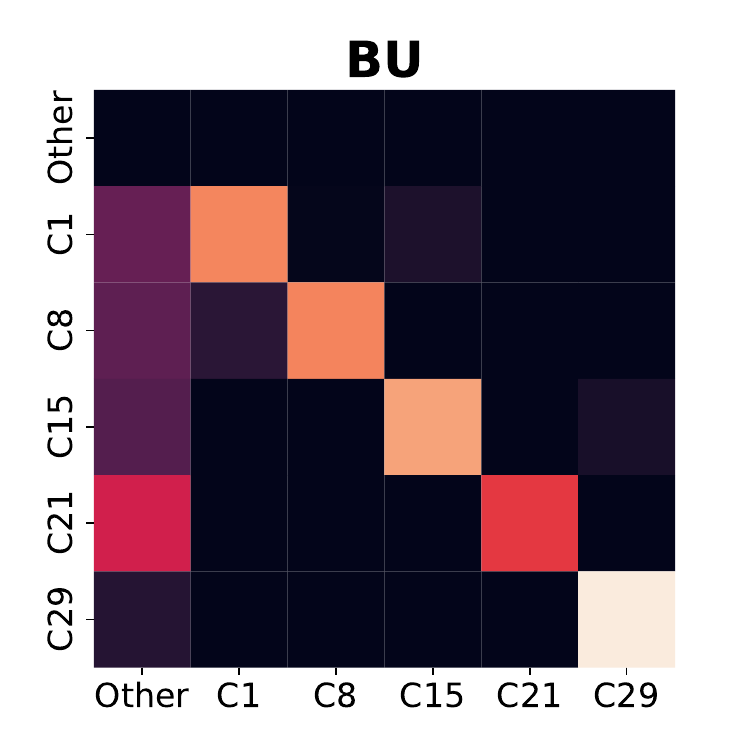}
            \includegraphics[width=0.243\linewidth]{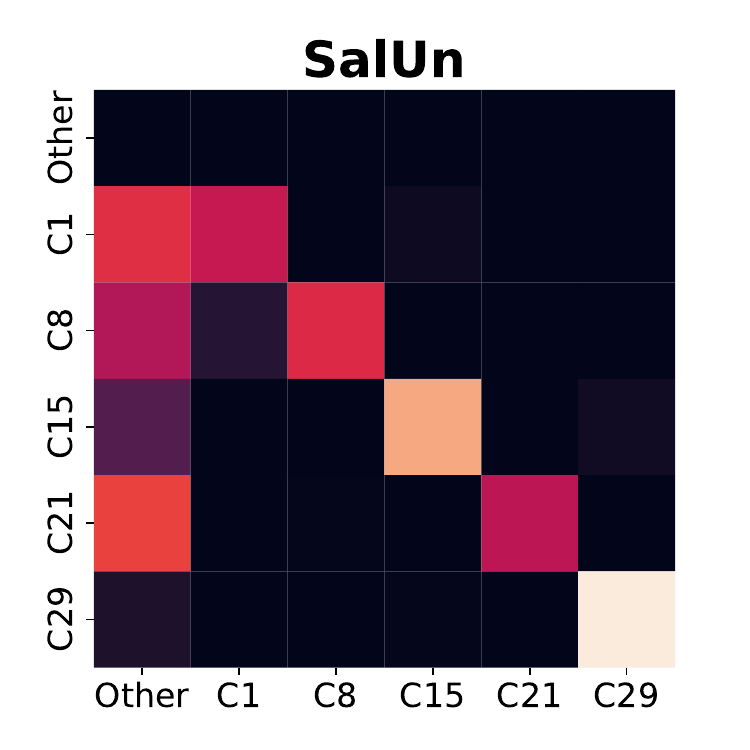}
            \includegraphics[width=0.243\linewidth]{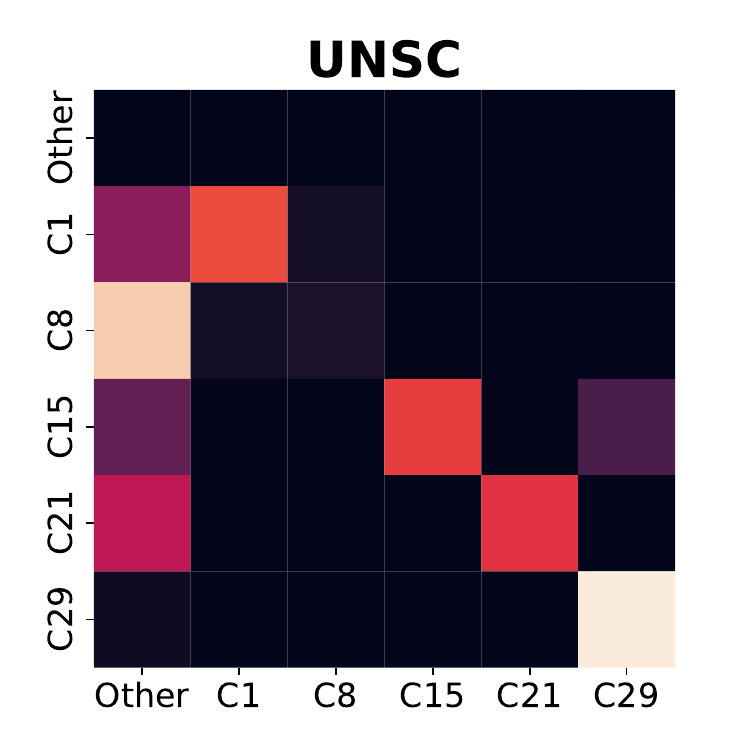}
        \end{minipage}
        }
     \subfigure[\textbf{DST+STU+TCH}]{
        \begin{minipage}[t]{0.50\linewidth}
            \includegraphics[width=0.243\linewidth]{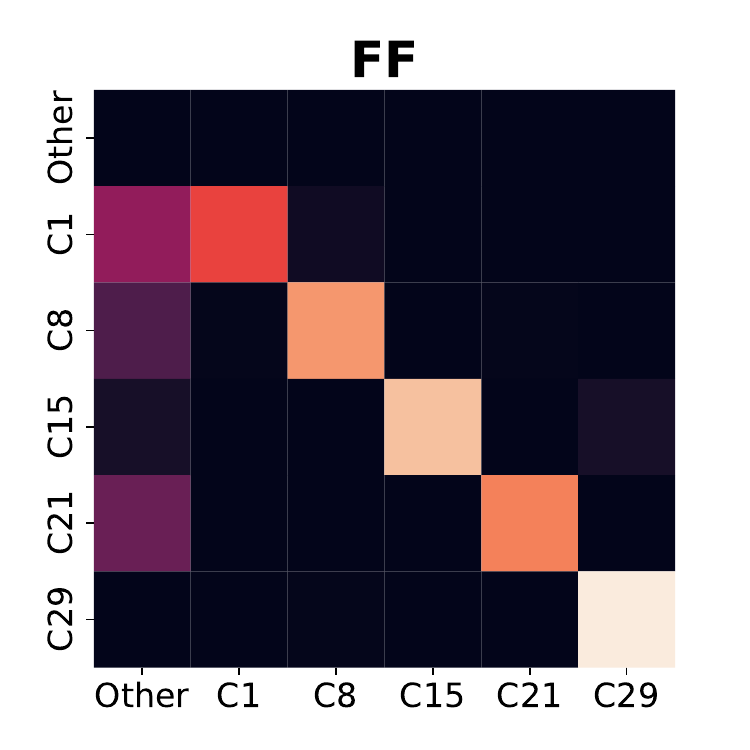}
            \includegraphics[width=0.243\linewidth]{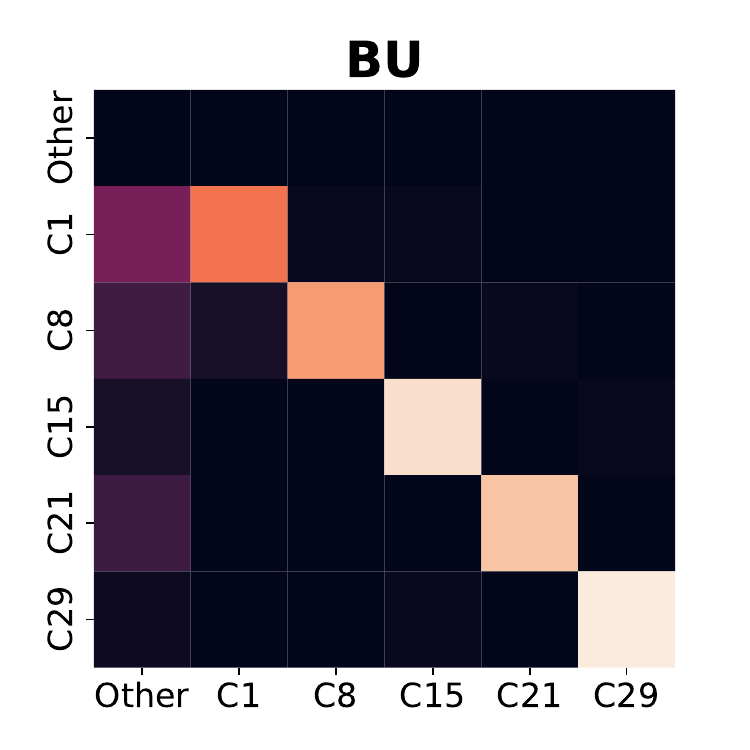}
            \includegraphics[width=0.243\linewidth]{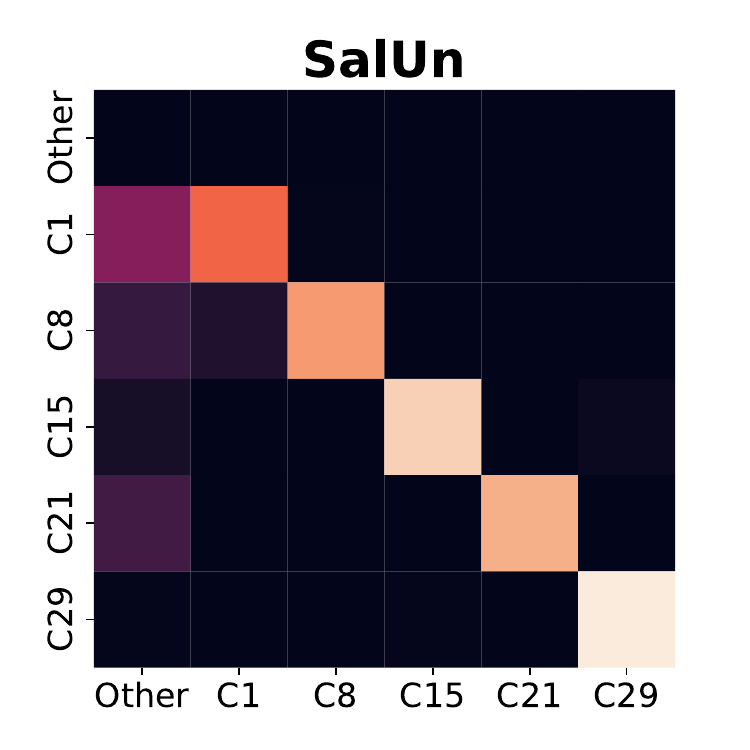}
            \includegraphics[width=0.243\linewidth]{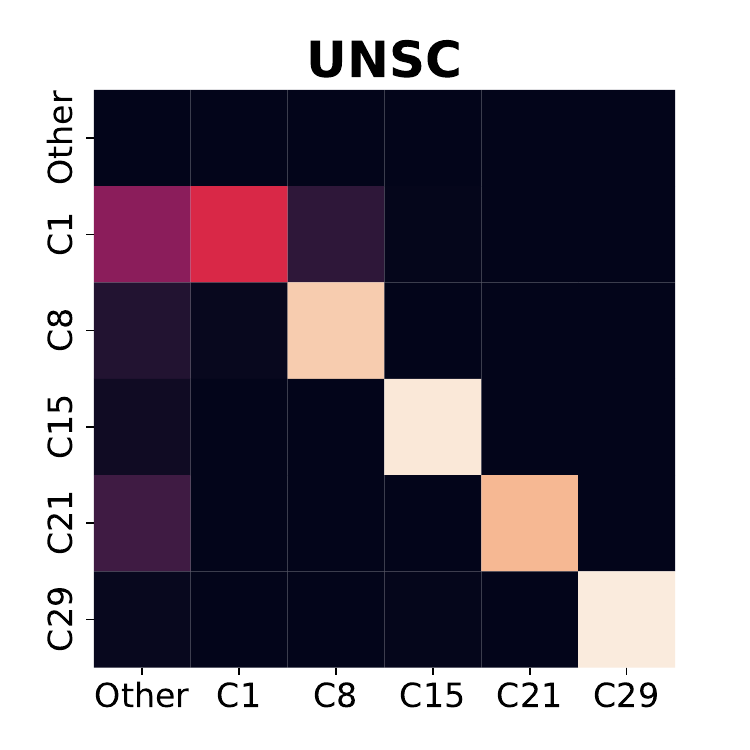}
        \end{minipage}
        }
        \caption{Confusion matrices with different configurations of MRA components. (a) demonstrates the confusion matrices after unlearning with different MU methods. Due to space limit, \textbf{LARGER figures} can be found in the online extended version.}
        \label{fig:ab_cmt}
\end{figure*}

\subsubsection{Comparison Results}

\Cref{tab:ablation} reports the results of MRA on Pet-37, where the \textit{Acc} on testing dataset $\mathcal{D}_{ts}$ and forgetting dataset $\mathcal{D}_f$ is reported. Additional results for ablation on other datasets can be found in the online extended version.

From the results, we can easily find that \textbf{DST} underperforms \textbf{DST+STU} and \textbf{DST+STU+TCH}. This is because the \textbf{DST} component only distills knowledge inferred from the prediction dataset $\mathcal{D}_p$ from the ULM to the student model, where the distilled knowledge inevitably contains label noise from the ULM. As a result, the \textit{Acc} of \textbf{DST} is close to that of \textbf{ULM}.
In comparison, \textbf{DST+STU} and \textbf{DST+STU+TCH} are based on the alternative learning process with both distillation and recall steps (cf. \Cref{alg:mra}). In the recall step, samples with high-confidence agreement are extracted, which can effectively recall forgotten class memberships. The model with all components, i.e., \textbf{DST+STU+TCH}, achieves the best performance because the additional \textbf{TCH} component can further recall the knowledge retained by the ULM, as illustrated in \Cref{sec:exp_mra_open_source}.

\subsubsection{Visualization of Confusion Matrices} 
\Cref{fig:ab_cmt} (a-d) visualize the normalized confusion matrices on $\mathcal{D}_f$ w.r.t. \textbf{ULM}, \textbf{DST}, \textbf{DST+STU} and \textbf{DST+STU+TCH}. From \Cref{fig:ab_cmt} (a) \textbf{ULM}, we can observe the highlights in the first column of each subfigure, which illustrates that MU methods have successfully unlearned true class memberships of forgetting instances.
From \Cref{fig:ab_cmt} (b) to (c), the diagonals of confusion matrices w.r.t. \textbf{DST}, \textbf{DST+STU} and \textbf{DST+STU+TCH} become more and more noticeable, that is, more and more forgotten class memberships have been successfully recalled, which shows that each component plays an important role in the MRA framework.

\section{Conclusion}
This study is the first attempt to explore MRA against MU techniques to recall unlearned class memberships, highlighting vulnerabilities of MU in data privacy protections. By using ULMs as noisy labelers, our implementation of MRA can recall forgotten class memberships from the ULMs without the need for the original model. Extensive experiments on four real-world datasets show that the proposed MRA framework exhibits high efficacy in recovering forgotten class memberships carried out by various MU methods. In particular, it reveals the phenomenon that \emph{``The MU methods in precise unlearning may lead to high success rate to recall the forgotten class memberships via MRA''}. Consequently, the proposed MRA can serve as a valuable tool to assess the potential risk of privacy leakage for existing and new MU methods, thus gaining deeper insights into MU.


\section*{Acknowledgments}
This work is partially supported by the National Natural Science Foundation of China (NSFC Granted No. 62276190).

\bibliographystyle{named}
\bibliography{ijcai25}
\clearpage
\appendix
\section*{Appendix}

\section{Source Code}
\label{sec:code}
To ensure a clear and concise code structure, we provide only the core implementation in the supplementary materials. Experimental datasets, trained model checkpoints, and other auxiliary files are excluded. The source code is available for review in the anonymous repository: \url{https://anonymous.4open.science/r/mra4mu-ijcai}.

\section{Additional Experiment Details}
This section provides additional information on \textbf{Data Preparation} and \textbf{Model Configuration} that are not specified in the main paper due to the limitation of space.

\subsection{Dataset Preparation}
For each dataset used in the experiments, five classes are selected to construct the forgetting dataset ${\mathcal{D}_f}$. 
\Cref{tab:forget_cls} lists the class ID and name for each forgetting class.

\begin{table*}[ht!]
	\centering
	\scalebox{1}{
		\begin{tabular}{l|l}
			\toprule
			\multicolumn{1}{c|}{\textbf{Dataset}} & \multicolumn{1}{c}{\textbf{ Forgetting Classes ID and Name}} \\
			\hline
			CIFAR-10 & C1: automobile; C3: cat; C5: dog; C7: horse; C9: truck \\
			CIFAR-100 & C10: bowl; C30: dolphin; C50: mouse; C70: rose; C90: train\\
			Pet-37 & C1: Bengal; C8: Ragdoll; C15: Basset Hound C21: Beagle; C29: Japanese Chin\\
			Flower-102 & C50: petunia; C72: water lily; C76: passion flower; C88: watercress; C93: foxglove; \\
			\bottomrule
		\end{tabular}%
	}
	\caption{Details of five forgetting classes for each dataset}
	\label{tab:forget_cls}%
\end{table*}%

\subsection{Model Configuration}
\Cref{tab:model_cfg_closesource,tab:model_cfg_opensource} list the detailed model configurations of MRA across all the experiments for the closed-source and open-source cases respectively. 
The MRA model configuration can be separated into three groups: \textbf{Hyperparameter} and \textbf{Model Optimization}.

For \textbf{Network Architecture}, we use EfficientNet (EFN, small version) \cite{efficientnet} on CIFAR datasets, ResNet (ResNet18) \cite{resnet18} on Pet-37, and Swin-Transformer V2
(Swin-T, tiny version) \cite{liu2022swin} on Flower-102 for a comprehensive study.
In particular, the ULM is based on Swin-T while the STM is based on ResNet for the Flower-102 dataset.

For \textbf{Hyperparameter}, it is further categorized into three subgroups,
\textbf{(1) Mixup}: $\alpha_{x}$ is the beta distribution hyperparameter for Mixup on images (cf. Eq. (1) in the main paper), and $\alpha_{y}$ is the hyperparameter of beta distribution for the Mixup on labels (cf. Eq. (2) in the main paper); 
\textbf{(2) Smoothing}: $\gamma_l$ is the hyperparameter of Laplace smoothing (cf. Eq. (6) in the main paper), and $\gamma_s$ is the hyperparameter of label smoothing (cf. Eq. (8) in the main paper);
\textbf{(3) Top-K}: $\tau$ is the hyperparameter to control how many confidence-agreement samples are selected (cf. Eq. (8) in the main paper).

For \textbf{Model Optimization}, we use AdamW as the optimizer across all experiments, $LR_S$ denotes the initial learning rate for STM, and $LR_U$ denotes the initial learning rate for ULM, and the batch size for each dataset is also reported. 

\begin{table*}[ht!]
	\centering
	\resizebox{\linewidth}{!}
	{
		\begin{tabular}{l|c|c|c|c|c|c|c|c|r|c|r}
			\toprule
			\multicolumn{1}{c|}{\multirow{3}[6]{*}{\textbf{Dataset}}} & \multicolumn{2}{c|}{\multirow{2}[4]{*}{\textbf{\shortstack{Network\\Architecture}}}} & \multicolumn{5}{c|}{\textbf{Hyperparameter}} & \multicolumn{4}{c}{\multirow{2}[4]{*}{\textbf{\shortstack{Model\\Optimization}}}} \\
			\cmidrule{4-8}          & \multicolumn{2}{c|}{} & \multicolumn{2}{c|}{\textbf{Mixup}} & \multicolumn{2}{c|}{\textbf{Smoothing}} & \textbf{Top-K} & \multicolumn{4}{c}{} \\
			\cmidrule{2-12}          & \textbf{ULM} & \textbf{STM} & $\alpha_x$ & $\alpha_y$ & $\gamma_l$ & $\gamma_s$ & $\tau$ & Optimizer & \multicolumn{1}{c|}{$LR_S$} & $LR_U$ & \multicolumn{1}{c}{Batch Size} \\
			\midrule
			CIFAR-10 & EFN   & EFN   & 0.20  & 0.75  & 0.01  & 0.05  & 0.60   & AdamW & 1.00E-04 & NA    & 256 \\
			CIFAR-100 & EFN   & EFN   & 0.20  & 0.75  & 0.01  & 0.05  & 0.60  & AdamW & 2.00E-04 & NA    & 256 \\
			Pet-37 & ResNet & ResNet & 0.20  & 0.75  & 0.01  & 0.05  & 0.05  & AdamW & 2.00E-04 & NA    & 64 \\
			Flower-102 & Swin-T & ResNet & 0.20  & 0.75  & 0.01  & 0.05  & 0.05  & AdamW & 2.00E-04 & NA    & 32 \\
			\bottomrule
		\end{tabular}%
	}
	\caption{Detailed model configuration for MRA in the closed-source case (ULM is not updated so $LR_U$ is NA)}
	\label{tab:model_cfg_closesource}%
\end{table*}%

\begin{table*}[ht!]
	\centering
	\resizebox{\linewidth}{!}
	{
		\begin{tabular}{l|c|c|c|c|c|c|c|c|r|c|r}
			\toprule
			\multicolumn{1}{c|}{\multirow{3}[6]{*}{\textbf{Dataset}}} & \multicolumn{2}{c|}{\multirow{2}[4]{*}{\textbf{\shortstack{Network\\Architecture}}}} & \multicolumn{5}{c|}{\textbf{Hyperparameter}} & \multicolumn{4}{c}{\multirow{2}[4]{*}{\textbf{\shortstack{Model\\Optimization}}}} \\
			\cmidrule{4-8}          & \multicolumn{2}{c|}{} & \multicolumn{2}{c|}{\textbf{Mixup}} & \multicolumn{2}{c|}{\textbf{Smoothing}} & \textbf{Top-K} & \multicolumn{4}{c}{} \\
			\cmidrule{2-12}          & \textbf{ULM} & \textbf{STM} & $\alpha_x$ & $\alpha_y$ & $\gamma_l$ & $\gamma_s$ & $\tau$ & Optimizer & \multicolumn{1}{c|}{$LR_S$} & $LR_U$ & \multicolumn{1}{c}{Batch Size} \\
			\midrule
			CIFAR-10 & EFN   & EFN   & 0.20  & 0.75  & 0.01  & 0.05  & 0.05   & AdamW & 1.00E-04 & 5.00E-05    & 256 \\
			CIFAR-100 & EFN   & EFN   & 0.20  & 0.75  & 0.01  & 0.05  & 0.05   & AdamW & 2.00E-04 & 1.00E-04    & 256 \\
			Pet-37 & ResNet & ResNet & 0.20  & 0.75  & 0.01  & 0.05  & 0.05  & AdamW & 2.00E-04 & 1.00E-04    & 64 \\
			Flower-102 & Swin-T & ResNet & 0.20  & 0.75  & 0.01  & 0.05  & 0.05  & AdamW & 1.00E-04 & 1.00E-04  & 32 \\
			\bottomrule
		\end{tabular}%
	}
	\caption{Detailed model configuration for MRA in the open-source case (ULM is optimized with $LR_U$)}
	\label{tab:model_cfg_opensource}%
\end{table*}%

\section{Additional Results of MRA Efficacy}

To intuitively demonstrate the capacity of MRA to recall the forgotten class memberships on the prediction images $\mathcal{X}_f$, we conducted in-detail evaluations w.r.t. each forgetting class by comparing the \textit{Acc} between ULM and RCM after MRA on CIFAR-10, CIFAR-10, Pet-37 and Flower-102. 

From the illustrations in both the closed-source and the open-source case, we can clearly observe the phenomenon that \emph{``An MU method with more precise unlearning may lead to higher success rate of MRA to recall the forgotten class memberships''}, e.g., SalUn and UNSC. As a result, MRA can serve as a valuable tool to assess the potential risk of privacy leakage for MU methods, thus facilitating the development of more robust MU models.

\subsection{Class-specific Recovery Efficacy (Close-source Case)}

\begin{figure*}[ht!]
	\begin{minipage}[c]{\linewidth}
		\flushleft
		\includegraphics[width=0.24\linewidth]{figures/results/cifar-10/fr_0.5_ijcai/fisher_efficientnet_s_student_only_bar.pdf}
		\includegraphics[width=0.24\linewidth]{figures/results/cifar-10/fr_0.5_ijcai/RL_efficientnet_s_student_only_bar.pdf}
		\includegraphics[width=0.24\linewidth]{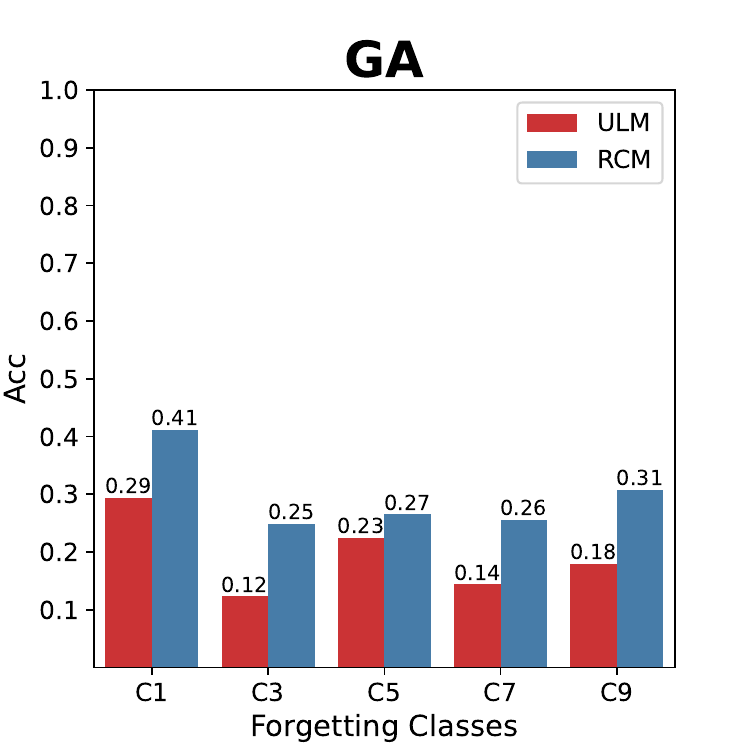}
		\includegraphics[width=0.24\linewidth]{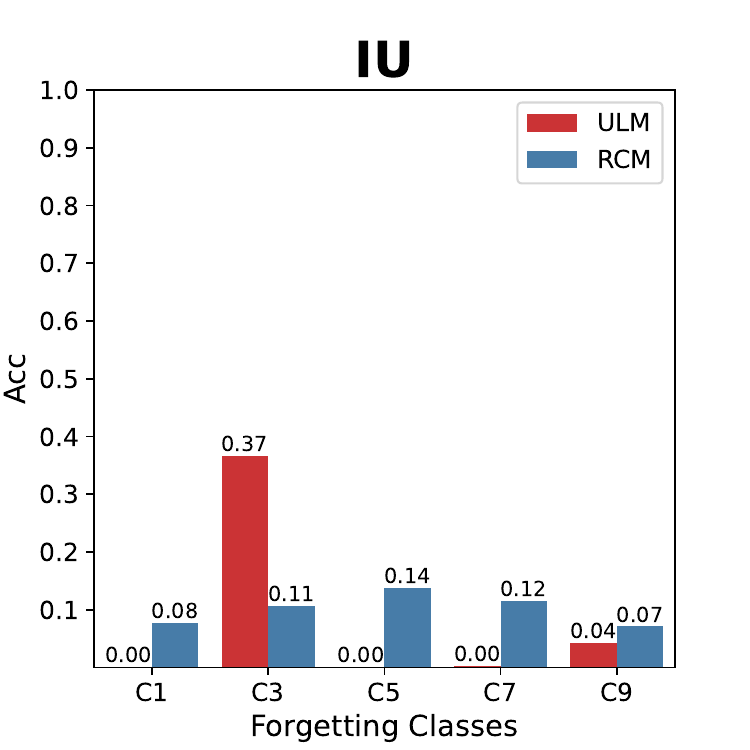}
		\includegraphics[width=0.24\linewidth]{figures/results/cifar-10/fr_0.5_ijcai/BU_efficientnet_s_student_only_bar.pdf}
		\includegraphics[width=0.24\linewidth]{figures/results/cifar-10/fr_0.5_ijcai/GA_l1_efficientnet_s_student_only_bar.pdf}
		\includegraphics[width=0.24\linewidth]{figures/results/cifar-10/fr_0.5_ijcai/SalUn_efficientnet_s_student_only_bar.pdf}
		\includegraphics[width=0.24\linewidth]{figures/results/cifar-10/fr_0.5_ijcai/UNSC_efficientnet_s_student_only_bar.pdf}
		\caption{Comparison of the \textit{Acc} between ULM and RCM (\textbf{closed-source case}) w.r.t. each forgetting class on CIFAR-10 dataset.}
	\end{minipage}
	\label{fig:closed_source_cifar10}
\end{figure*}

\begin{figure*}[ht!]
	\begin{minipage}[c]{\linewidth}
		\flushleft
		\includegraphics[width=0.24\linewidth]{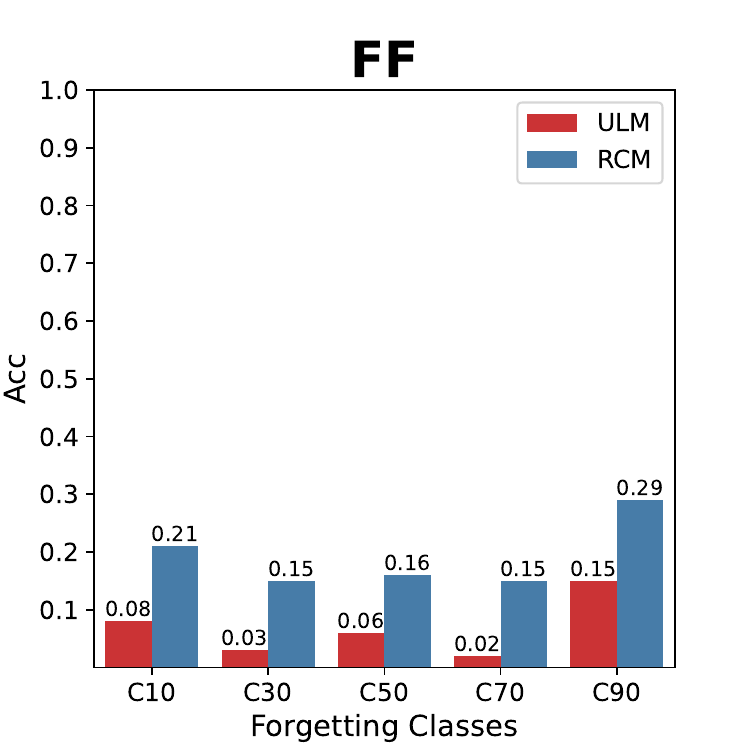}
		\includegraphics[width=0.24\linewidth]{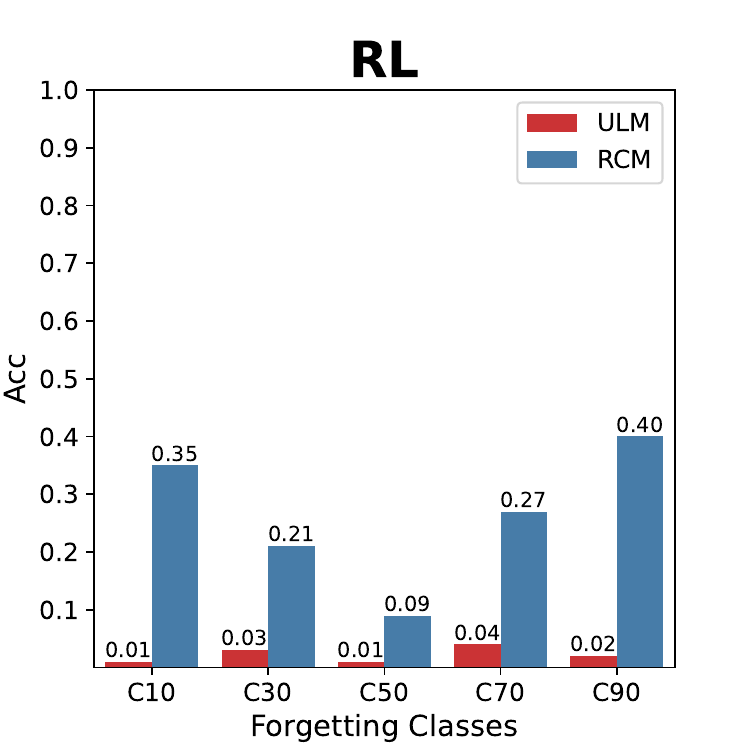}
		\includegraphics[width=0.24\linewidth]{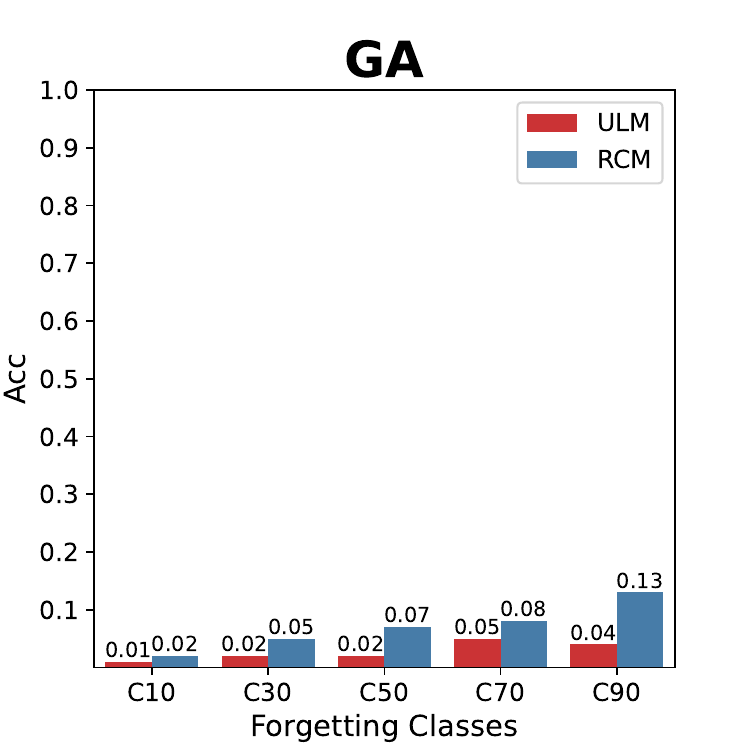}
		\includegraphics[width=0.24\linewidth]{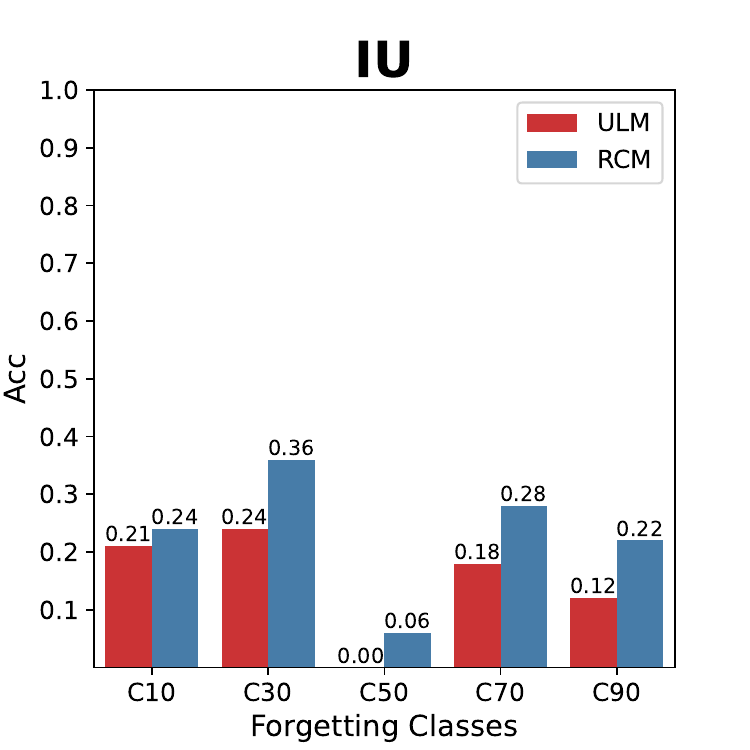}
		\includegraphics[width=0.24\linewidth]{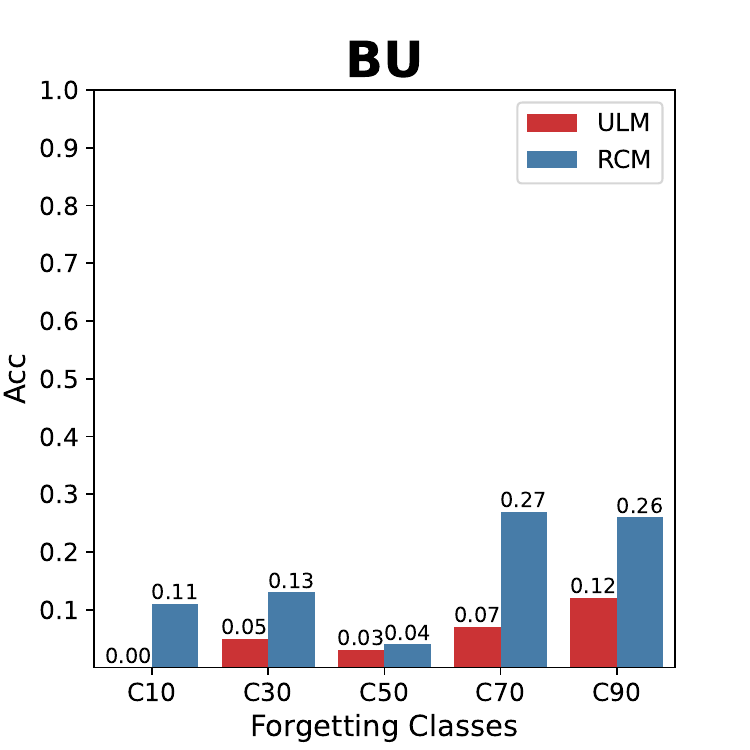}
		\includegraphics[width=0.24\linewidth]{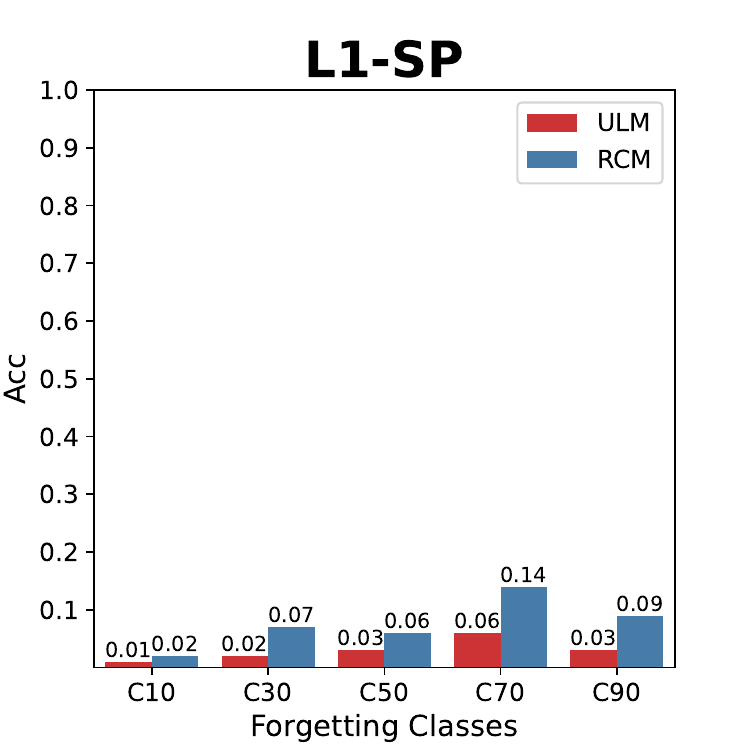}
		\includegraphics[width=0.24\linewidth]{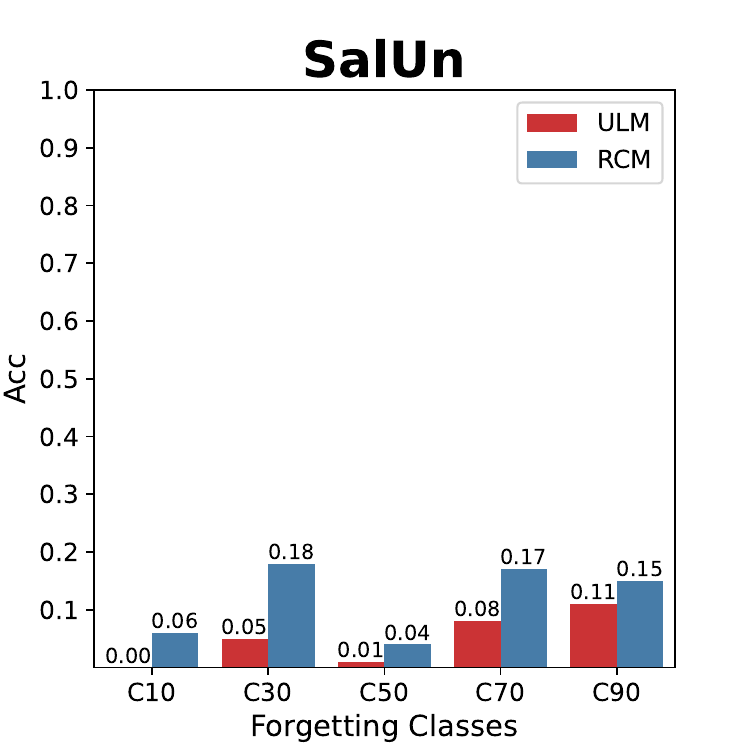}
		\includegraphics[width=0.24\linewidth]{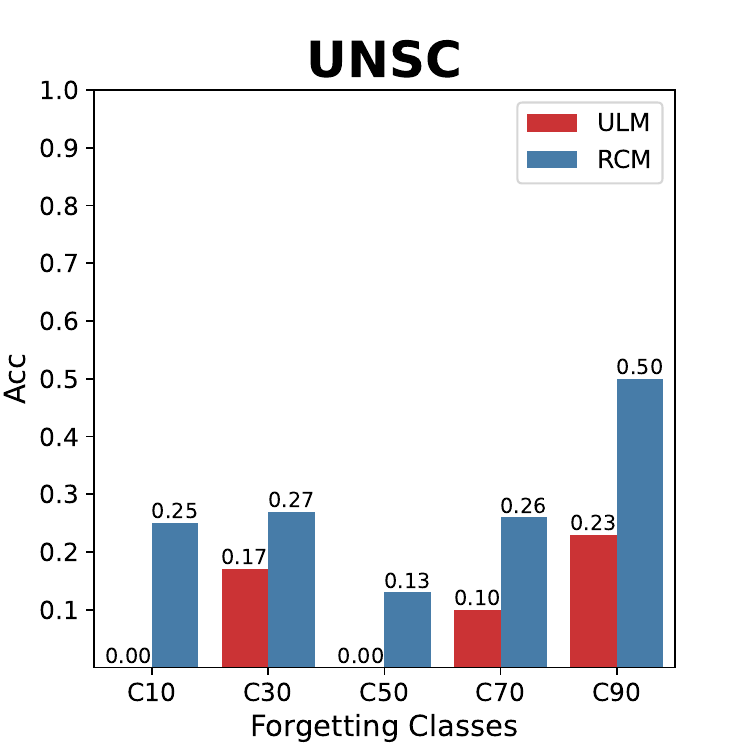}
		\caption{Comparison of the \textit{Acc} between ULM and RCM (\textbf{closed-source case}) w.r.t. each forgetting class on CIFAR-100 dataset.}
	\end{minipage}
	\label{fig:open_source_pet37}
\end{figure*}

\begin{figure*}[ht!]
	\begin{minipage}[c]{\linewidth}
		\flushleft
		\includegraphics[width=0.24\linewidth]{figures/results/pet-37/fr_0.5_ijcai/fisher_resnet18_student_only_bar.pdf}
		\includegraphics[width=0.24\linewidth]{figures/results/pet-37/fr_0.5_ijcai/RL_resnet18_student_only_bar.pdf}
		\includegraphics[width=0.24\linewidth]{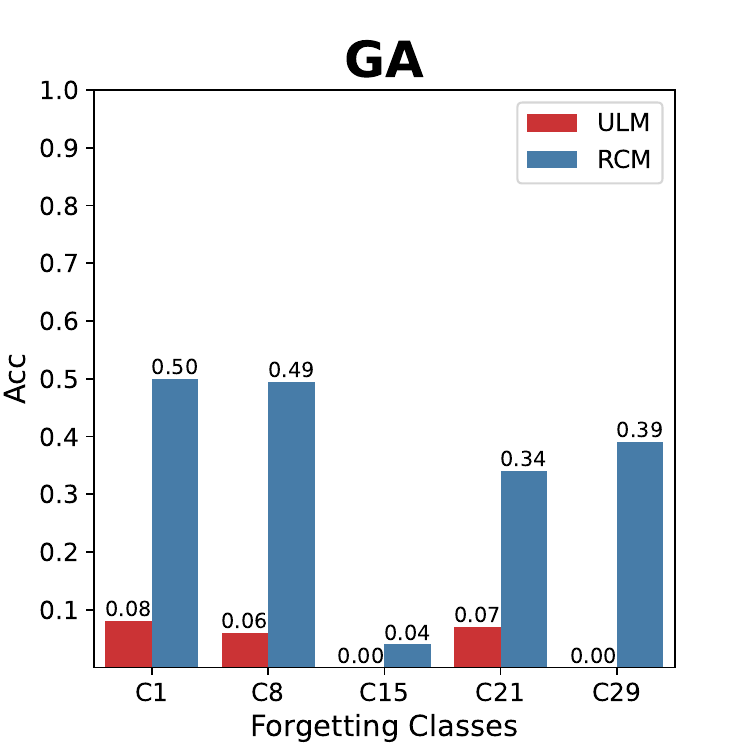}
		\includegraphics[width=0.24\linewidth]{figures/results/pet-37/fr_0.5_ijcai/IU_resnet18_student_only_bar.pdf}
		\includegraphics[width=0.24\linewidth]{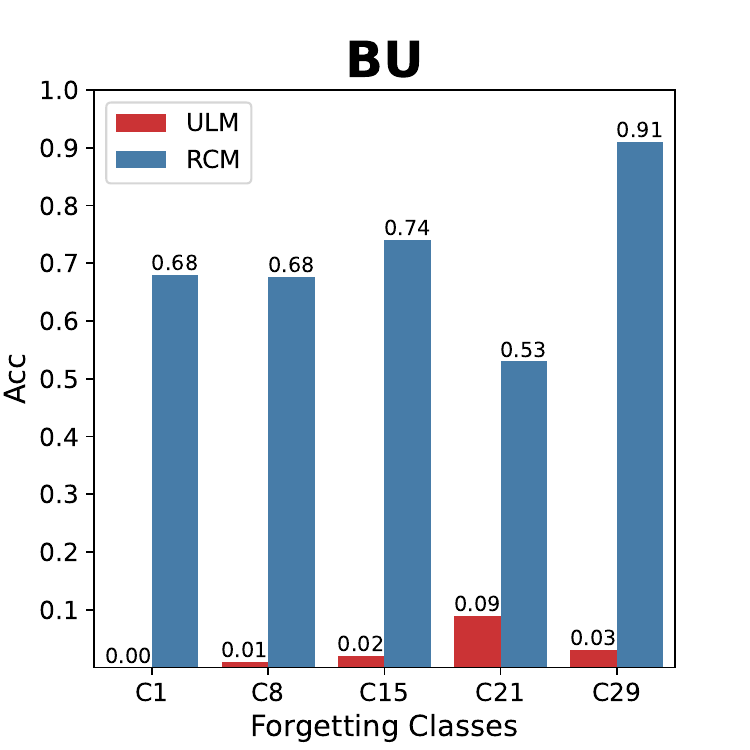}
		\includegraphics[width=0.24\linewidth]{figures/results/pet-37/fr_0.5_ijcai/GA_l1_resnet18_student_only_bar.pdf}
		\includegraphics[width=0.24\linewidth]{figures/results/pet-37/fr_0.5_ijcai/SalUn_resnet18_student_only_bar.pdf}
		\includegraphics[width=0.24\linewidth]{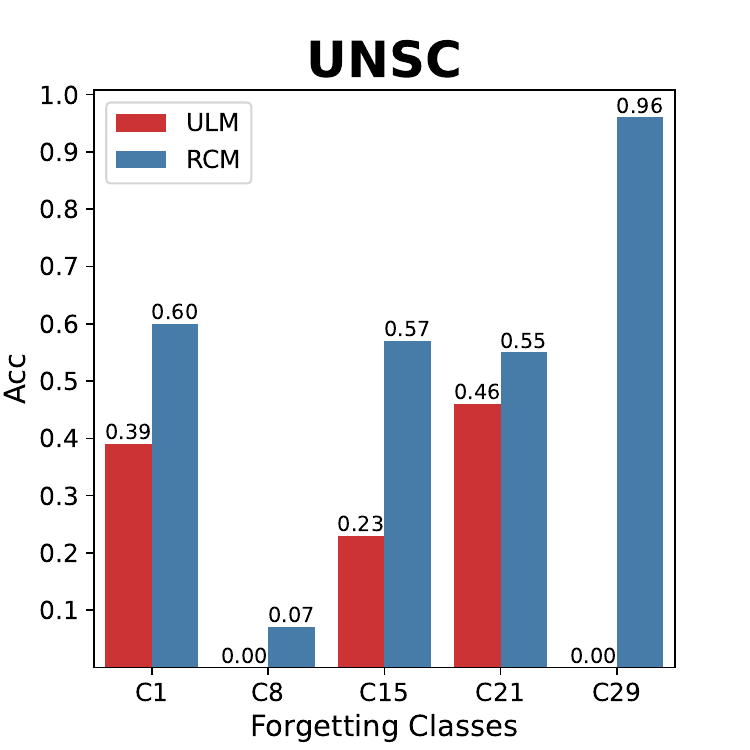}
		\caption{Comparison of the \textit{Acc} between ULM and RCM (\textbf{closed-source case}) w.r.t. each forgetting class on Pet-37 dataset.}
	\end{minipage}
	\label{fig:open_source_pet37}
\end{figure*}

\begin{figure*}[ht!]
	\begin{minipage}[c]{\linewidth}
		\flushleft
		\includegraphics[width=0.24\linewidth]{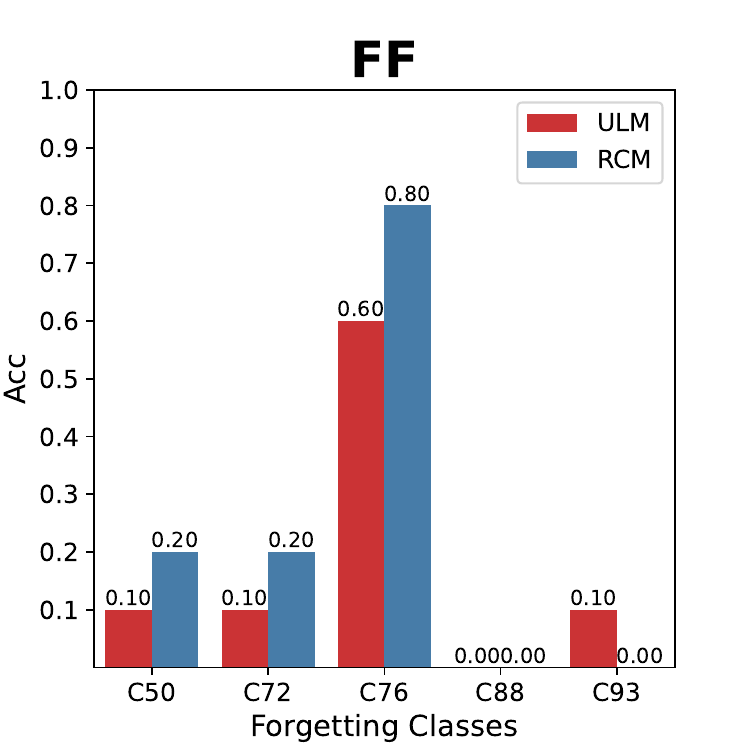}
		\includegraphics[width=0.24\linewidth]{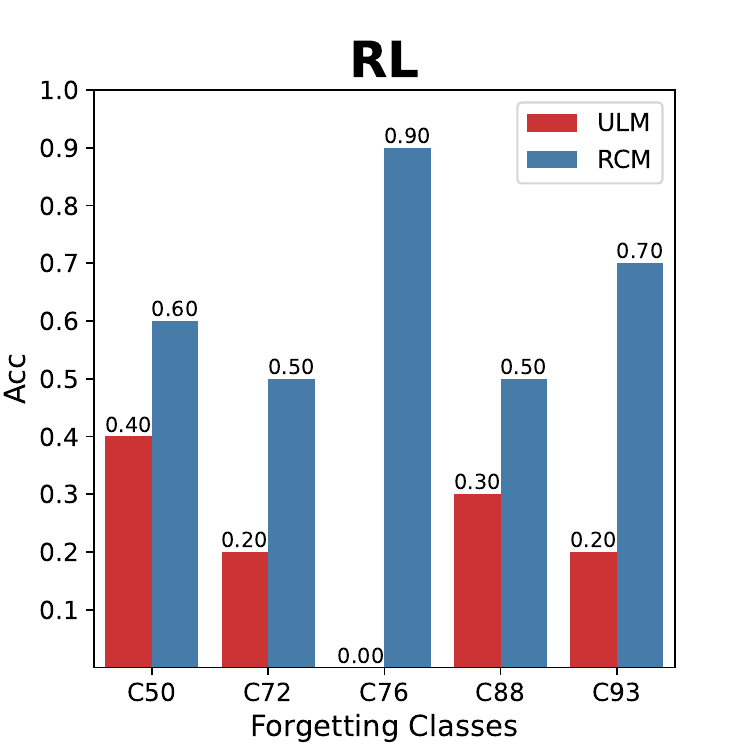}
		\includegraphics[width=0.24\linewidth]{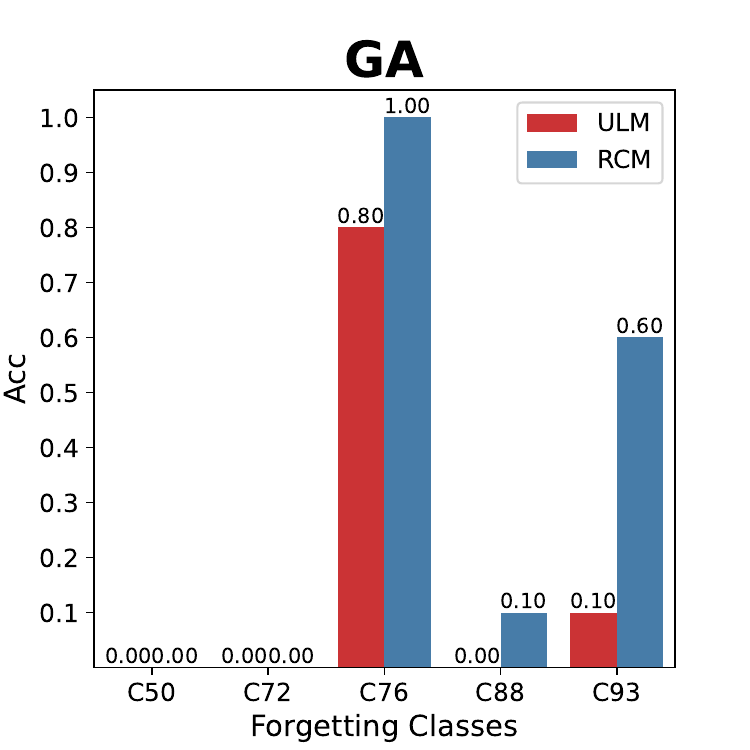}
		\includegraphics[width=0.24\linewidth]{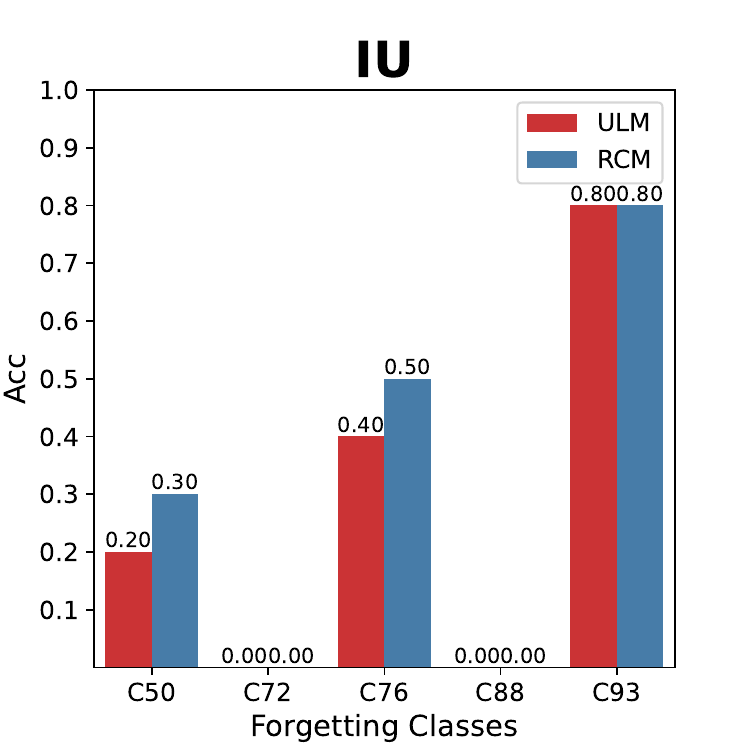}
		\includegraphics[width=0.24\linewidth]{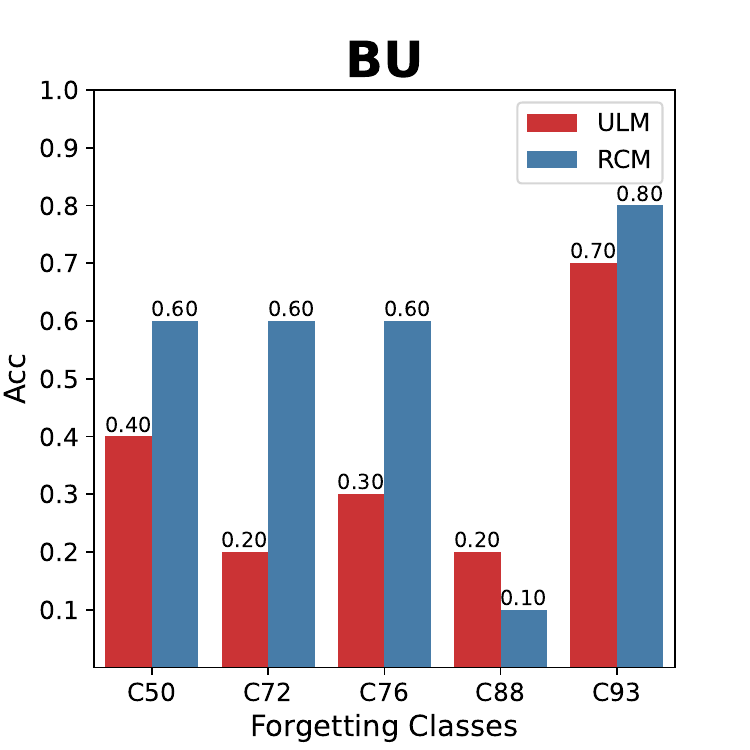}
		\includegraphics[width=0.24\linewidth]{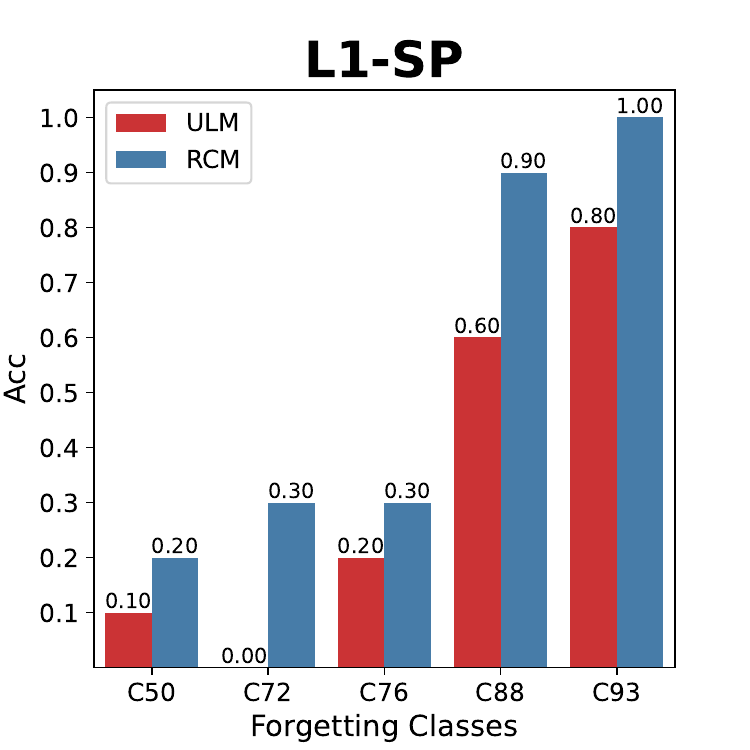}
		\includegraphics[width=0.24\linewidth]{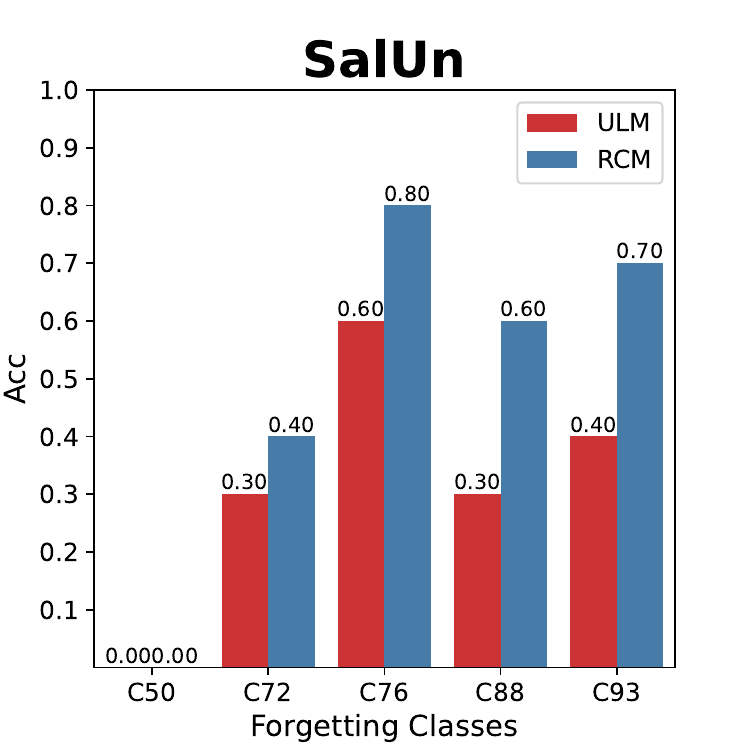}
		\caption{Comparison of the \textit{Acc} between ULM and RCM (\textbf{closed-source case}) w.r.t. each forgetting class on Flower-102 dataset.}
	\end{minipage}
	\label{fig:open_source_pet37}
\end{figure*}

\subsection{Class-specific Recovery Efficacy (Open-source Case)}

\begin{figure*}[ht!]
	\begin{minipage}[c]{\linewidth}
		\flushleft
		\includegraphics[width=0.24\linewidth]{figures/results/cifar-10/fr_0.5_ijcai/fisher_efficientnet_s_restore_bar.pdf}
		\includegraphics[width=0.24\linewidth]{figures/results/cifar-10/fr_0.5_ijcai/RL_efficientnet_s_restore_bar.pdf}
		\includegraphics[width=0.24\linewidth]{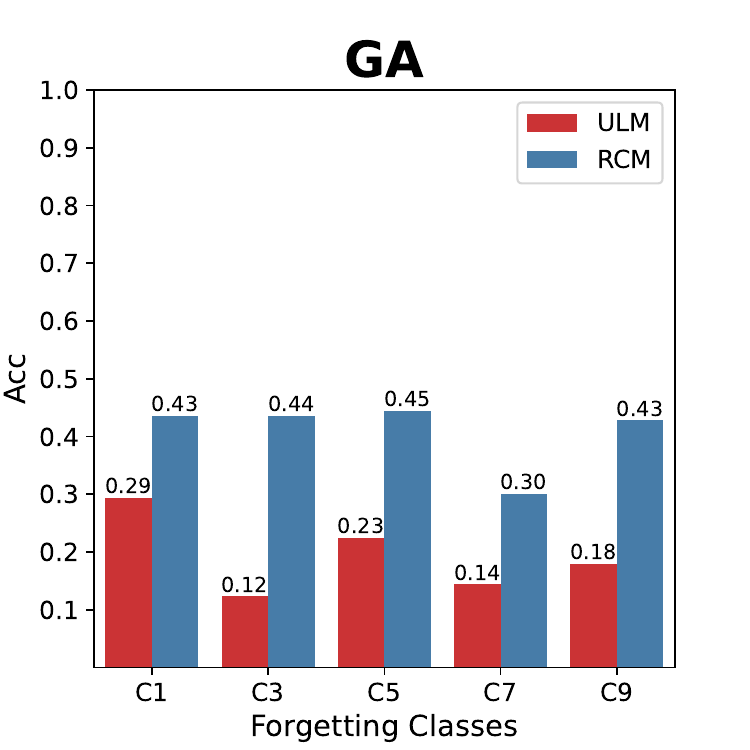}
		\includegraphics[width=0.24\linewidth]{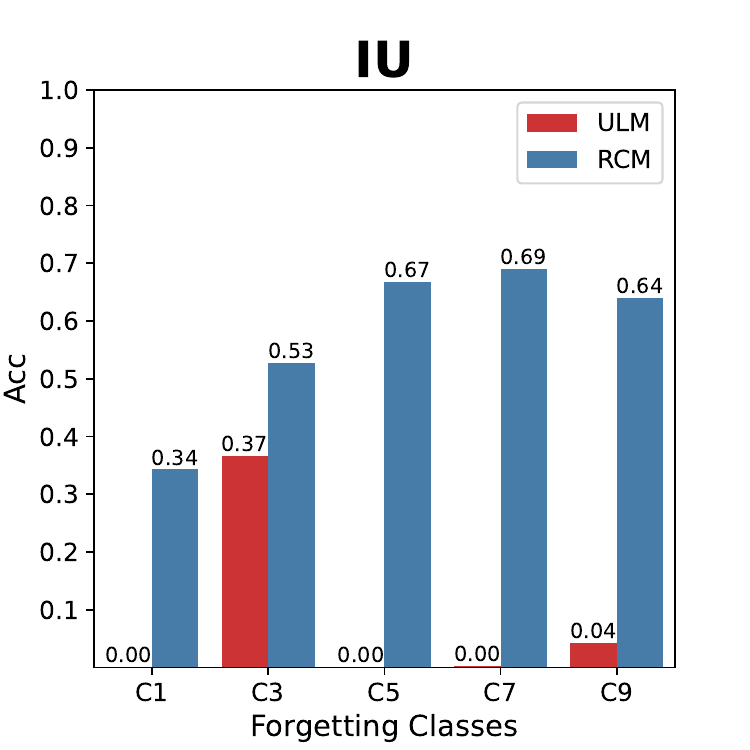}
		\includegraphics[width=0.24\linewidth]{figures/results/cifar-10/fr_0.5_ijcai/BU_efficientnet_s_restore_bar.pdf}
		\includegraphics[width=0.24\linewidth]{figures/results/cifar-10/fr_0.5_ijcai/GA_l1_efficientnet_s_restore_bar.pdf}
		\includegraphics[width=0.24\linewidth]{figures/results/cifar-10/fr_0.5_ijcai/SalUn_efficientnet_s_restore_bar.pdf}
		\includegraphics[width=0.24\linewidth]{figures/results/cifar-10/fr_0.5_ijcai/UNSC_efficientnet_s_restore_bar.pdf}
		\caption{Comparison of the \textit{Acc} between ULM and RCM (\textbf{open-source case}) w.r.t. each forgetting class on CIFAR-10 dataset.}
	\end{minipage}
	\label{fig:open_source_pet37}
\end{figure*}

\begin{figure*}[ht!]
	\begin{minipage}[c]{\linewidth}
		\flushleft
		\includegraphics[width=0.24\linewidth]{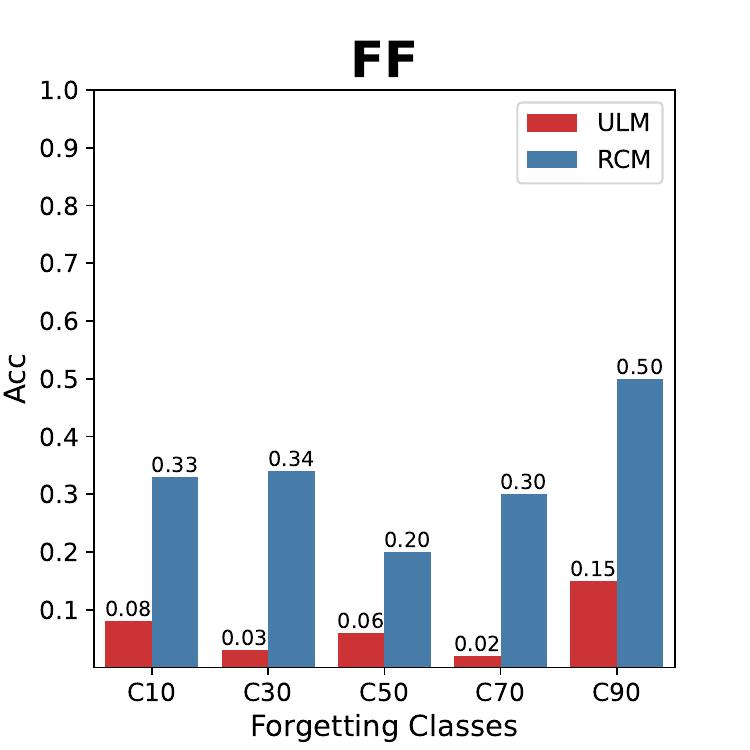}
		\includegraphics[width=0.24\linewidth]{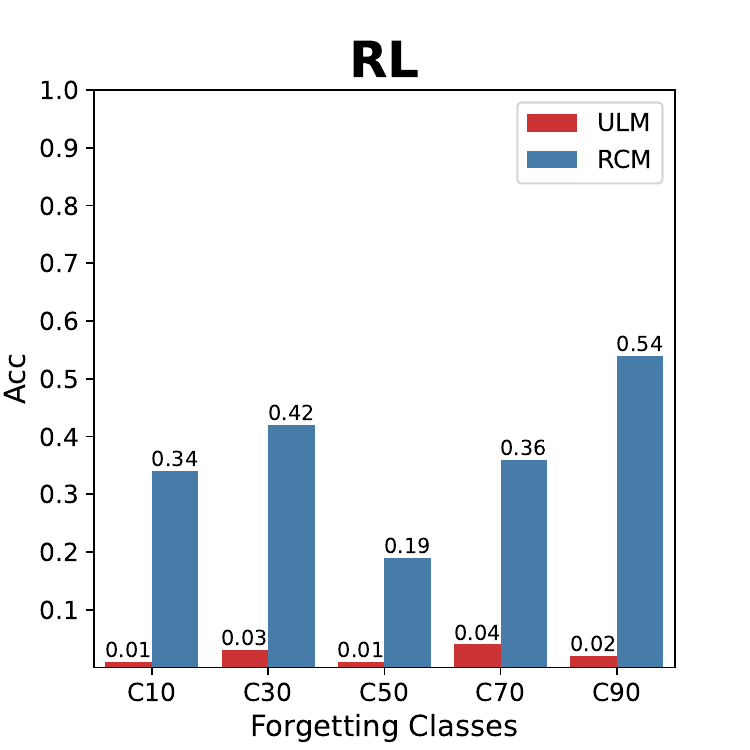}
		\includegraphics[width=0.24\linewidth]{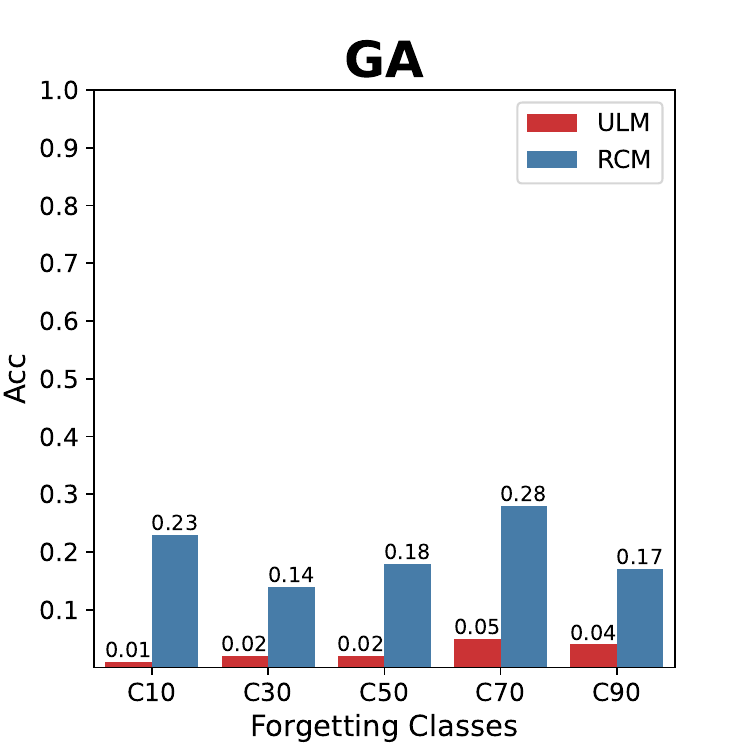}
		\includegraphics[width=0.24\linewidth]{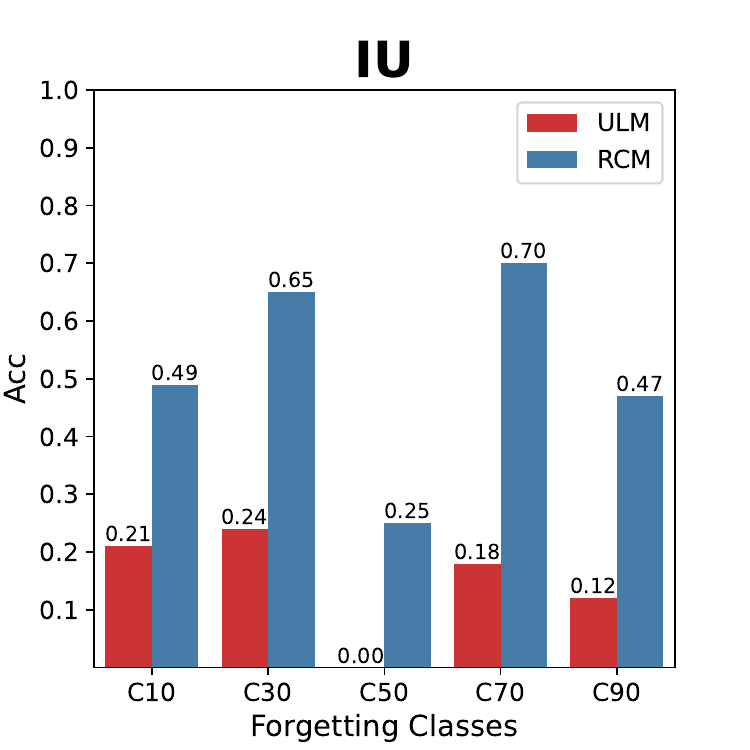}
		\includegraphics[width=0.24\linewidth]{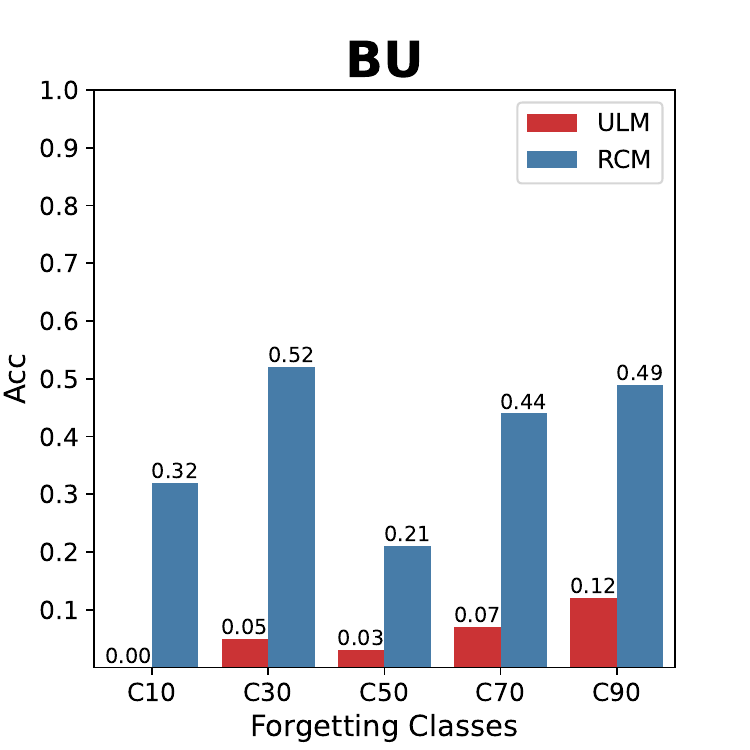}
		\includegraphics[width=0.24\linewidth]{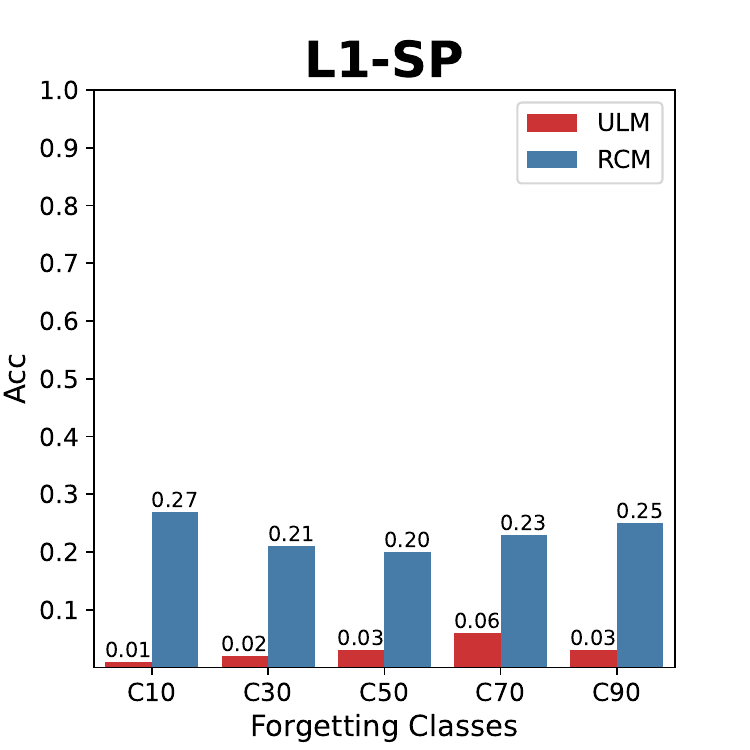}
		\includegraphics[width=0.24\linewidth]{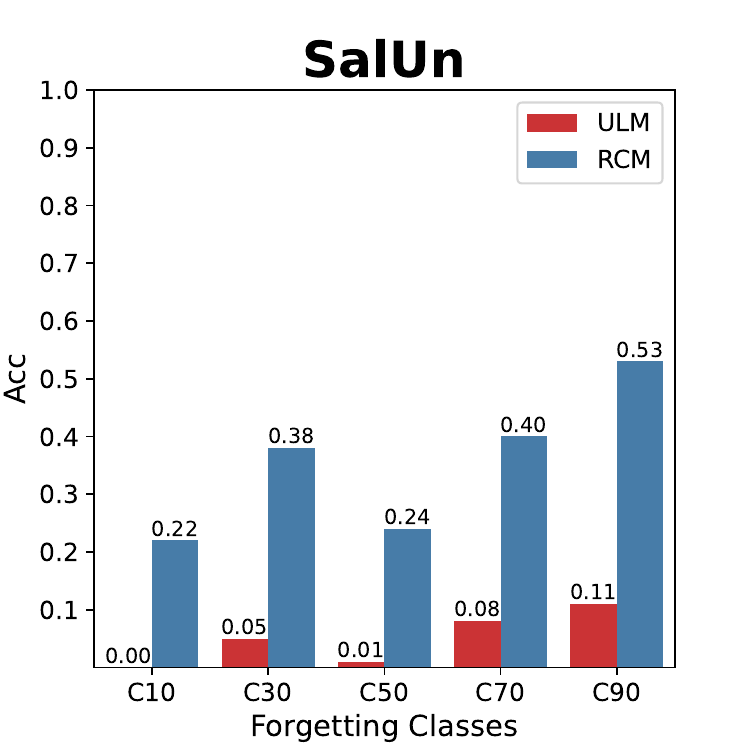}
		\includegraphics[width=0.24\linewidth]{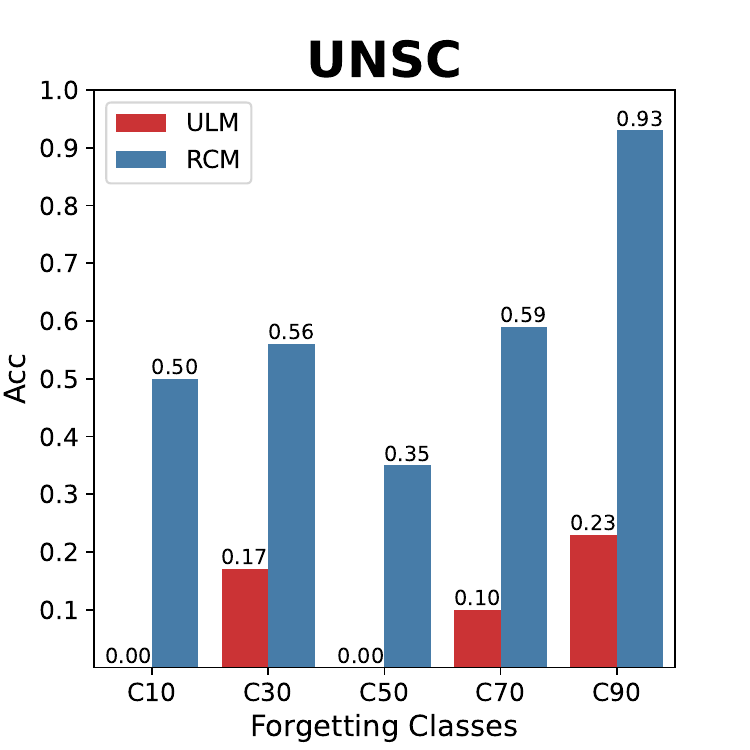}
		\caption{Comparison of the \textit{Acc} between ULM and RCM (\textbf{open-source case}) w.r.t. each forgetting class on CIFAR-100 dataset.}
	\end{minipage}
	\label{fig:open_source_pet37}
\end{figure*}

\begin{figure*}[ht!]
	\begin{minipage}[c]{\linewidth}
		\flushleft
		\includegraphics[width=0.24\linewidth]{figures/results/pet-37/fr_0.5_ijcai/fisher_resnet18_restore_bar.pdf}
		\includegraphics[width=0.24\linewidth]{figures/results/pet-37/fr_0.5_ijcai/RL_resnet18_restore_bar.pdf}
		\includegraphics[width=0.24\linewidth]{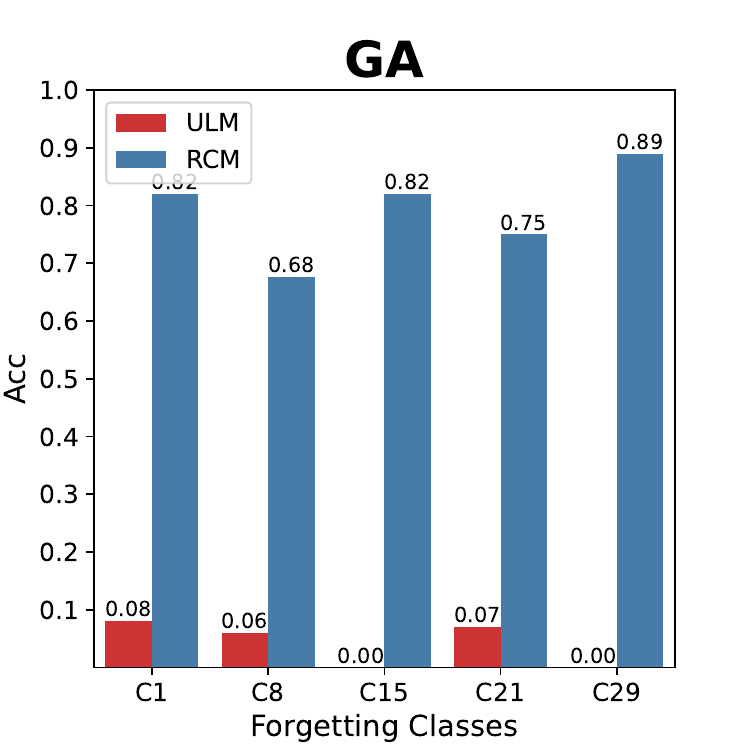}
		\includegraphics[width=0.24\linewidth]{figures/results/pet-37/fr_0.5_ijcai/IU_resnet18_restore_bar.pdf}
		\includegraphics[width=0.24\linewidth]{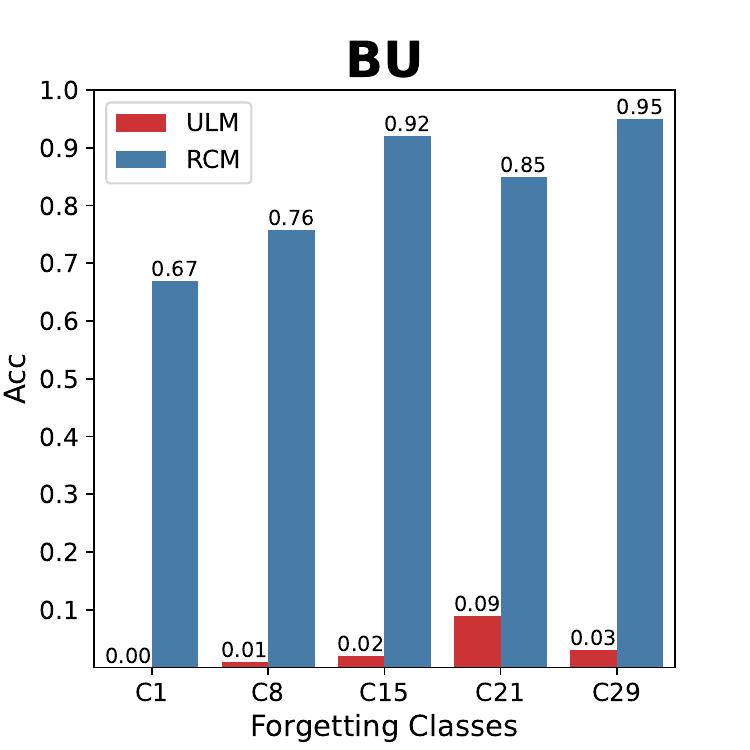}
		\includegraphics[width=0.24\linewidth]{figures/results/pet-37/fr_0.5_ijcai/GA_l1_resnet18_restore_bar.pdf}
		\includegraphics[width=0.24\linewidth]{figures/results/pet-37/fr_0.5_ijcai/SalUn_resnet18_restore_bar.pdf}
		\includegraphics[width=0.24\linewidth]{figures/results/pet-37/fr_0.5_ijcai/UNSC_resnet18_restore_bar.pdf}
		\caption{Comparison of the \textit{Acc} between ULM and RCM (\textbf{open-source case}) w.r.t. each forgetting class on Pet-37 dataset.}
	\end{minipage}
	\label{fig:open_source_pet37}
\end{figure*}

\begin{figure*}[ht!]
	\begin{minipage}[c]{\linewidth}
		\flushleft
		\includegraphics[width=0.24\linewidth]{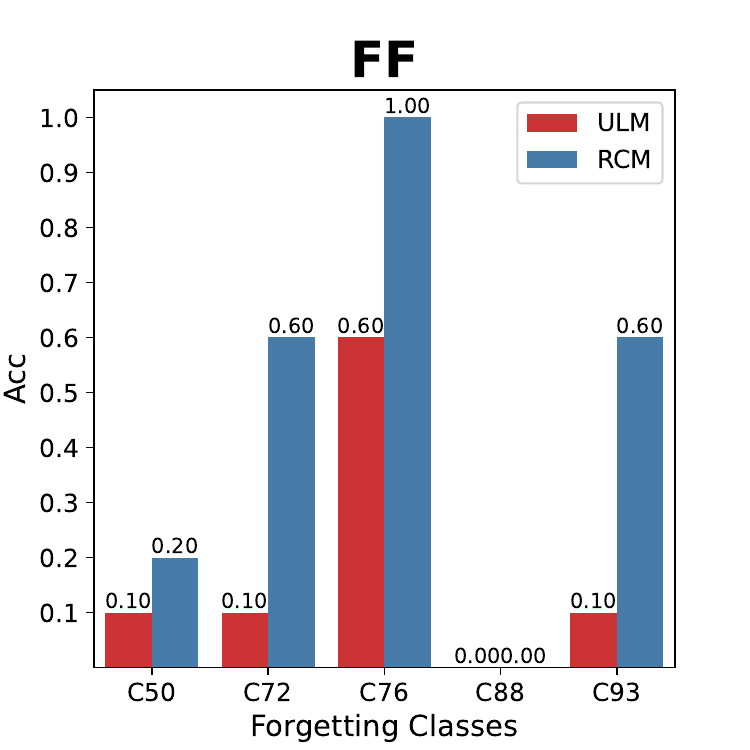}
		\includegraphics[width=0.24\linewidth]{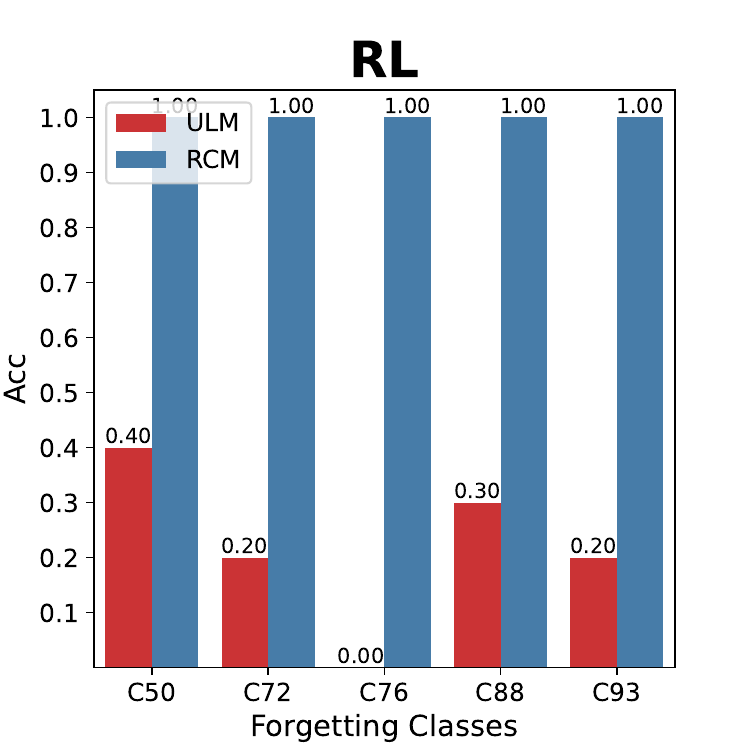}
		\includegraphics[width=0.24\linewidth]{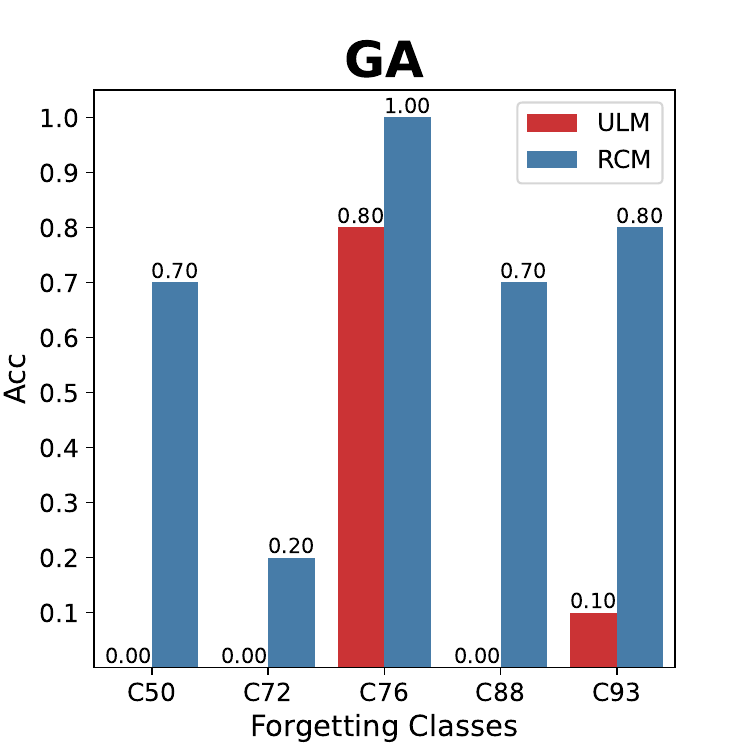}
		\includegraphics[width=0.24\linewidth]{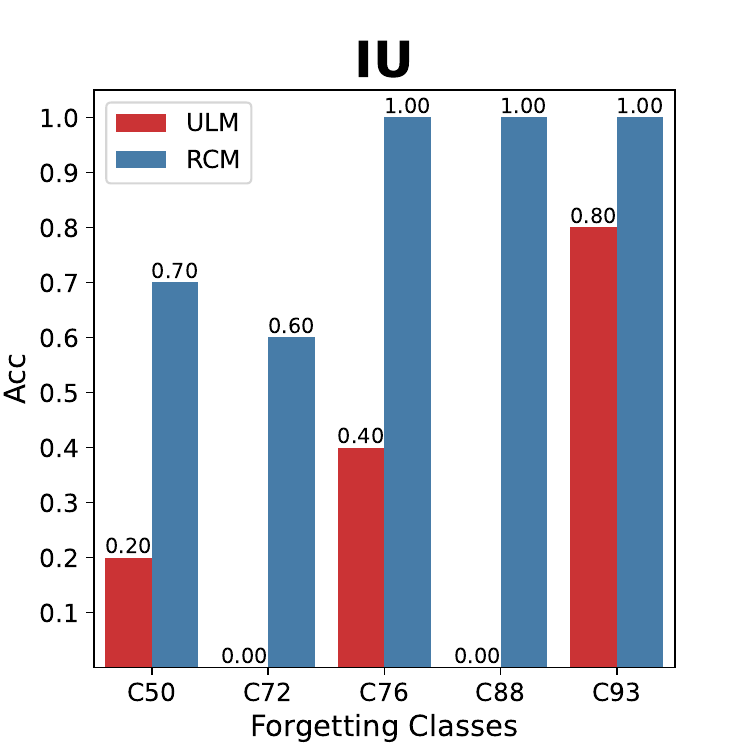}
		\includegraphics[width=0.24\linewidth]{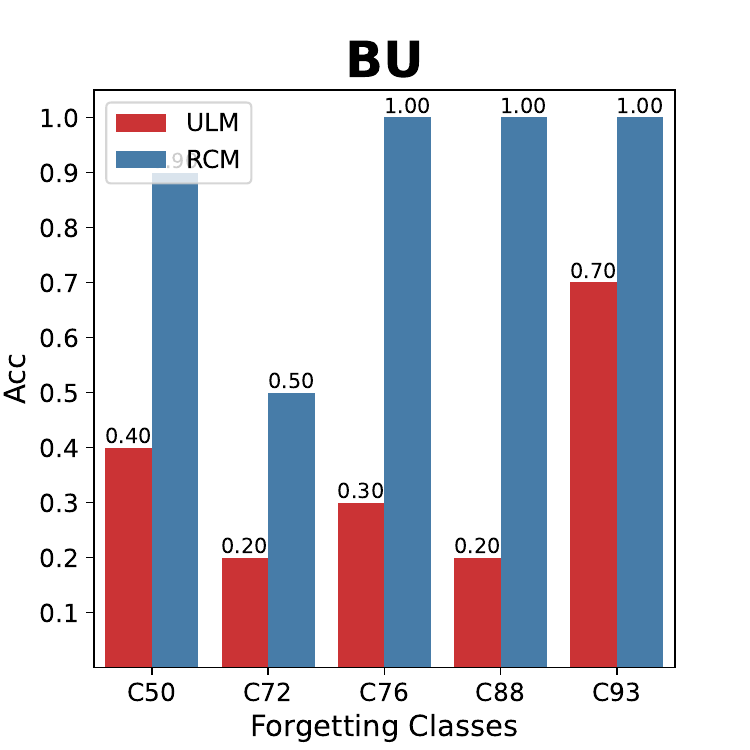}
		\includegraphics[width=0.24\linewidth]{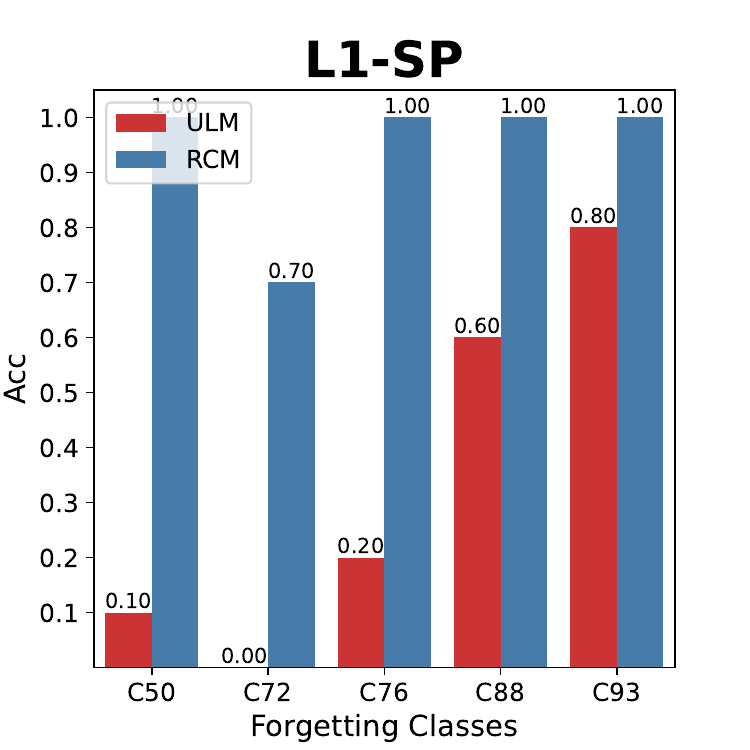}
		\includegraphics[width=0.24\linewidth]{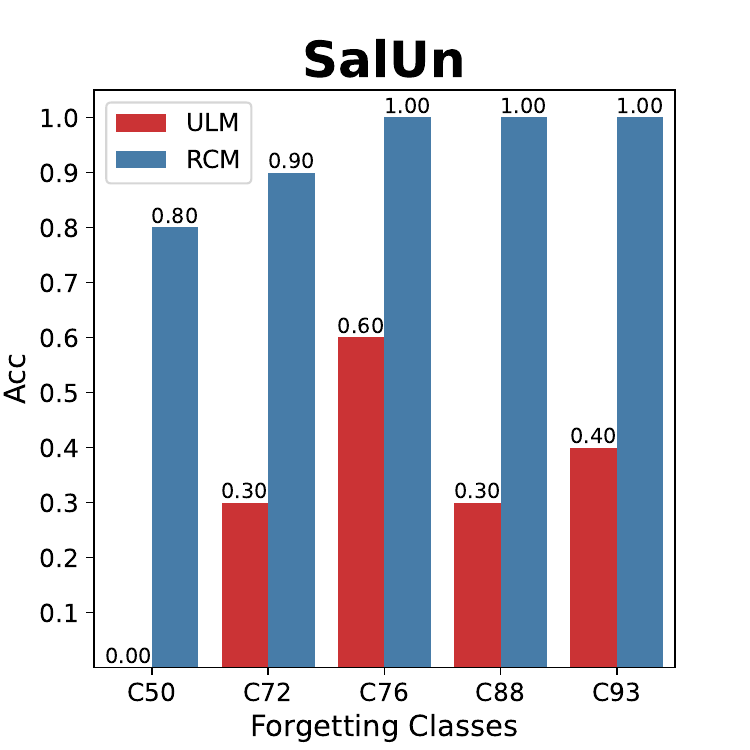}
		\caption{Comparison of the \textit{Acc} between ULM and RCM (\textbf{open-source case}) w.r.t. each forgetting class on Flower-102 dataset.}
	\end{minipage}
	\label{fig:open_source_pet37}
\end{figure*}

\section{Additional Results of Ablation Study}

In this section, we demonstrate more results of the ablation study on CIFAR-10, CIFAR-100, Pet-37, and Flower-102. The ablation components have been presented in the main paper, that is:

\begin{itemize}
	\item \textbf{DST:} This component is the \textit{Denosing Knowledge Distillation} step presented in the main paper, which aims to distill knowledge from the teacher model to the student model.
	
	\item \textbf{STU:} The component serves as the class membership recall process for STM, as presented in the \textit{Confident Membership Recall} step. That is, \textbf{DST+STU} is equivalent to the closed-source case of MRA.
	
	\item \textbf{TCH:} The component serves as the recall of class membership for teacher models as presented in the \textit{Confident Membership Recall} step. That is, \textbf{DST+STU+TCH} is equivalent to the open-source case of MRA.
\end{itemize}

\subsection{Ablation Study on CIFAR-10}

\begin{table*}[ht!]
	\centering
	\resizebox{\linewidth}{!}
	{
		\begin{tabular}{c|c|c||c|c|c|c|c|c|c|c|c|c|c|c|c|c|c|c}
			\toprule
			\multicolumn{3}{c||}{\textbf{Component}} & \multicolumn{2}{c|}{\textbf{FF}} & \multicolumn{2}{c|}{\textbf{RL}} & \multicolumn{2}{c|}{\textbf{GA}} & \multicolumn{2}{c|}{\textbf{IU}} & \multicolumn{2}{c|}{\textbf{BU}} & \multicolumn{2}{c|}{\textbf{L1-SP}} & \multicolumn{2}{c|}{\textbf{SalUn}} & \multicolumn{2}{c}{\textbf{UNSC}} \\
			\midrule
			\textbf{DST} & \textbf{STU} & \textbf{TCH} & $\mathcal{D}_{ts}$ & $\mathcal{D}_f$  & $\mathcal{D}_{ts}$ & $\mathcal{D}_f$  & $\mathcal{D}_{ts}$ & $\mathcal{D}_f$  & $\mathcal{D}_{ts}$ & $\mathcal{D}_f$  & $\mathcal{D}_{ts}$ & $\mathcal{D}_f$  & $\mathcal{D}_{ts}$ & $\mathcal{D}_f$  & $\mathcal{D}_{ts}$ & $\mathcal{D}_f$  & $\mathcal{D}_{ts}$ & $\mathcal{D}_f$ \\
			\midrule
			\checkmark &       &       & 0.163  & 0.126  & 0.393  & 0.144  & 0.501  & 0.252  & 0.114  & 0.101  & 0.458  & 0.059  & 0.448  & 0.131  & 0.451  & 0.027  & 0.199  & 0.030  \\
			\midrule
			\checkmark & \checkmark &       & 0.252  & 0.195  & 0.479  & 0.328  & 0.528  & 0.358  & 0.115  & 0.096  & 0.651  & 0.507  & 0.456  & 0.231  & 0.697  & 0.615  & 0.568  & 0.513  \\
			\midrule
			\checkmark & \checkmark & \checkmark & 0.708  & 0.726  & 0.471  & 0.351  & 0.582  & 0.509  & 0.689  & 0.759  & 0.783  & 0.877  & 0.479  & 0.338  & 0.829  & 0.997  & 0.774  & 0.921  \\
			\bottomrule
		\end{tabular}%
	}
	\caption{Ablation results (\textit{Acc}) of MRA on CIFAR-10 dataset w.r.t. different MU methods}
	\label{tab:addlabel}%
\end{table*}%

\begin{figure*}[ht!]
	\begin{minipage}[c]{\linewidth}
		\flushleft
		\includegraphics[width=0.24\linewidth]{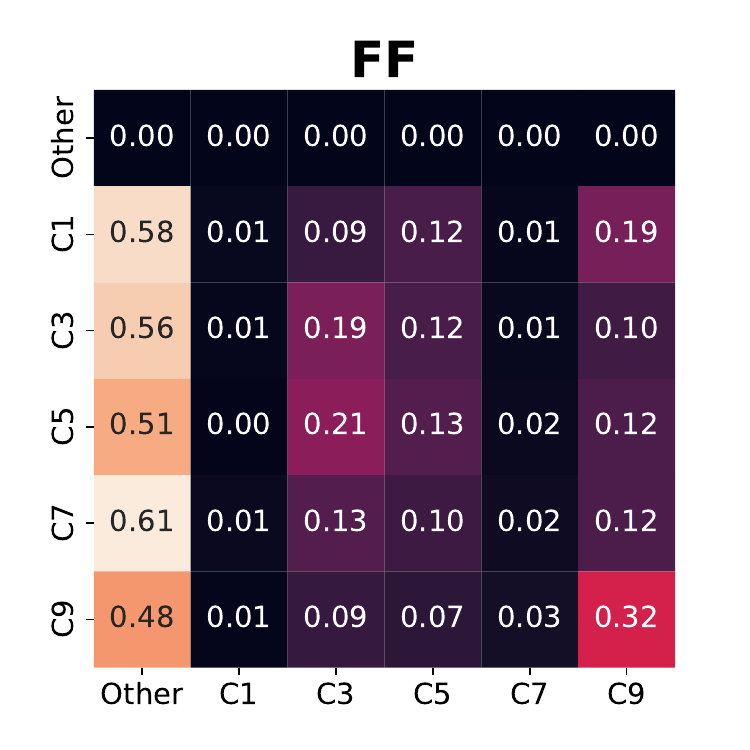}
		\includegraphics[width=0.24\linewidth]{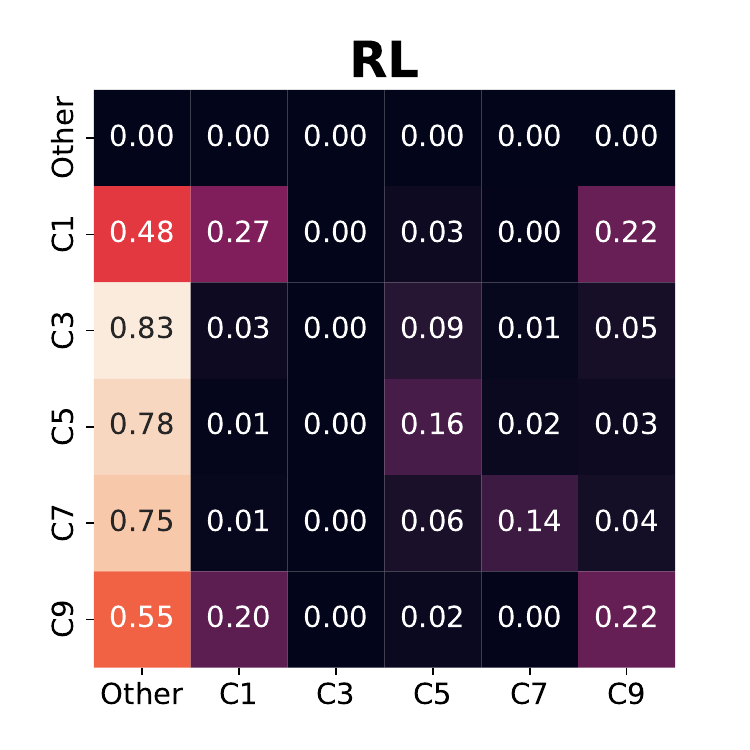}
		\includegraphics[width=0.24\linewidth]{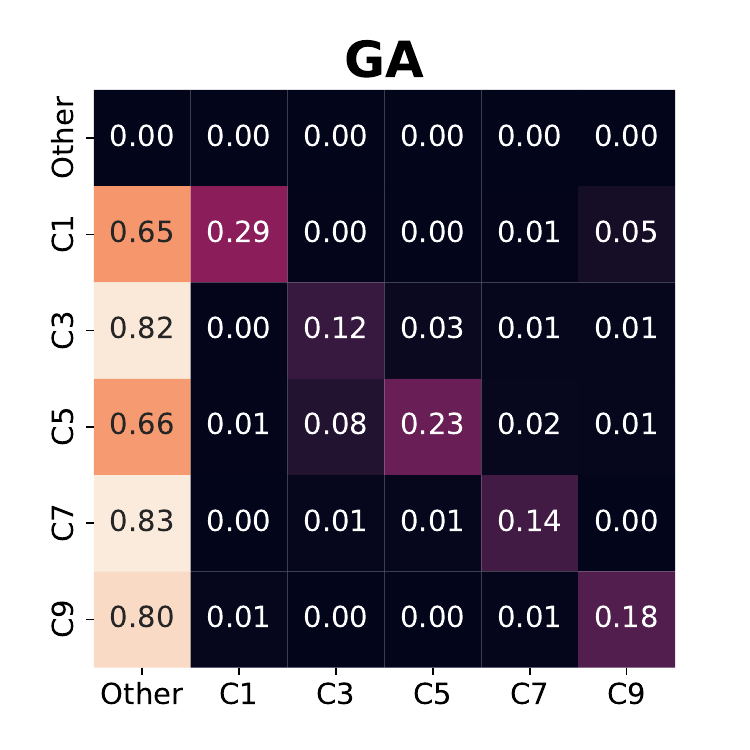}
		\includegraphics[width=0.24\linewidth]{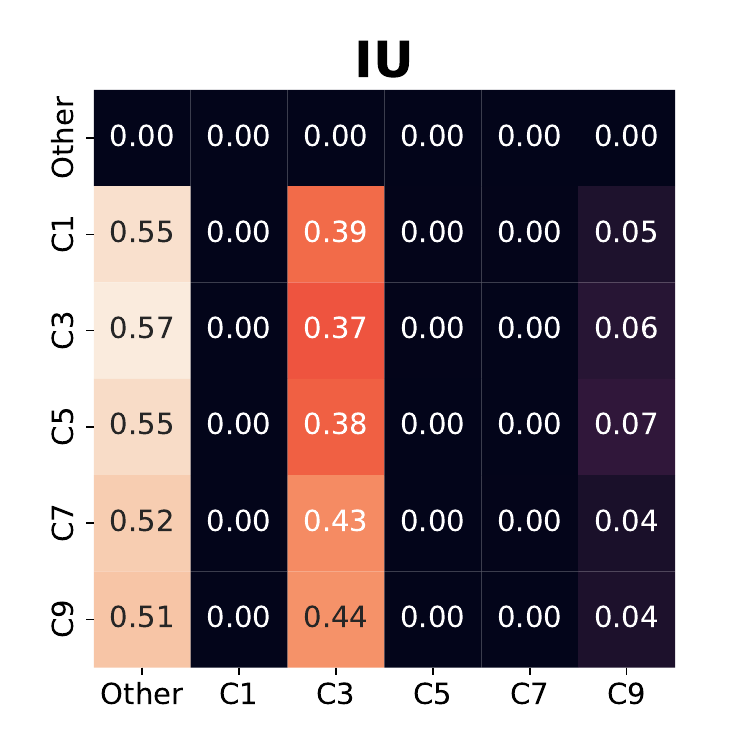}
		\includegraphics[width=0.24\linewidth]{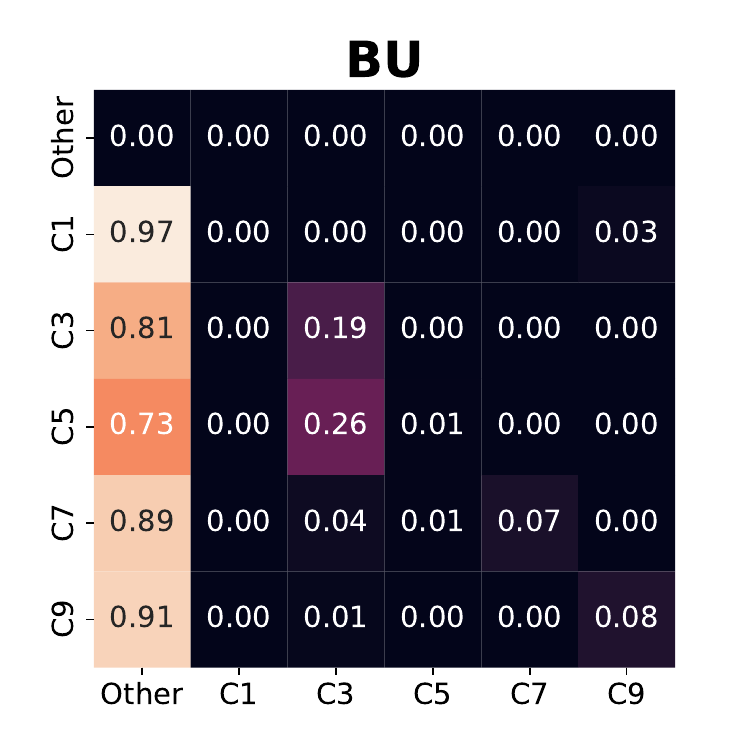}
		\includegraphics[width=0.24\linewidth]{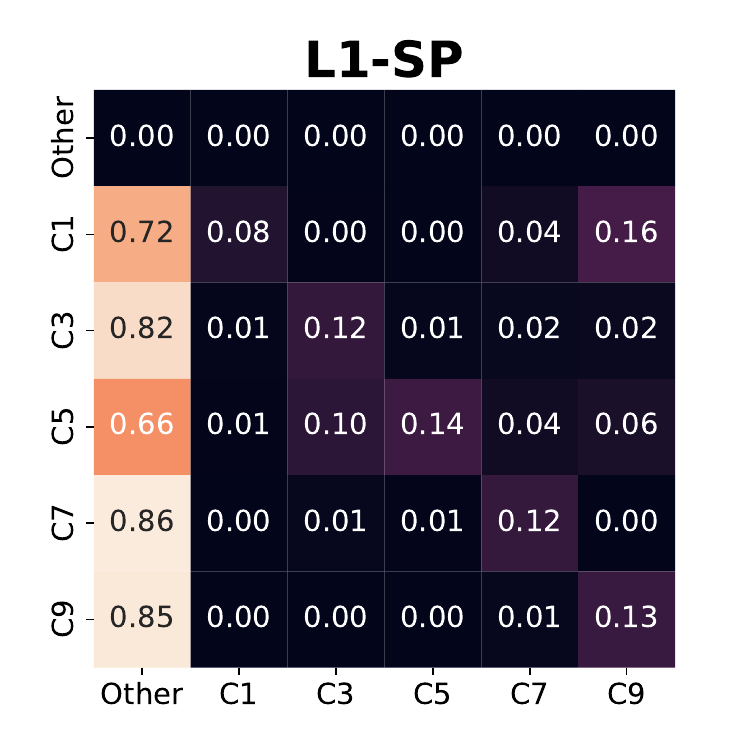}
		\includegraphics[width=0.24\linewidth]{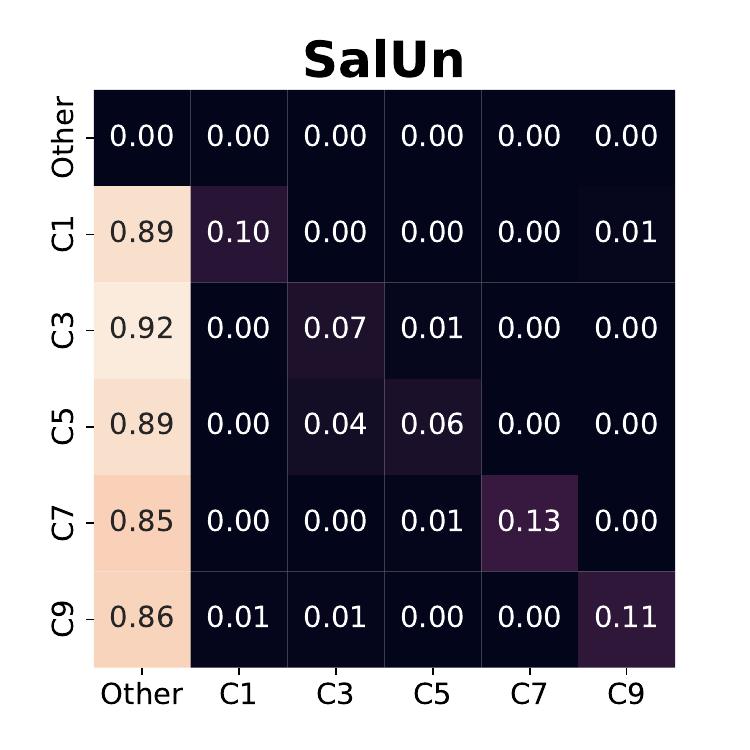}
		\includegraphics[width=0.24\linewidth]{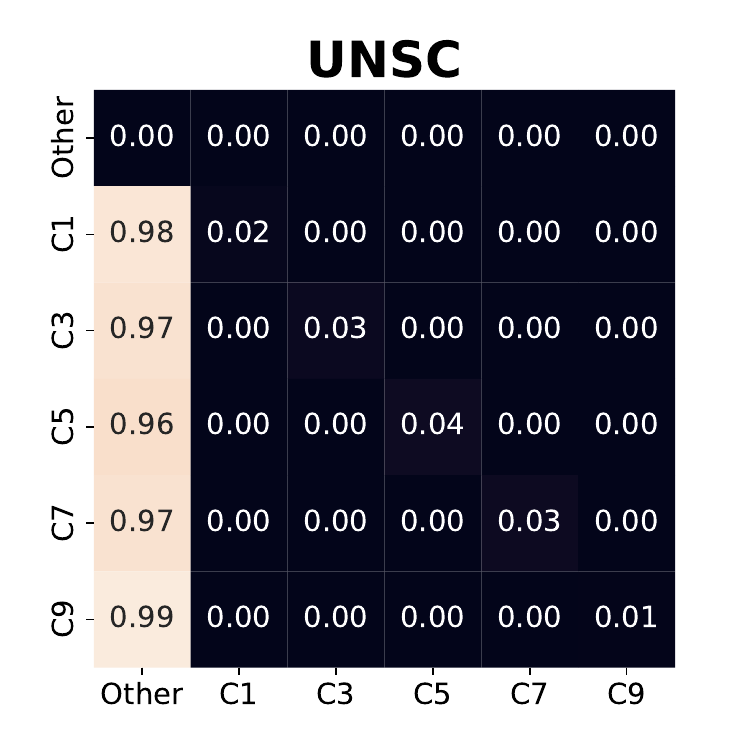}
		\caption{Confusion matrices of \textbf{ULM} w.r.t. different MU methods}
	\end{minipage}
\end{figure*}

\begin{figure*}[ht!]
	\begin{minipage}[c]{\linewidth}
		\flushleft
		\includegraphics[width=0.24\linewidth]{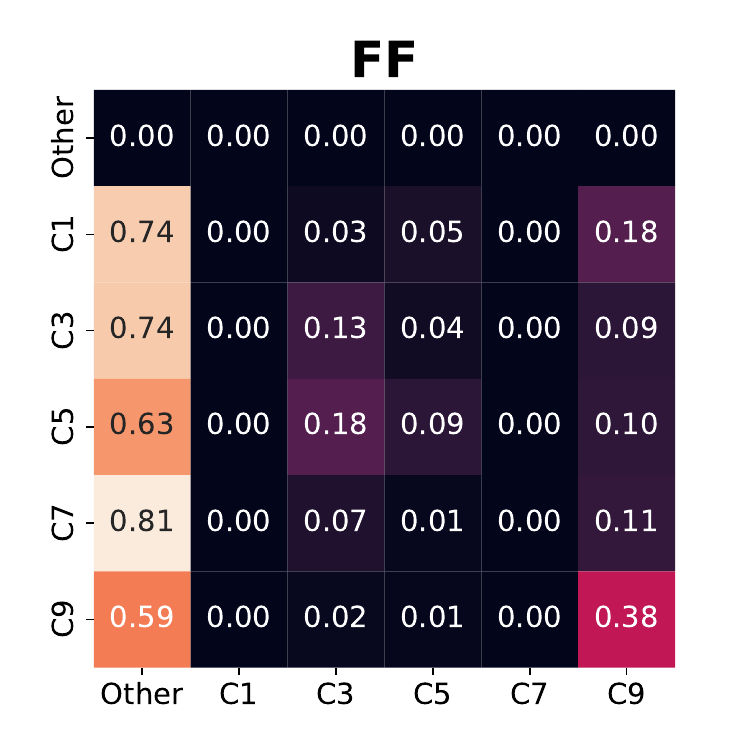}
		\includegraphics[width=0.24\linewidth]{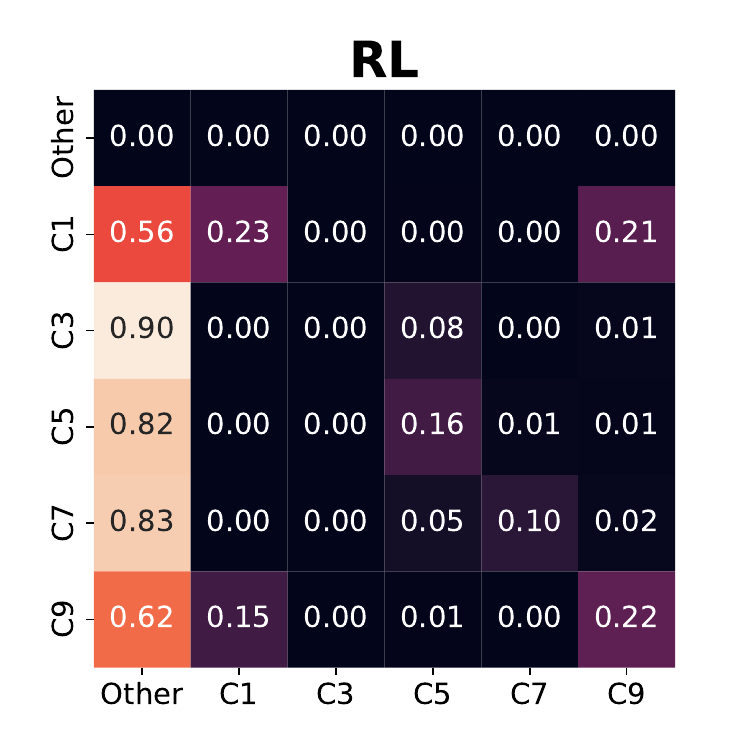}
		\includegraphics[width=0.24\linewidth]{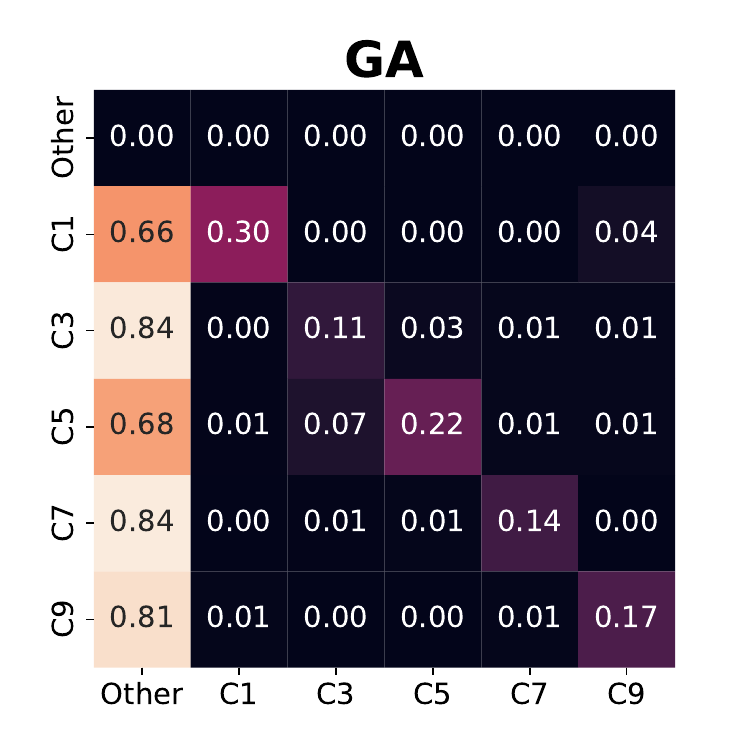}
		\includegraphics[width=0.24\linewidth]{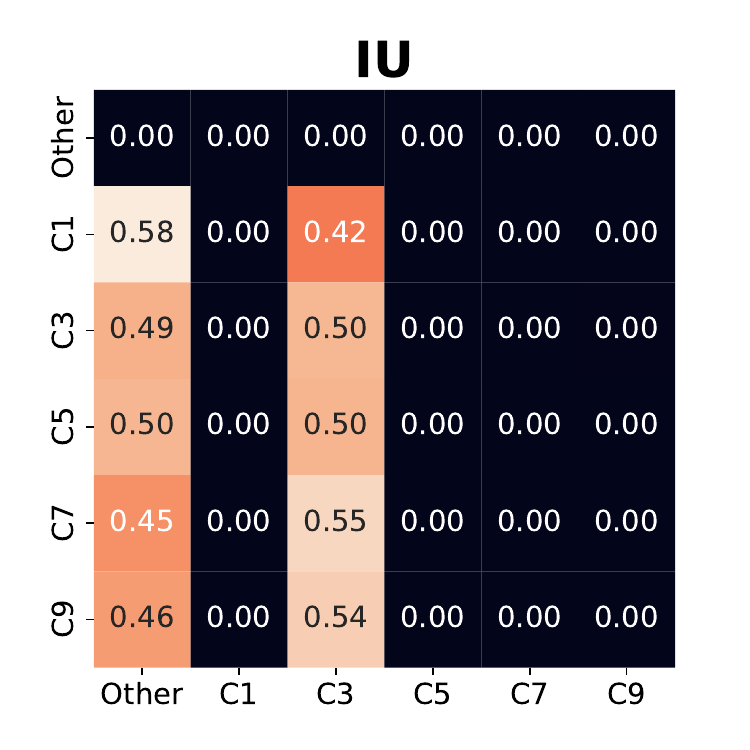}
		\includegraphics[width=0.24\linewidth]{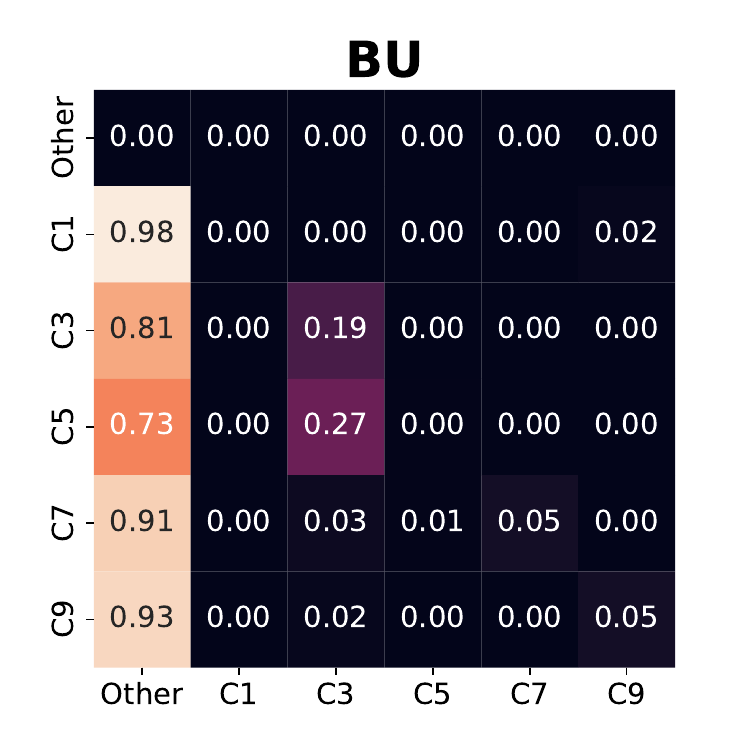}
		\includegraphics[width=0.24\linewidth]{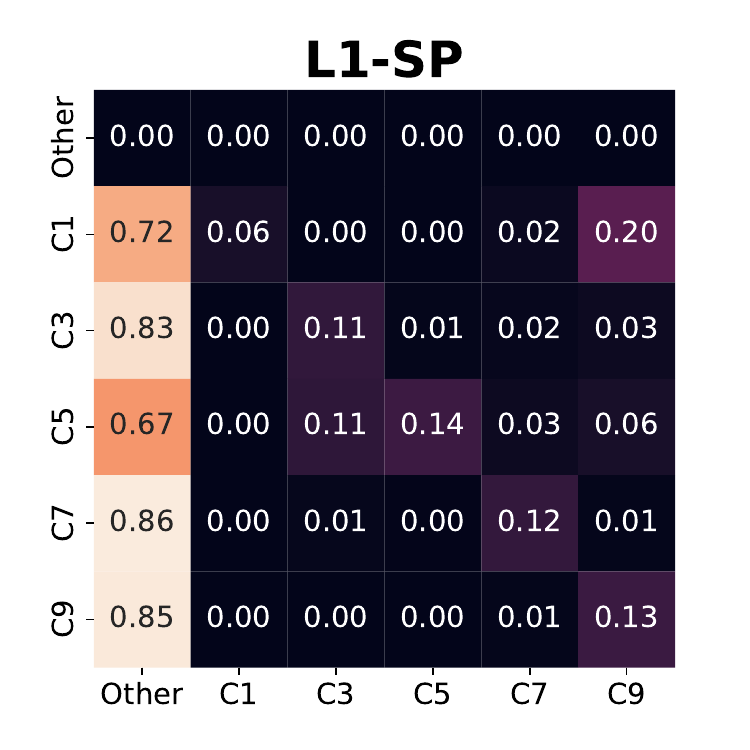}
		\includegraphics[width=0.24\linewidth]{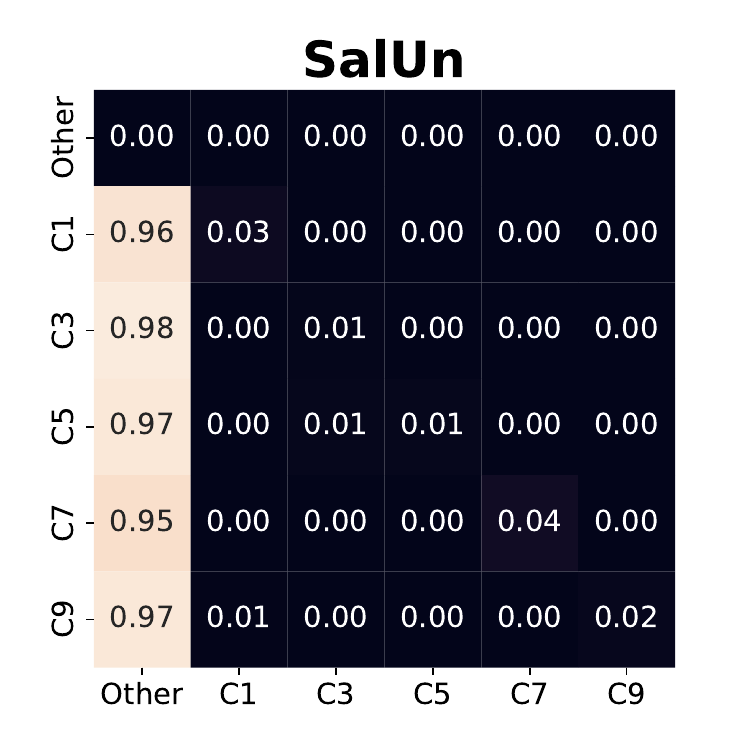}
		\includegraphics[width=0.24\linewidth]{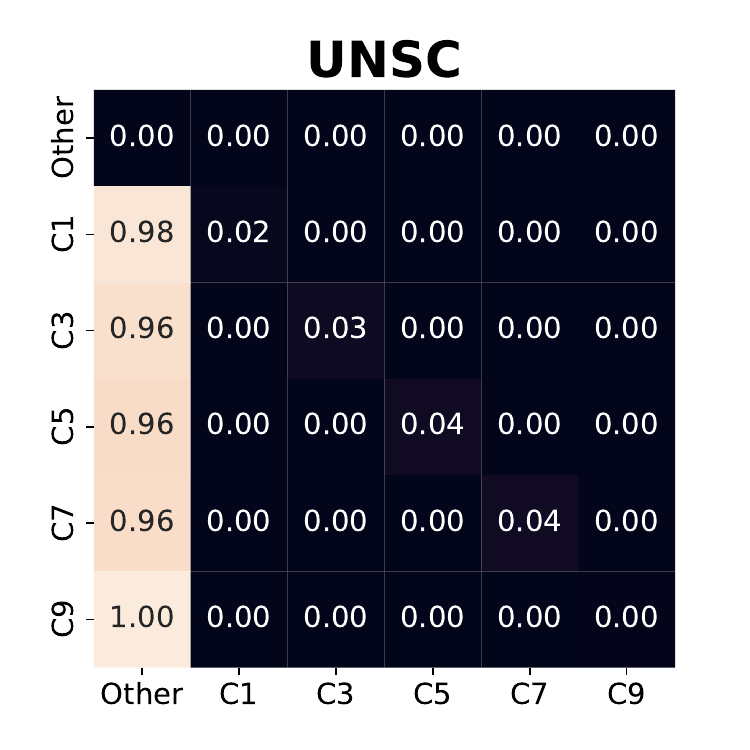}
		\caption{Confusion matrices of \textbf{DST} w.r.t. different MU methods}
	\end{minipage}
\end{figure*}

\begin{figure*}[ht!]
	\begin{minipage}[c]{\linewidth}
		\flushleft
		\includegraphics[width=0.24\linewidth]{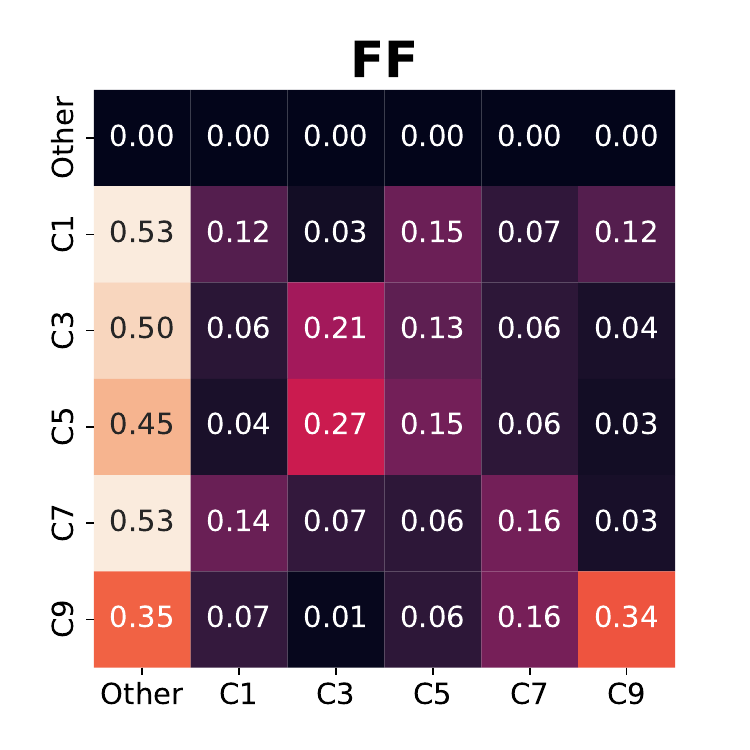}
		\includegraphics[width=0.24\linewidth]{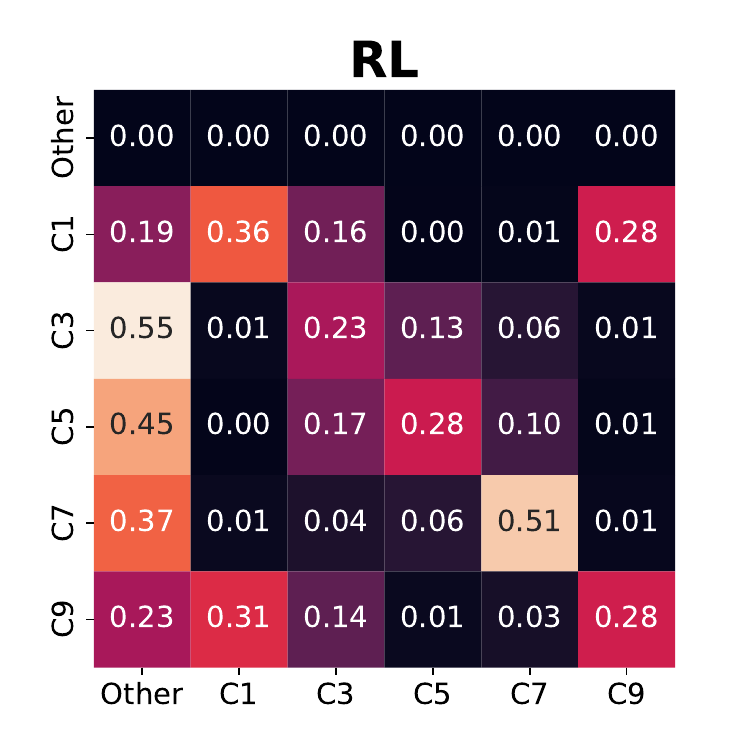}
		\includegraphics[width=0.24\linewidth]{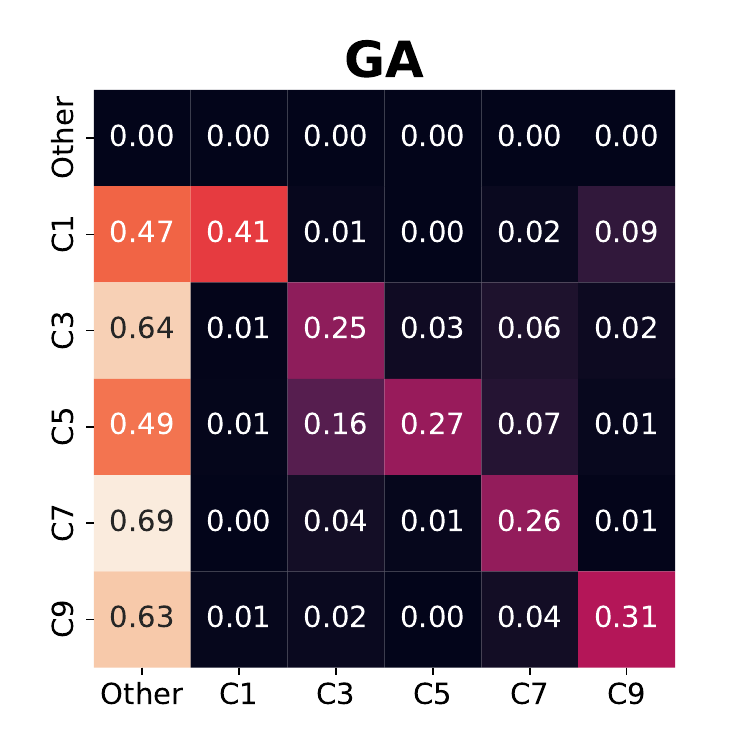}
		\includegraphics[width=0.24\linewidth]{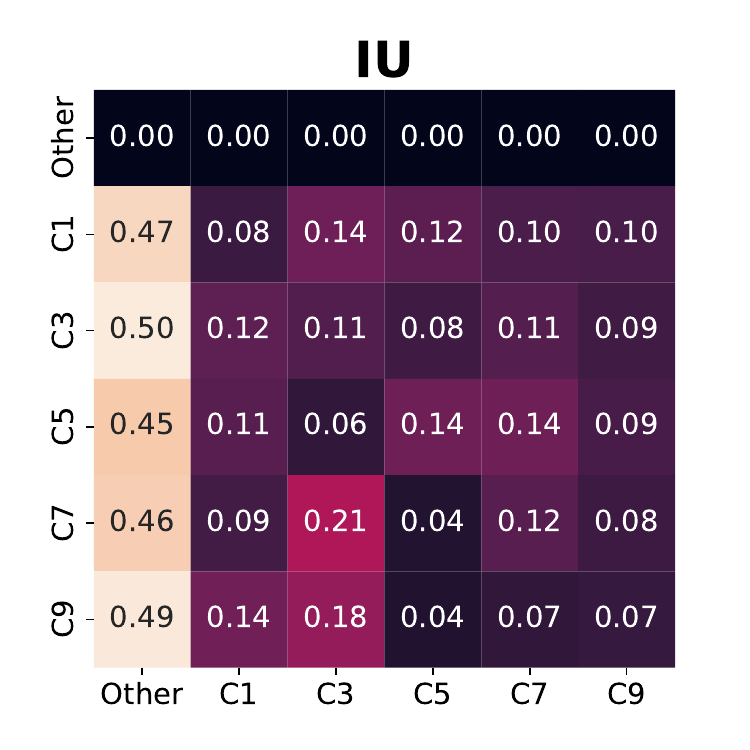}
		\includegraphics[width=0.24\linewidth]{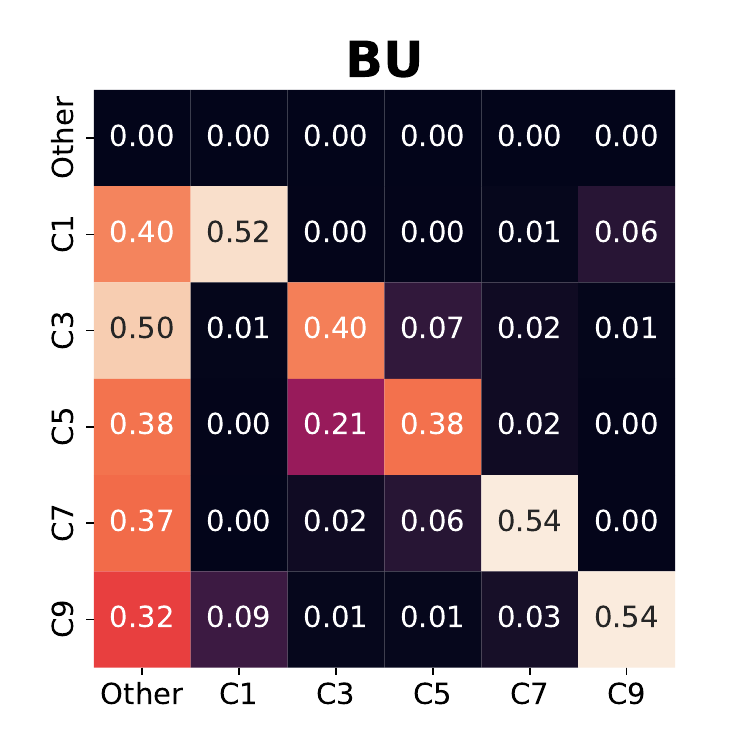}
		\includegraphics[width=0.24\linewidth]{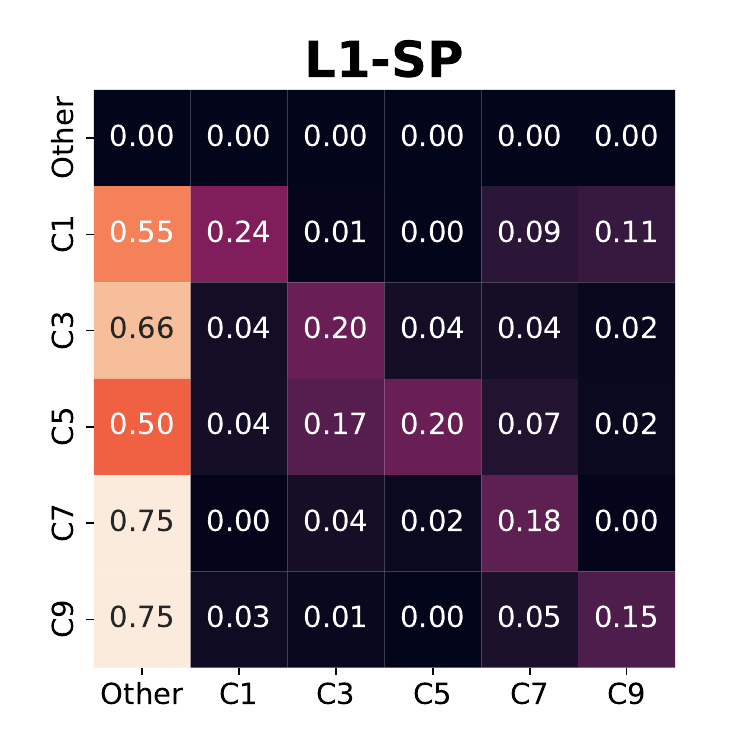}
		\includegraphics[width=0.24\linewidth]{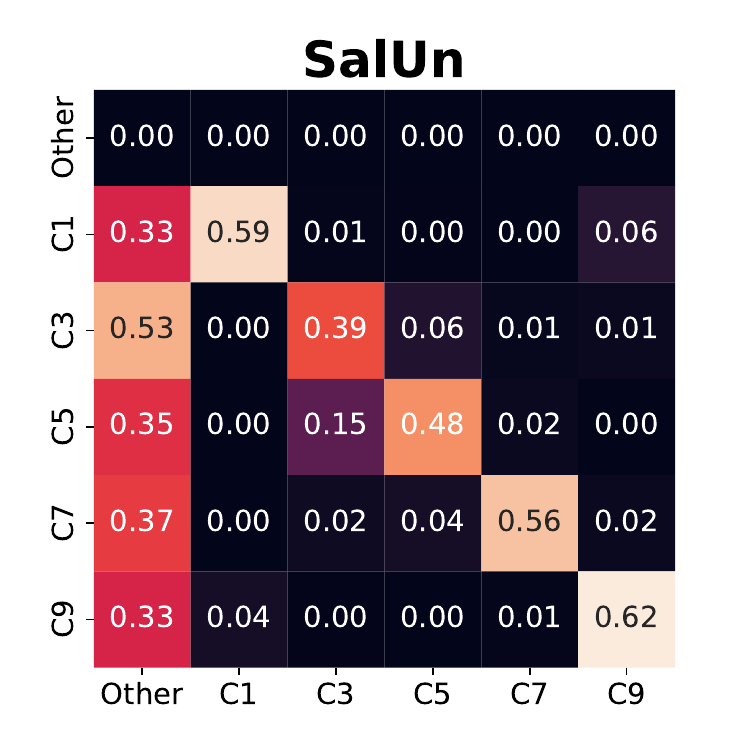}
		\includegraphics[width=0.24\linewidth]{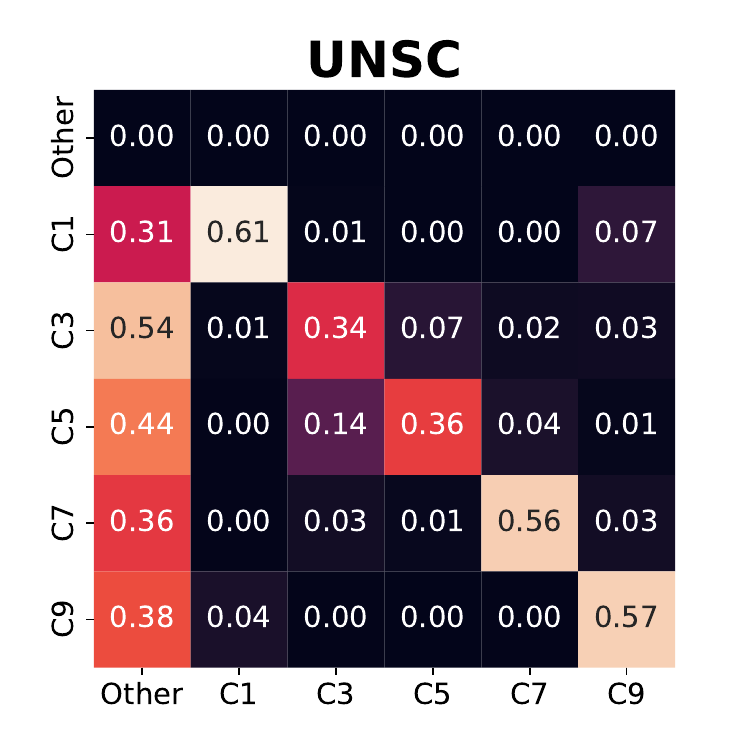}
		\caption{Confusion matrices of \textbf{DST+STU} w.r.t. different MU methods}
	\end{minipage}
\end{figure*}

\begin{figure*}[ht!]
	\begin{minipage}[c]{\linewidth}
		\flushleft
		\includegraphics[width=0.24\linewidth]{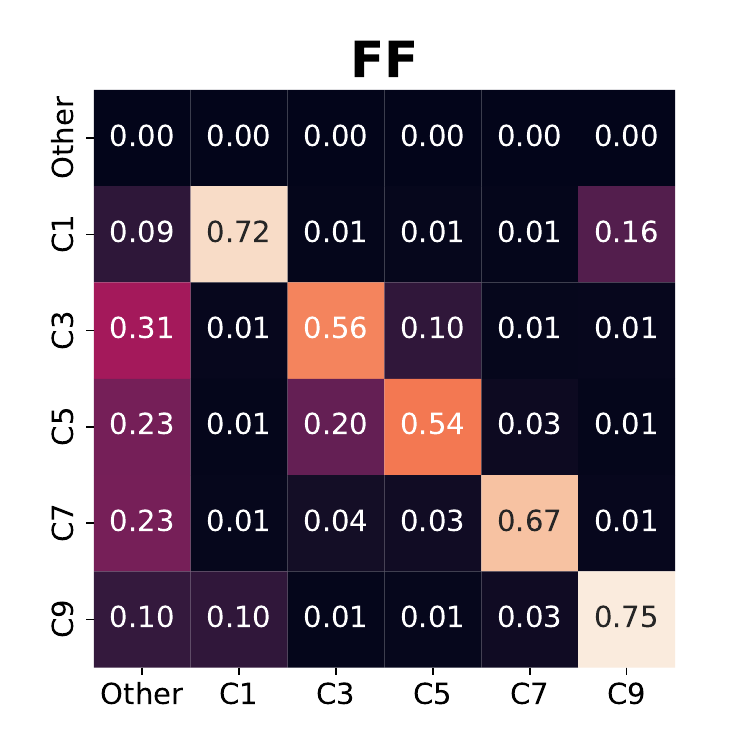}
		\includegraphics[width=0.24\linewidth]{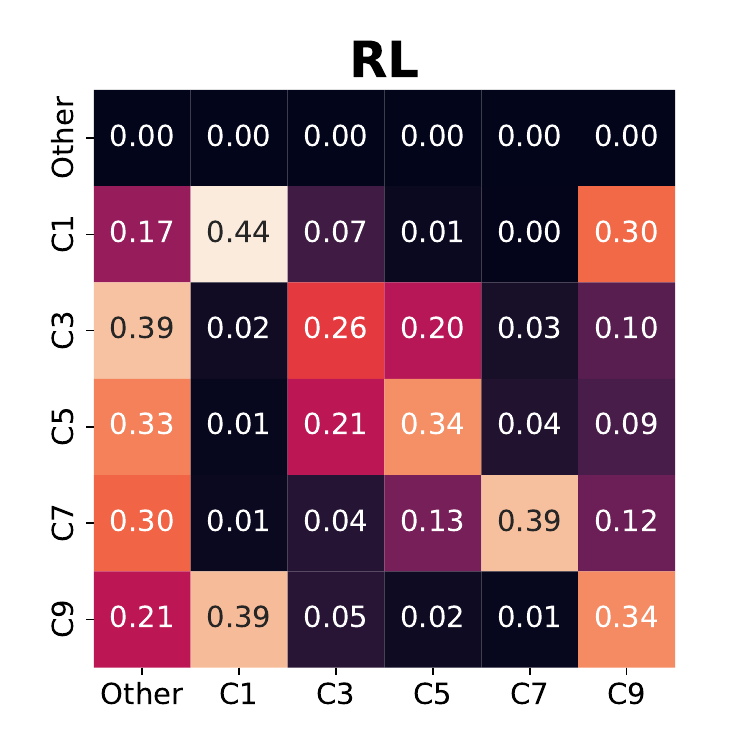}
		\includegraphics[width=0.24\linewidth]{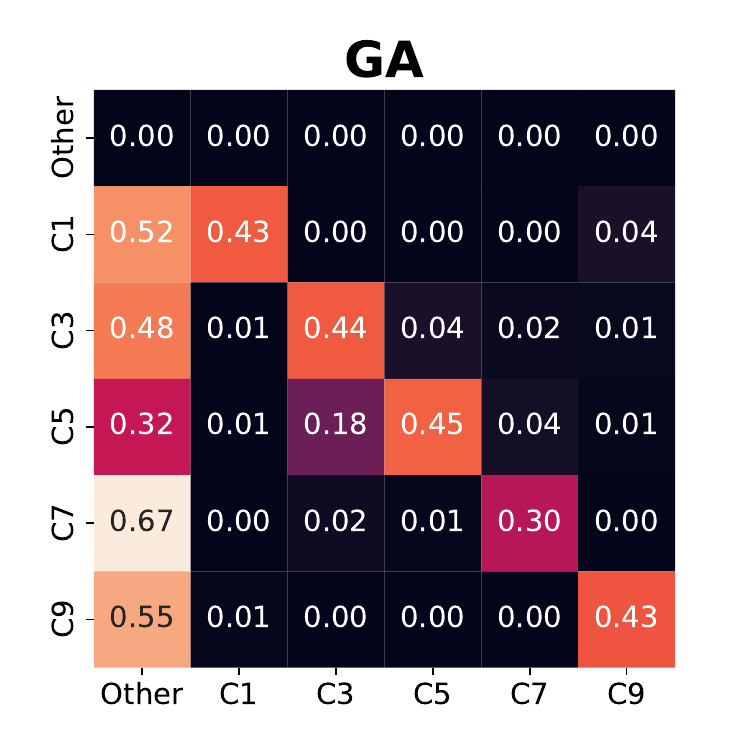}
		\includegraphics[width=0.24\linewidth]{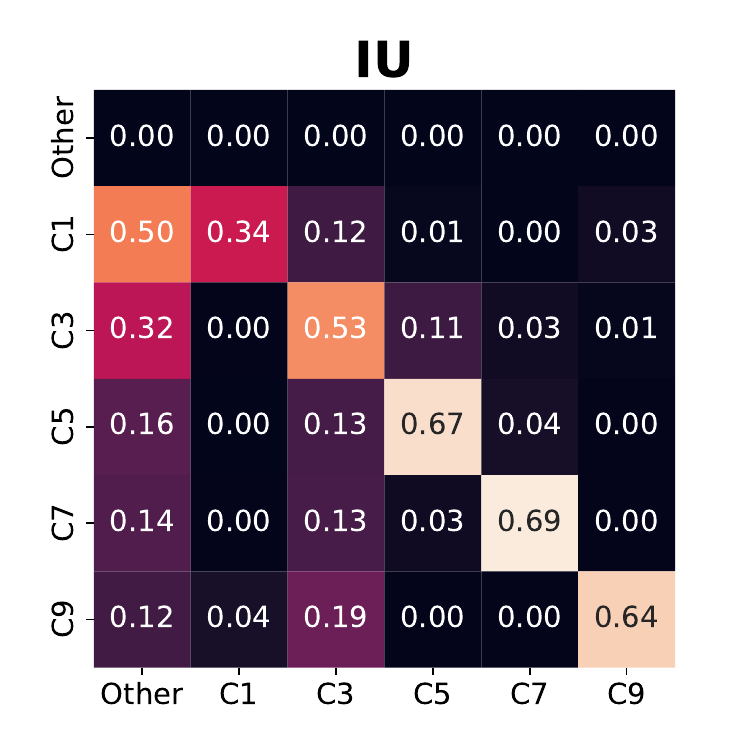}
		\includegraphics[width=0.24\linewidth]{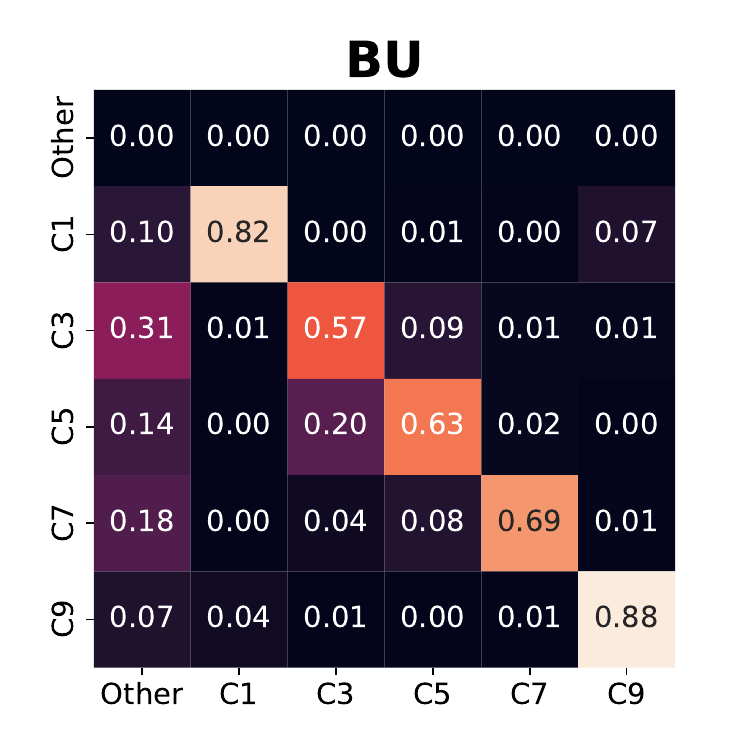}
		\includegraphics[width=0.24\linewidth]{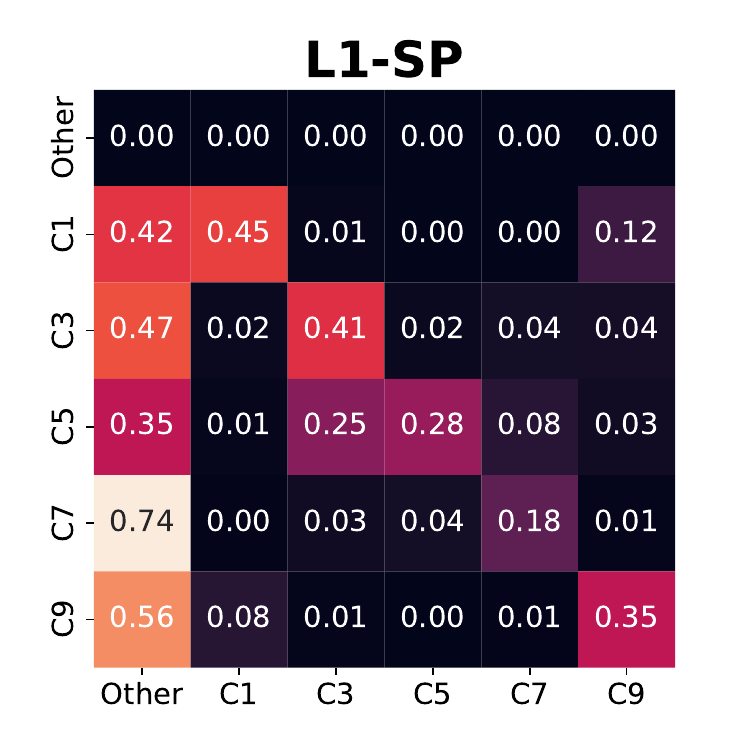}
		\includegraphics[width=0.24\linewidth]{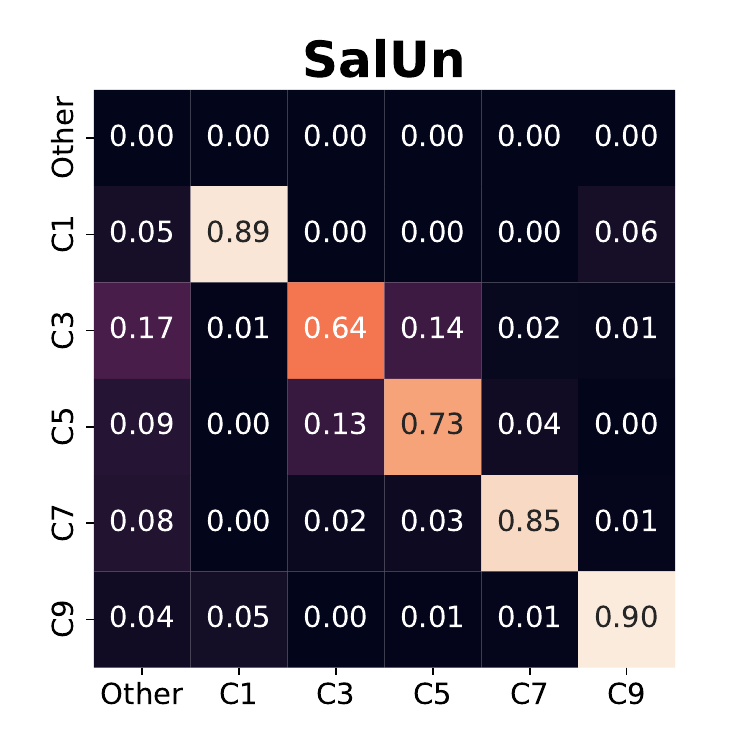}
		\includegraphics[width=0.24\linewidth]{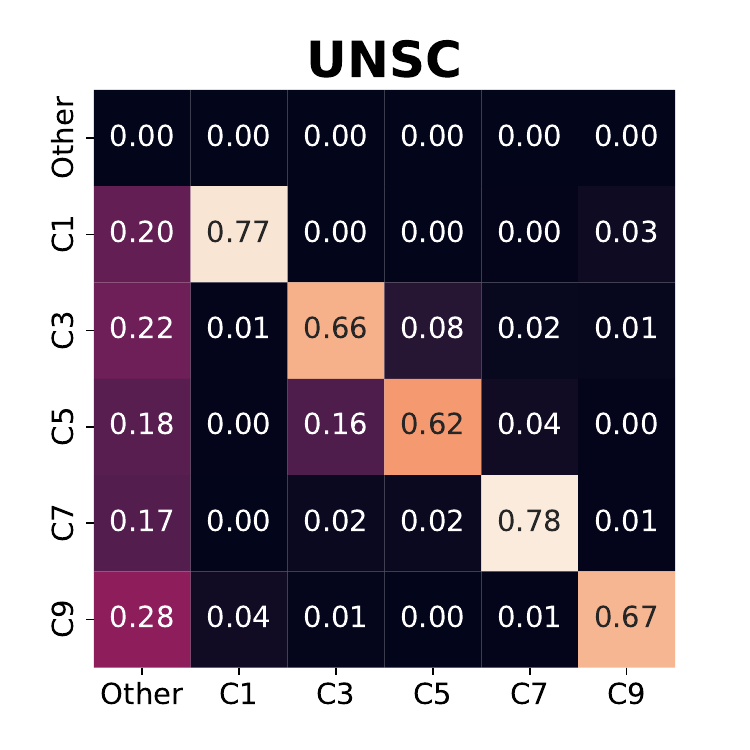}
		\caption{Confusion matrices of the \textbf{DST+STU+TCH} w.r.t. different MU methods}
	\end{minipage}
	\label{fig:open_source_pet37}
\end{figure*}

\subsection{Ablation Study on CIFAR-100}

\begin{table*}[ht!]
	\centering
	\resizebox{\linewidth}{!}
	{
		\begin{tabular}{c|c|c||c|c|c|c|c|c|c|c|c|c|c|c|c|c|c|c}
			\toprule
			\multicolumn{3}{c||}{\textbf{Component}} & \multicolumn{2}{c|}{\textbf{FF}} & \multicolumn{2}{c|}{\textbf{RL}} & \multicolumn{2}{c|}{\textbf{GA}} & \multicolumn{2}{c|}{\textbf{IU}} & \multicolumn{2}{c|}{\textbf{BU}} & \multicolumn{2}{c|}{\textbf{L1-SP}} & \multicolumn{2}{c|}{\textbf{SalUn}} & \multicolumn{2}{c}{\textbf{UNSC}} \\
			\midrule
			\textbf{DST} & \textbf{STU} & \textbf{TCH} & $\mathcal{D}_{ts}$ & $\mathcal{D}_f$  & $\mathcal{D}_{ts}$ & $\mathcal{D}_f$  & $\mathcal{D}_{ts}$ & $\mathcal{D}_f$  & $\mathcal{D}_{ts}$ & $\mathcal{D}_f$  & $\mathcal{D}_{ts}$ & $\mathcal{D}_f$  & $\mathcal{D}_{ts}$ & $\mathcal{D}_f$  & $\mathcal{D}_{ts}$ & $\mathcal{D}_f$  & $\mathcal{D}_{ts}$ & $\mathcal{D}_f$ \\
			\midrule
			\checkmark &       &       & \multicolumn{1}{c}{0.173 } & \multicolumn{1}{c}{0.103 } & \multicolumn{1}{c}{0.594 } & \multicolumn{1}{c}{0.015 } & \multicolumn{1}{c}{0.528 } & \multicolumn{1}{c}{0.180 } & \multicolumn{1}{c}{0.135 } & \multicolumn{1}{c}{0.194 } & \multicolumn{1}{c}{0.532 } & \multicolumn{1}{c}{0.210 } & \multicolumn{1}{c}{0.536 } & \multicolumn{1}{c}{0.183 } & \multicolumn{1}{c}{0.440 } & \multicolumn{1}{c}{0.142 } & \multicolumn{1}{c}{0.244 } & 0.131  \\
			\midrule
			\checkmark & \checkmark &       & 0.295  & 0.214  & 0.607  & 0.537  & 0.551  & 0.270  & 0.297  & 0.256  & 0.566  & 0.423  & 0.550  & 0.258  & 0.522  & 0.358  & 0.480  & 0.372  \\
			\midrule
			\checkmark & \checkmark & \checkmark & 0.521  & 0.686  & 0.599  & 0.898  & 0.589  & 0.682  & 0.596  & 0.967  & 0.587  & 0.970  & 0.598  & 0.764  & 0.564  & 0.945  & 0.598  & 0.985  \\
			\bottomrule
		\end{tabular}%
	}
	\caption{Ablation results (\textit{Acc}) of MRA on CIFAR-100 dataset w.r.t. different MU methods}
	\label{tab:addlabel}%
\end{table*}%

\begin{figure*}[ht!]
	\begin{minipage}[c]{\linewidth}
		\flushleft
		\includegraphics[width=0.24\linewidth]{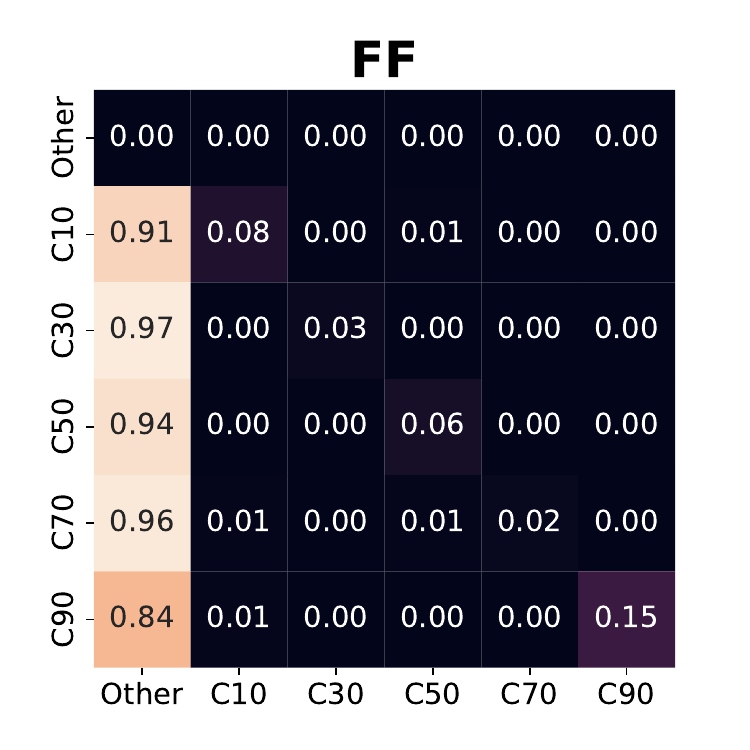}
		\includegraphics[width=0.24\linewidth]{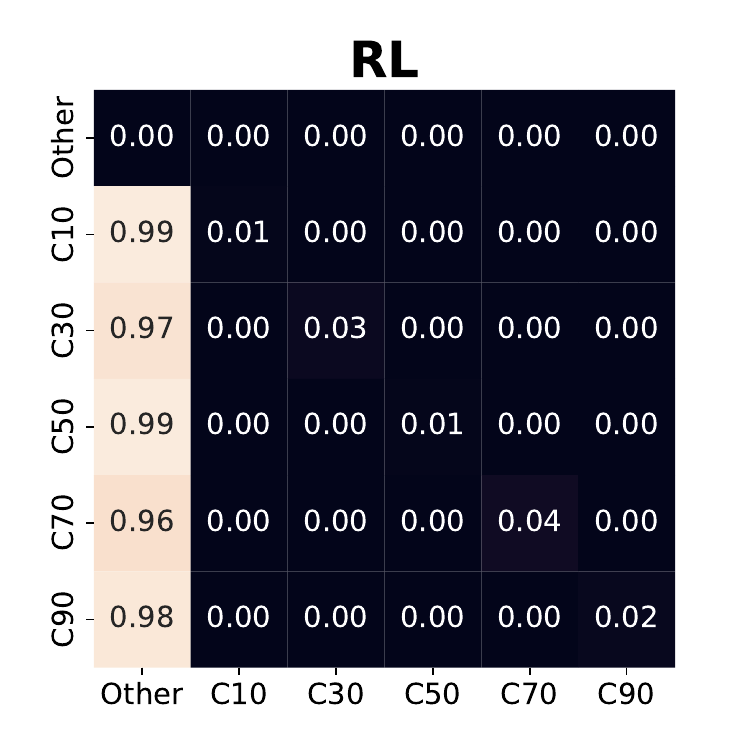}
		\includegraphics[width=0.24\linewidth]{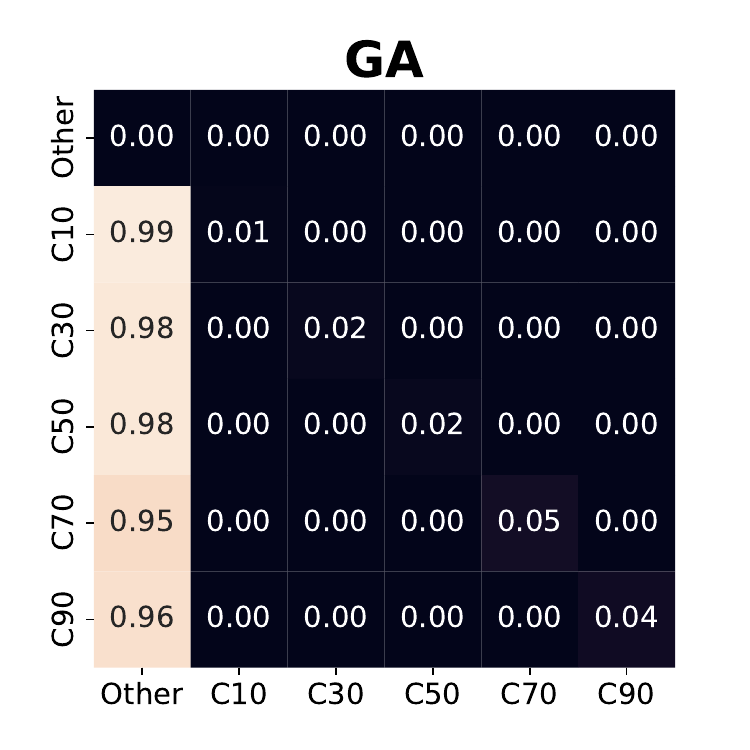}
		\includegraphics[width=0.24\linewidth]{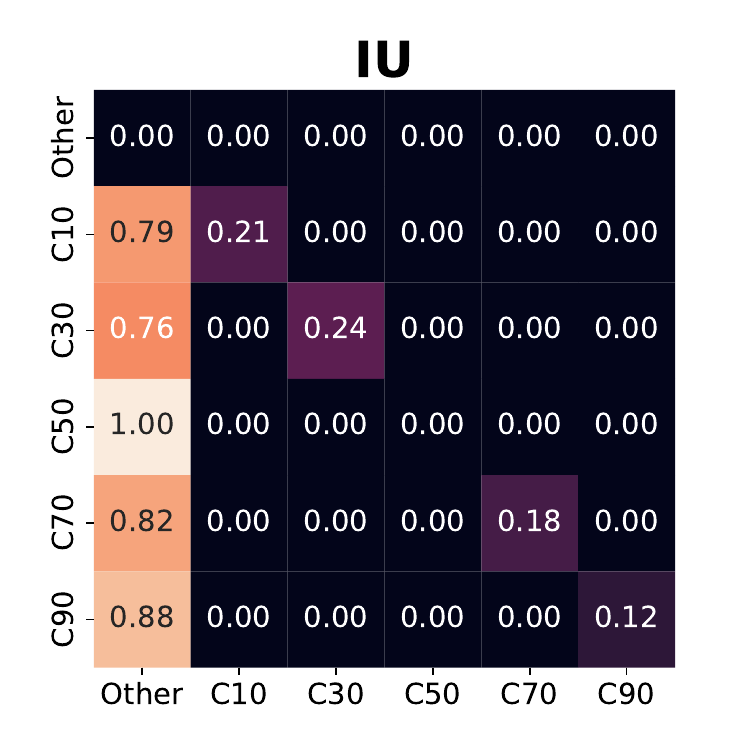}
		\includegraphics[width=0.24\linewidth]{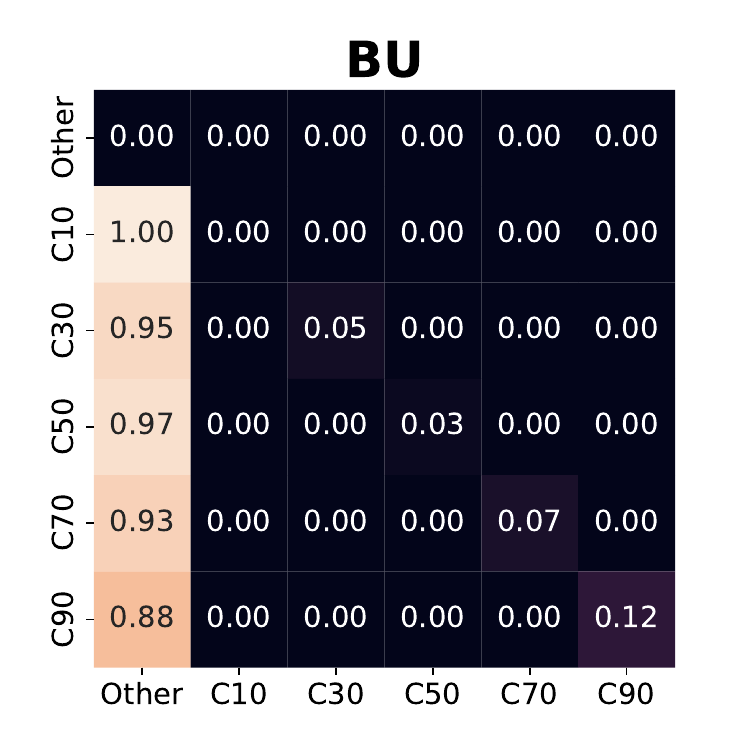}
		\includegraphics[width=0.24\linewidth]{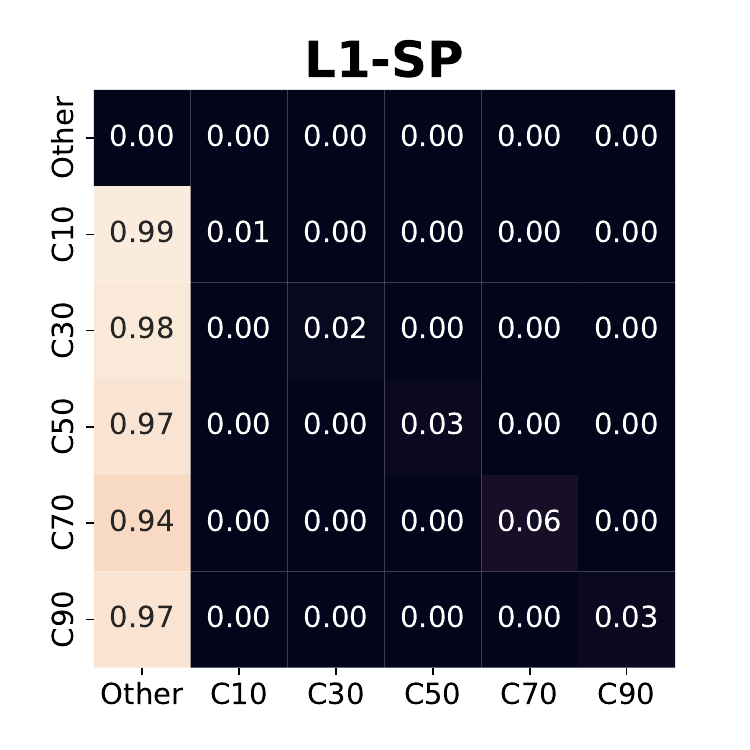}
		\includegraphics[width=0.24\linewidth]{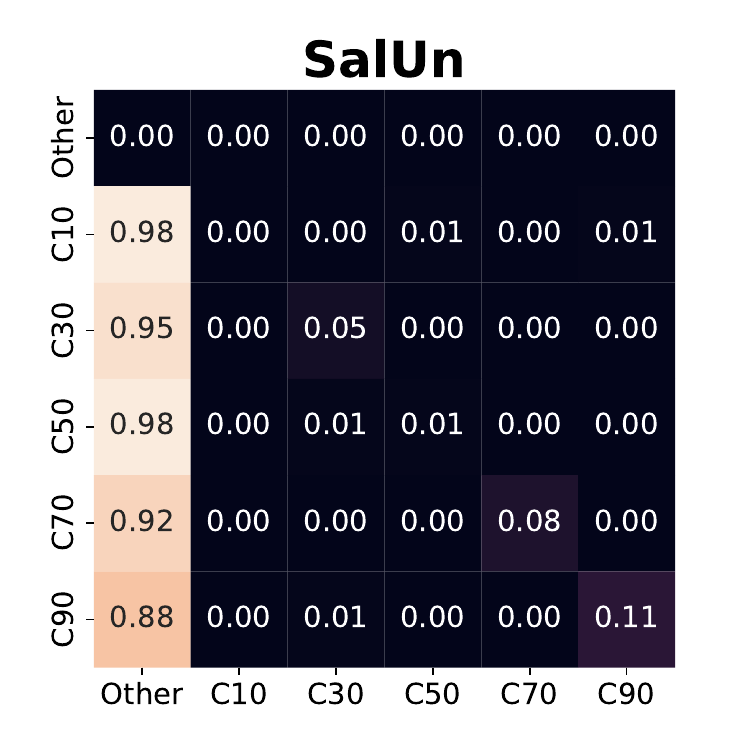}
		\includegraphics[width=0.24\linewidth]{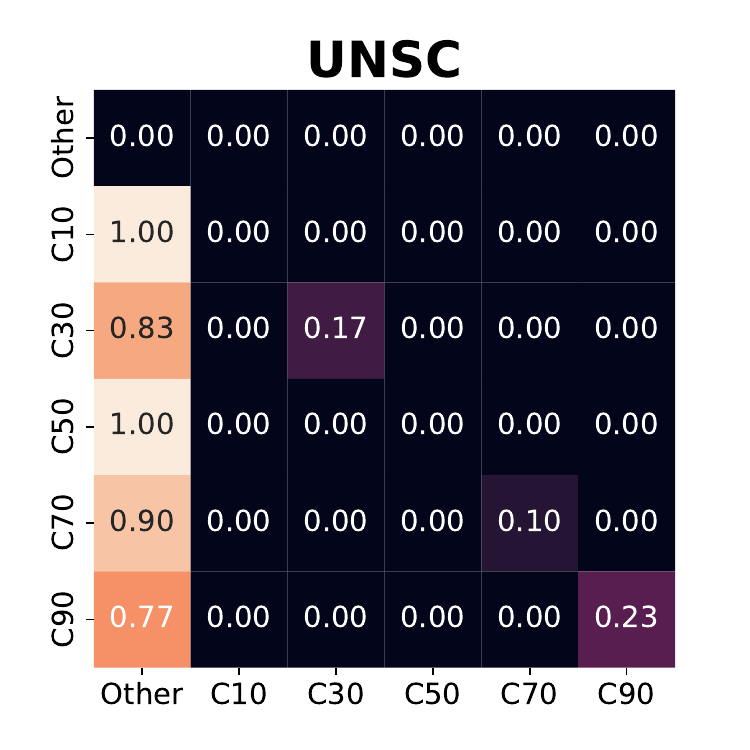}
		\caption{Confusion matrices of the \textbf{UML} w.r.t. different MU methods}
	\end{minipage}
\end{figure*}

\begin{figure*}[ht!]
	\begin{minipage}[c]{\linewidth}
		\flushleft
		\includegraphics[width=0.24\linewidth]{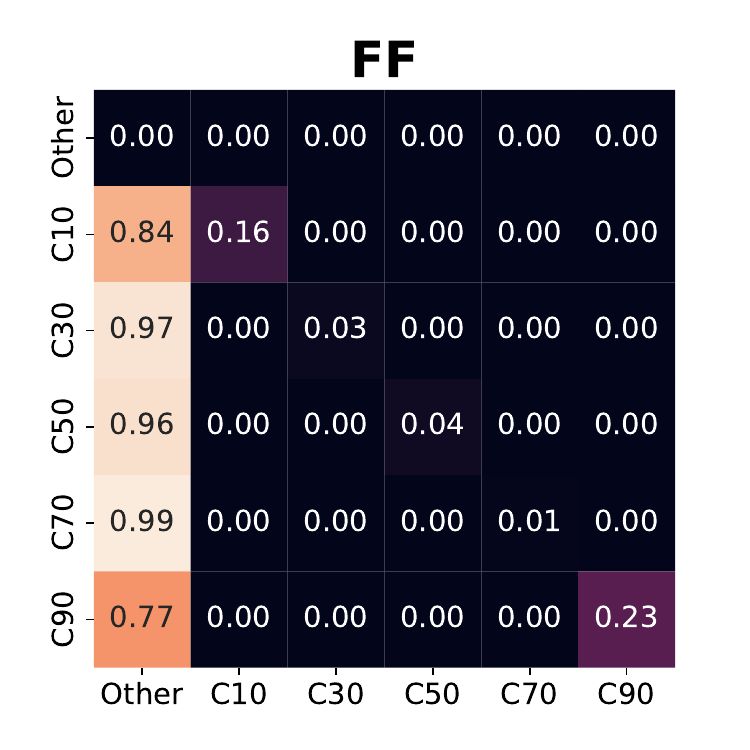}
		\includegraphics[width=0.24\linewidth]{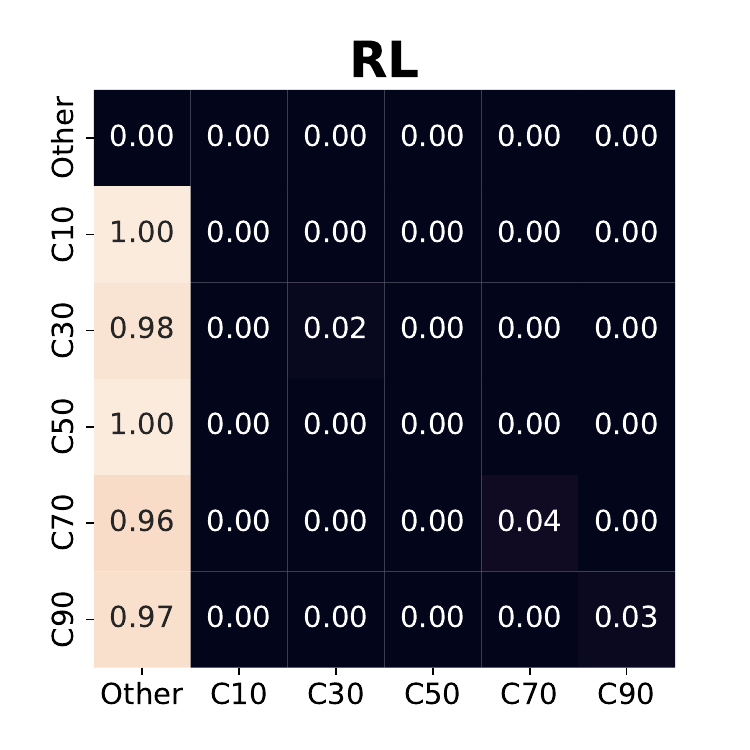}
		\includegraphics[width=0.24\linewidth]{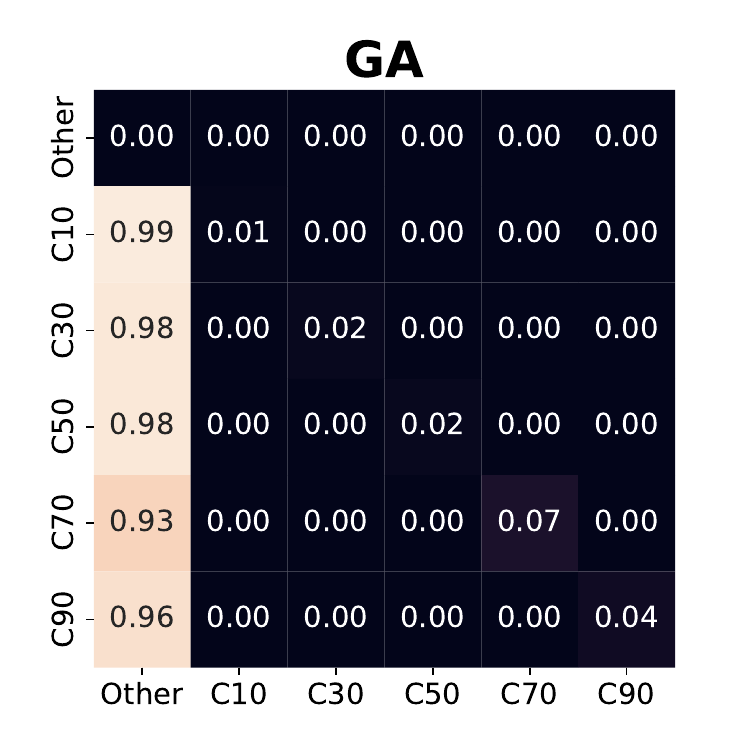}
		\includegraphics[width=0.24\linewidth]{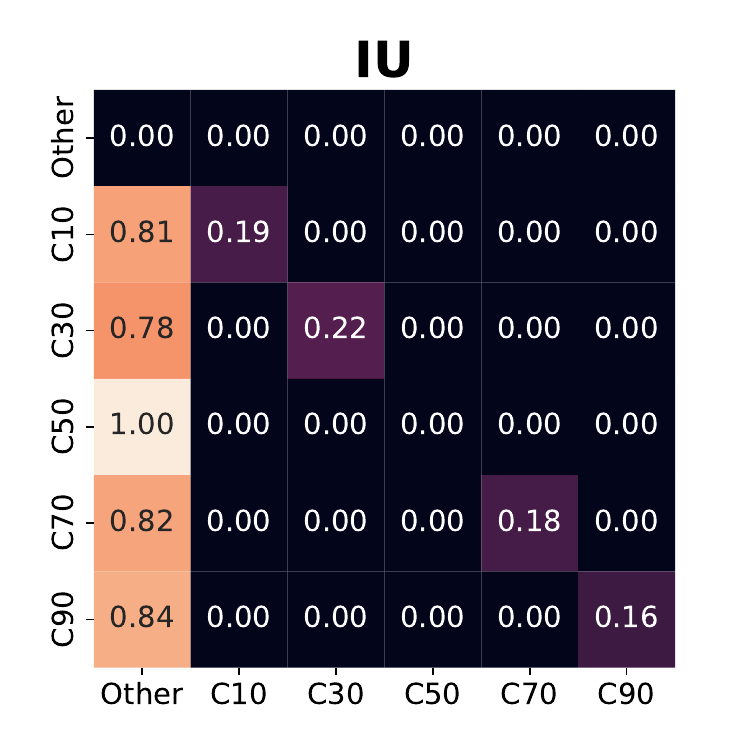}
		\includegraphics[width=0.24\linewidth]{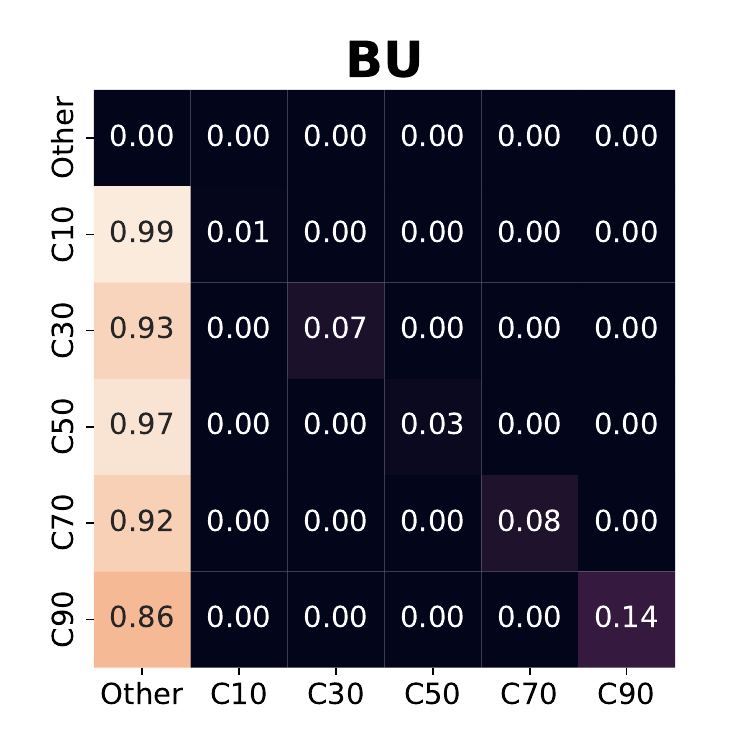}
		\includegraphics[width=0.24\linewidth]{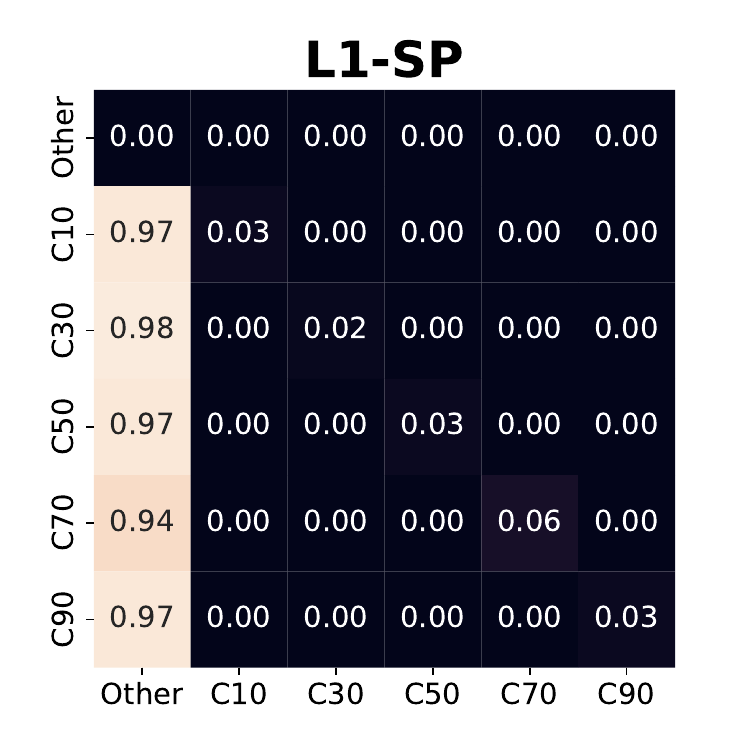}
		\includegraphics[width=0.24\linewidth]{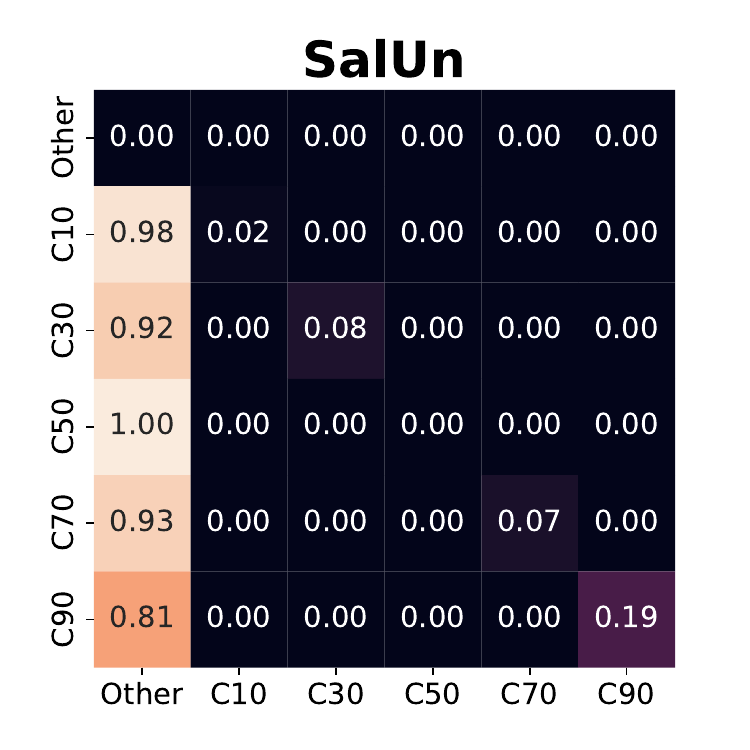}
		\includegraphics[width=0.24\linewidth]{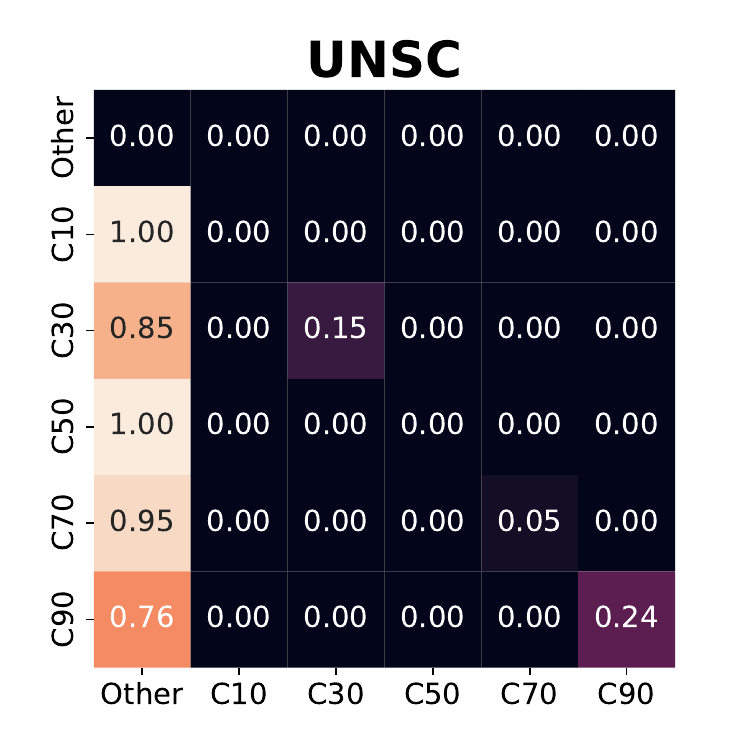}
		\caption{Confusion matrices of the \textbf{DST} w.r.t. different MU methods}
	\end{minipage}
\end{figure*}

\begin{figure*}[ht!]
	\begin{minipage}[c]{\linewidth}
		\flushleft
		\includegraphics[width=0.24\linewidth]{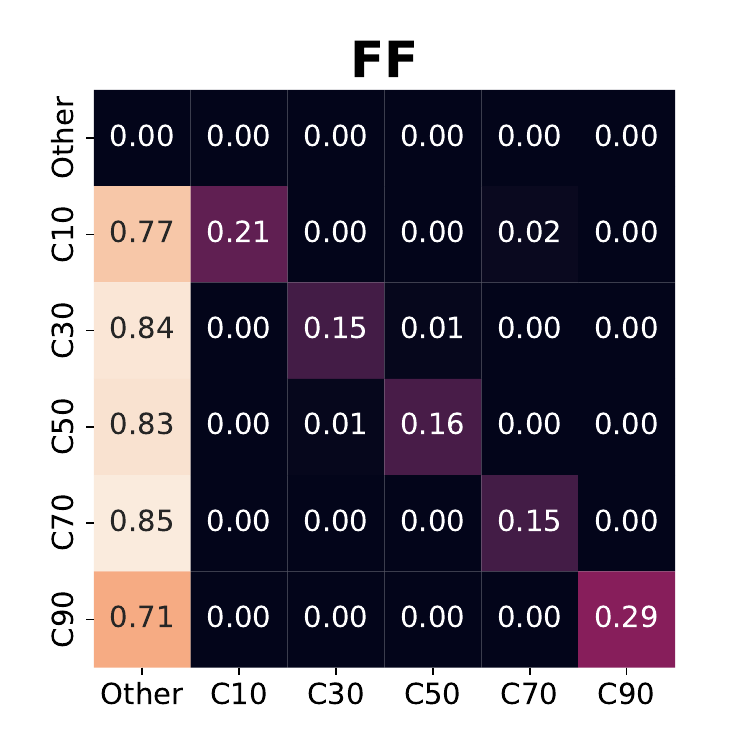}
		\includegraphics[width=0.24\linewidth]{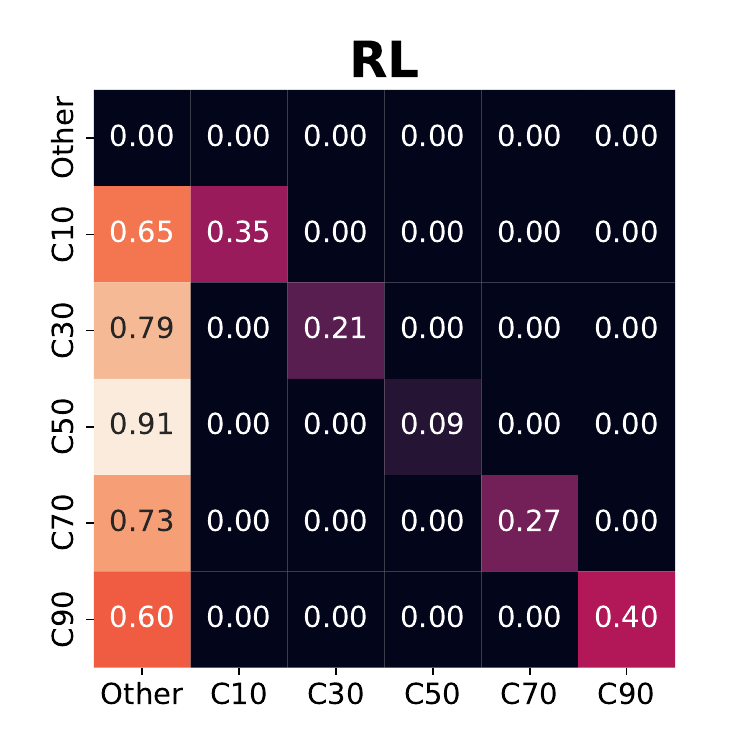}
		\includegraphics[width=0.24\linewidth]{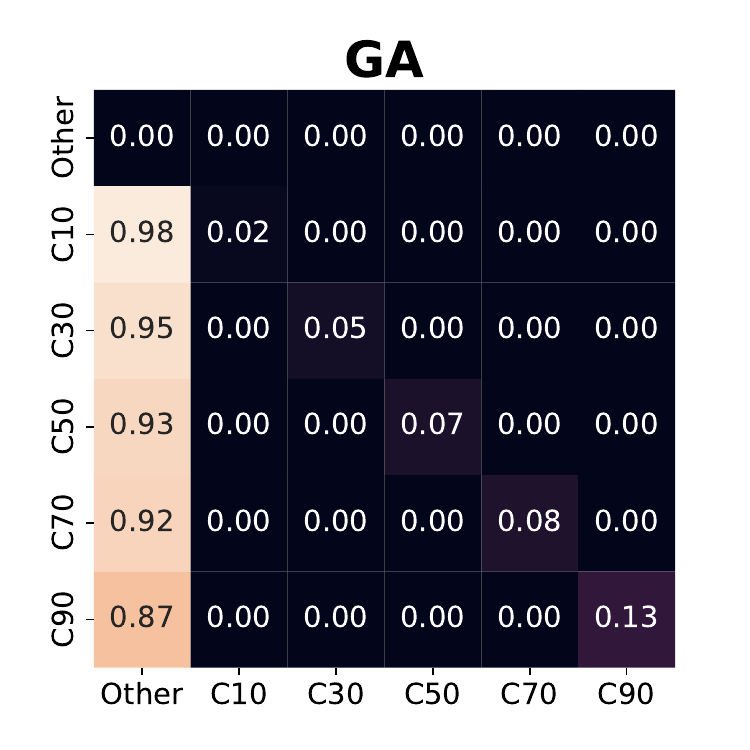}
		\includegraphics[width=0.24\linewidth]{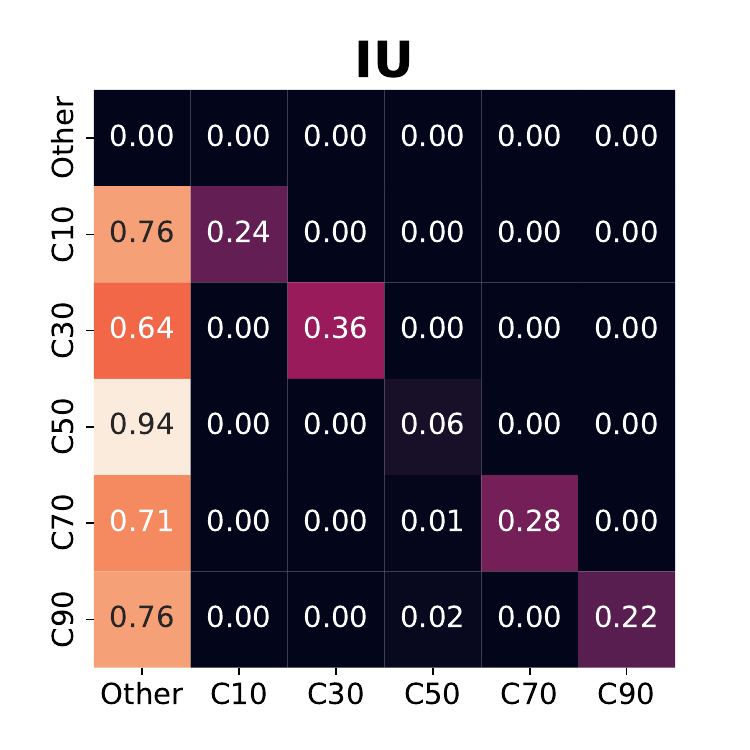}
		\includegraphics[width=0.24\linewidth]{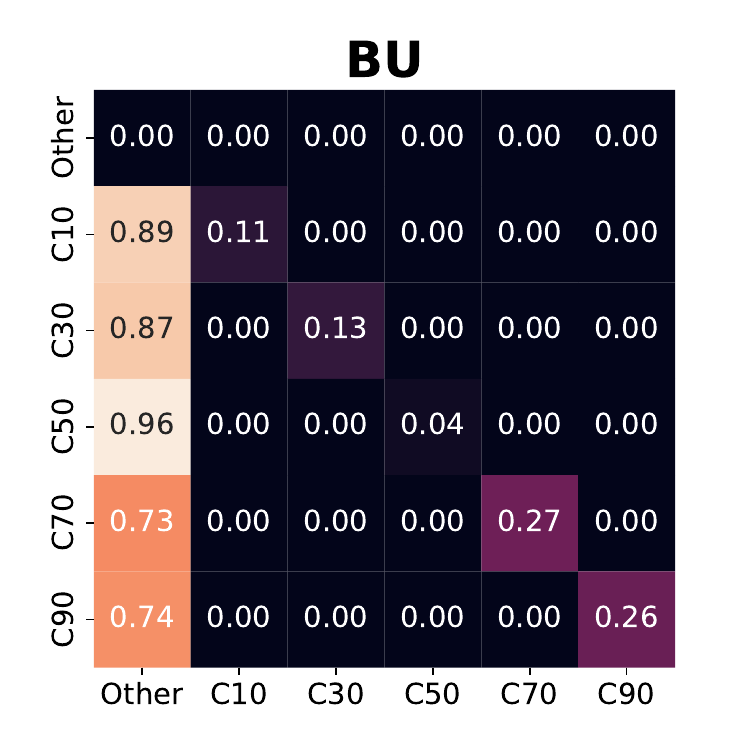}
		\includegraphics[width=0.24\linewidth]{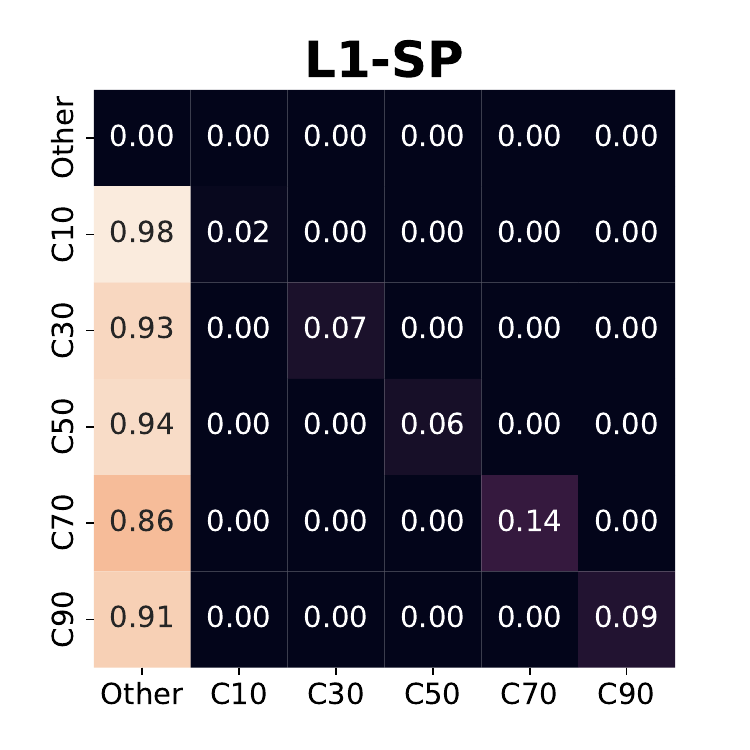}
		\includegraphics[width=0.24\linewidth]{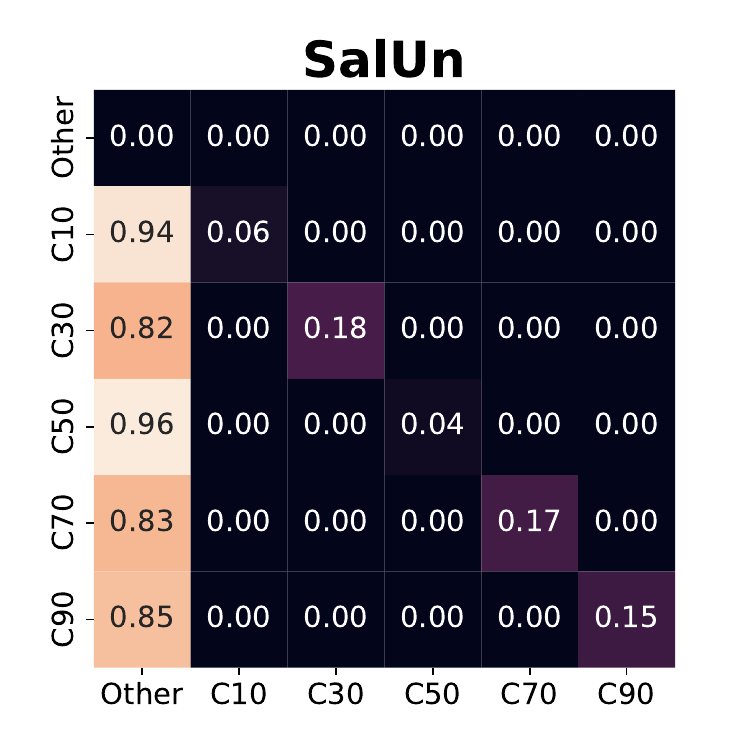}
		\includegraphics[width=0.24\linewidth]{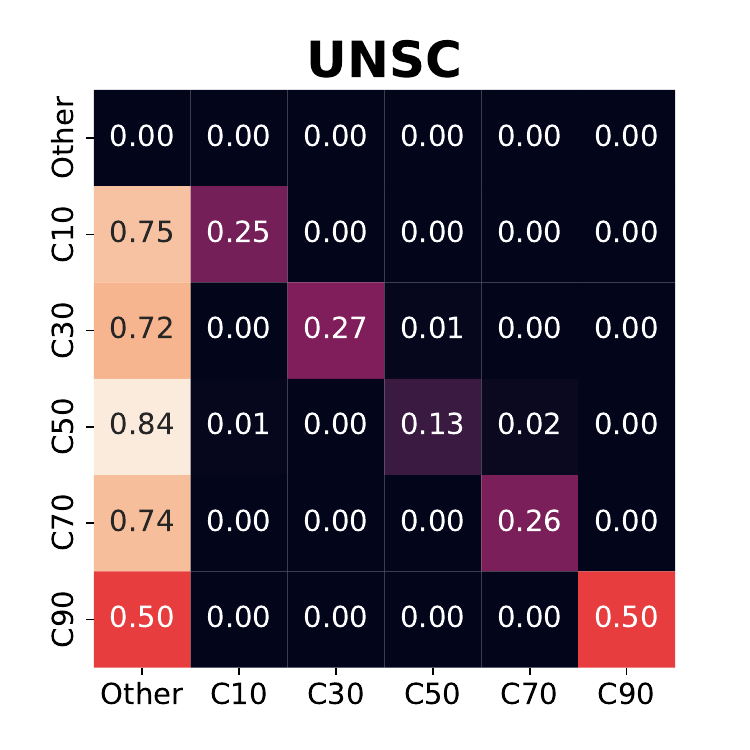}
		\caption{Confusion matrices of the \textbf{DST+STU} w.r.t. different MU methods}
	\end{minipage}
\end{figure*}

\begin{figure*}[ht!]
	\begin{minipage}[c]{\linewidth}
		\flushleft
		\includegraphics[width=0.24\linewidth]{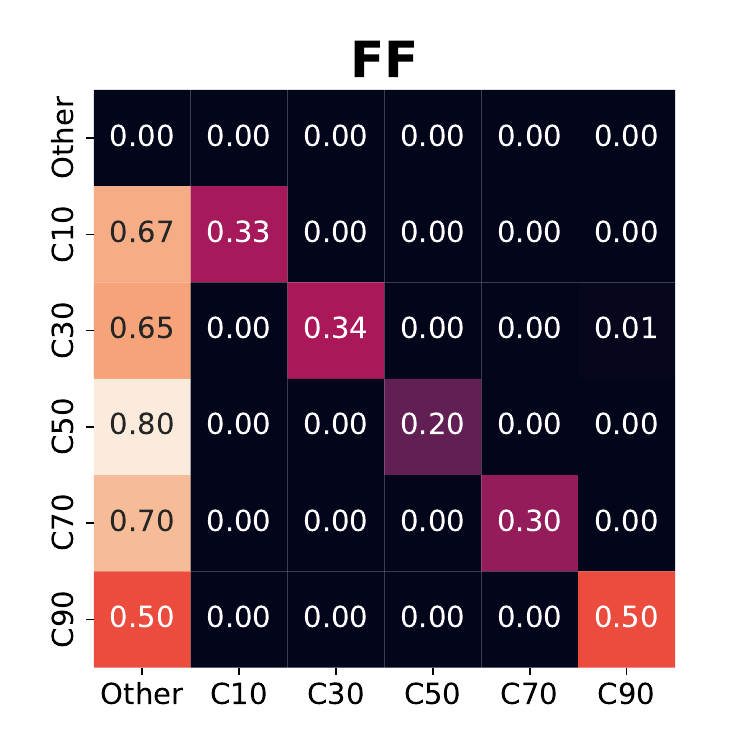}
		\includegraphics[width=0.24\linewidth]{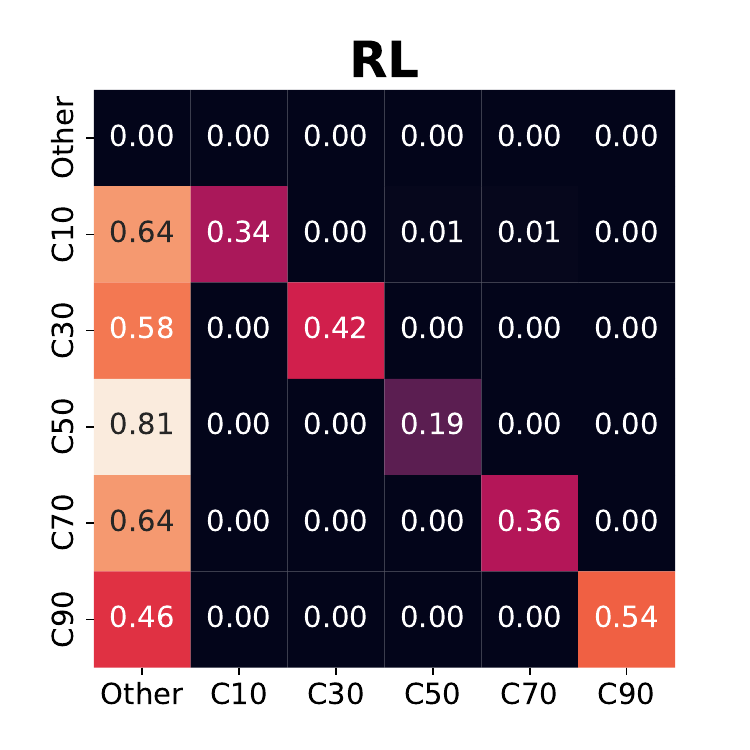}
		\includegraphics[width=0.24\linewidth]{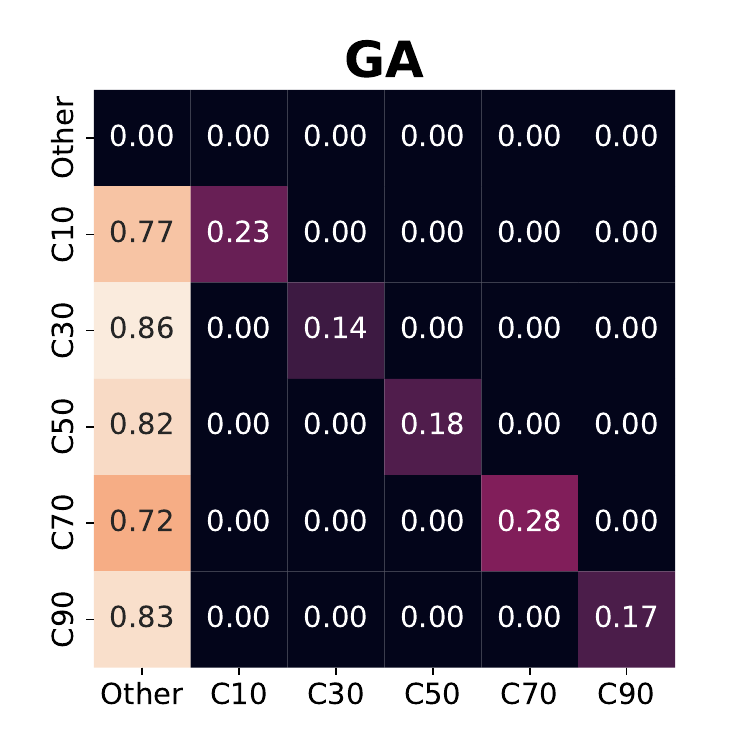}
		\includegraphics[width=0.24\linewidth]{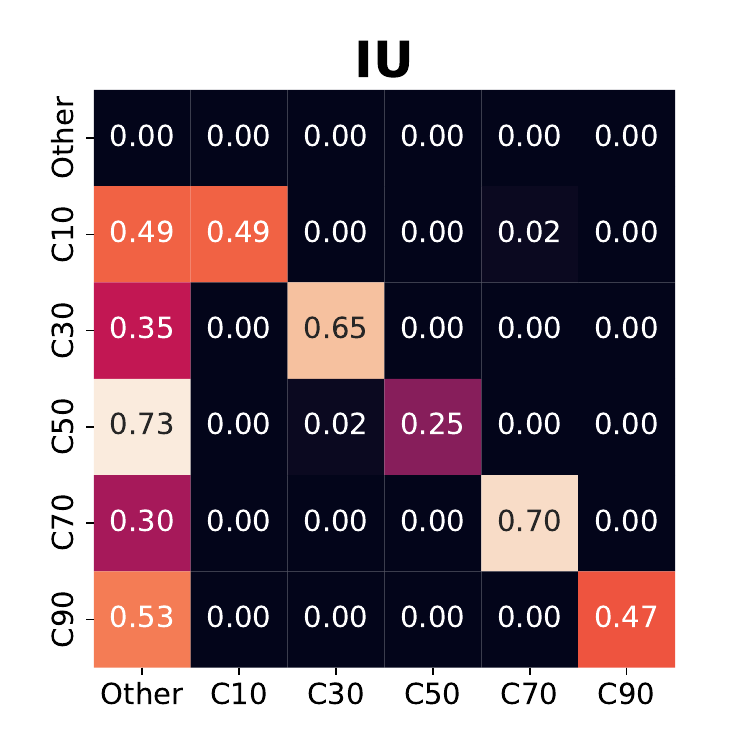}
		\includegraphics[width=0.24\linewidth]{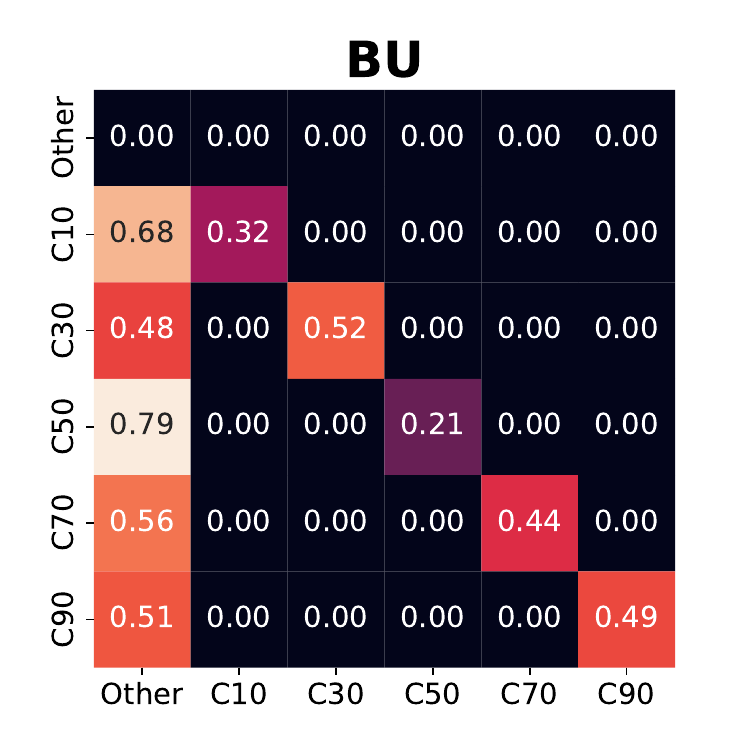}
		\includegraphics[width=0.24\linewidth]{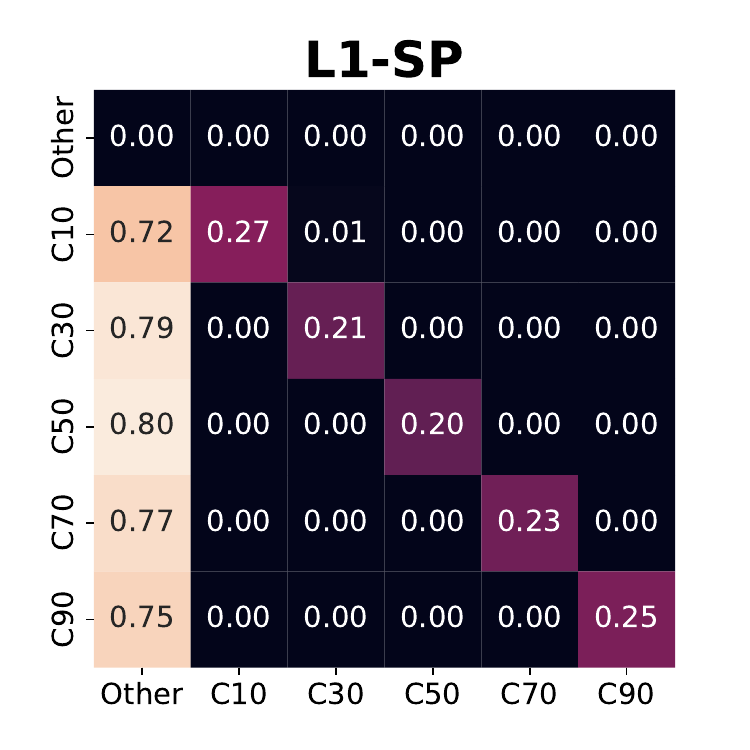}
		\includegraphics[width=0.24\linewidth]{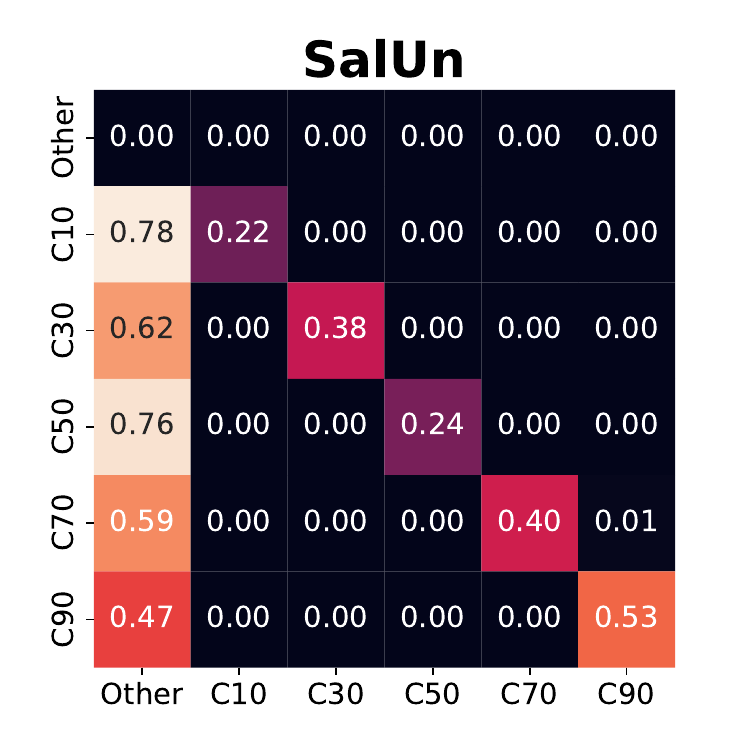}
		\includegraphics[width=0.24\linewidth]{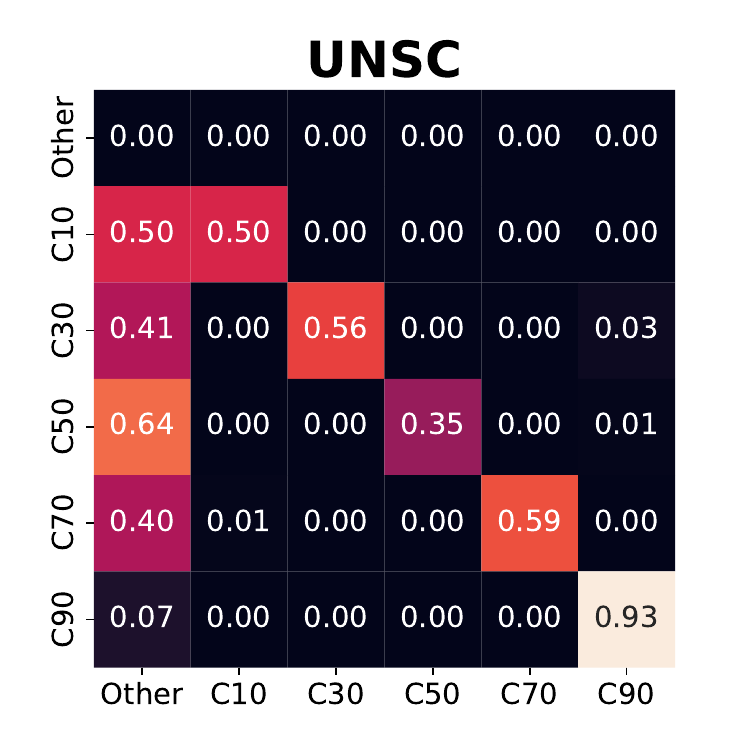}
		\caption{Confusion matrices of the \textbf{DST+STU+TCH} w.r.t. different MU methods}
	\end{minipage}
	\label{fig:open_source_pet37}
\end{figure*}

\subsection{Ablation Study on Pet-37}
\begin{table*}[ht!]
	\centering
	\resizebox{\linewidth}{!}
	{
		\begin{tabular}{c|c|c||c|c|c|c|c|c|c|c|c|c|c|c|c|c|c|c}
			\toprule
			\multicolumn{3}{c||}{\textbf{Component}} & \multicolumn{2}{c|}{\textbf{FF}} & \multicolumn{2}{c|}{\textbf{RL}} & \multicolumn{2}{c|}{\textbf{GA}} & \multicolumn{2}{c|}{\textbf{IU}} & \multicolumn{2}{c|}{\textbf{BU}} & \multicolumn{2}{c|}{\textbf{L1-SP}} & \multicolumn{2}{c|}{\textbf{SalUn}} & \multicolumn{2}{c}{\textbf{UNSC}} \\
			\midrule
			\textbf{DST} & \textbf{STU} & \textbf{TCH} & $\mathcal{D}_{ts}$ & $\mathcal{D}_f$  & $\mathcal{D}_{ts}$ & $\mathcal{D}_f$  & $\mathcal{D}_{ts}$ & $\mathcal{D}_f$  & $\mathcal{D}_{ts}$ & $\mathcal{D}_f$  & $\mathcal{D}_{ts}$ & $\mathcal{D}_f$  & $\mathcal{D}_{ts}$ & $\mathcal{D}_f$  & $\mathcal{D}_{ts}$ & $\mathcal{D}_f$  & $\mathcal{D}_{ts}$ & $\mathcal{D}_f$ \\
			\midrule
			\checkmark &       &       & \multicolumn{1}{c}{0.223 } & \multicolumn{1}{c}{0.168 } & \multicolumn{1}{c}{0.766 } & \multicolumn{1}{c}{0.364 } & \multicolumn{1}{c}{0.739 } & \multicolumn{1}{c}{0.176 } & \multicolumn{1}{c}{0.576 } & \multicolumn{1}{c}{0.144 } & \multicolumn{1}{c}{0.659 } & \multicolumn{1}{c}{0.128 } & \multicolumn{1}{c}{0.732 } & \multicolumn{1}{c}{0.208 } & \multicolumn{1}{c}{0.698 } & \multicolumn{1}{c}{0.108 } & \multicolumn{1}{c}{0.767 } & 0.320  \\
			\midrule
			\checkmark & \checkmark &       & 0.486  & 0.388  & 0.816  & 0.884  & 0.757  & 0.520  & 0.682  & 0.448  & 0.785  & 0.856  & 0.758  & 0.440  & 0.778  & 0.712  & 0.799  & 0.660  \\
			\midrule
			\checkmark & \checkmark & \checkmark & 0.782  & 0.904  & 0.850  & 0.960  & 0.843  & 0.952  & 0.838  & 0.936  & 0.846  & 0.960  & 0.841  & 0.920  & 0.849  & 0.948  & 0.856  & 0.976  \\
			\bottomrule
		\end{tabular}%
	}
	\caption{Ablation results (\textit{Acc}) of MRA on Pet-37 dataset w.r.t. different MU methods}
	\label{tab:addlabel}%
\end{table*}%

\begin{figure*}[ht!]
	\begin{minipage}[c]{\linewidth}
		\flushleft
		\includegraphics[width=0.24\linewidth]{figures/results/pet-37/fr_0.5_ijcai/fisher_resnet18_ul_cmt.pdf}
		\includegraphics[width=0.24\linewidth]{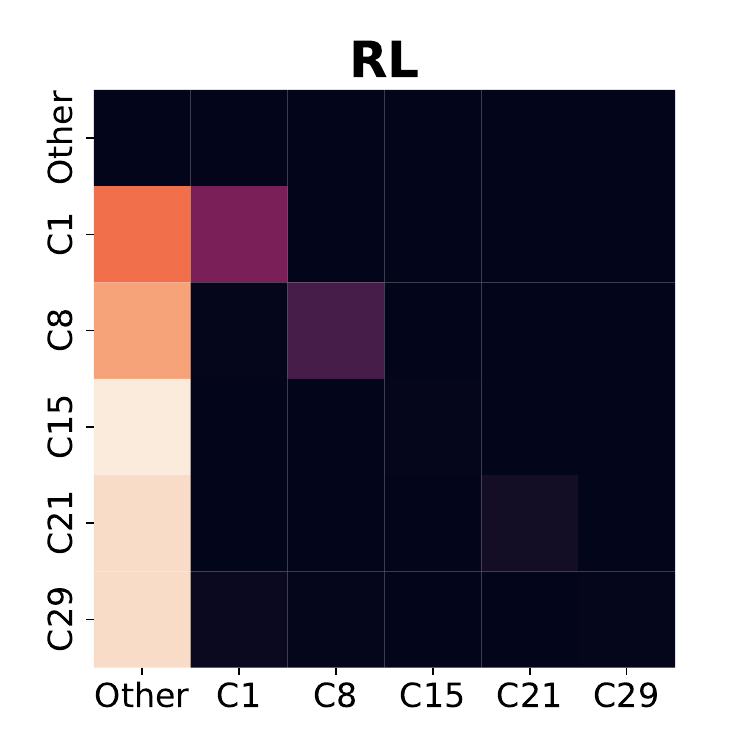}
		\includegraphics[width=0.24\linewidth]{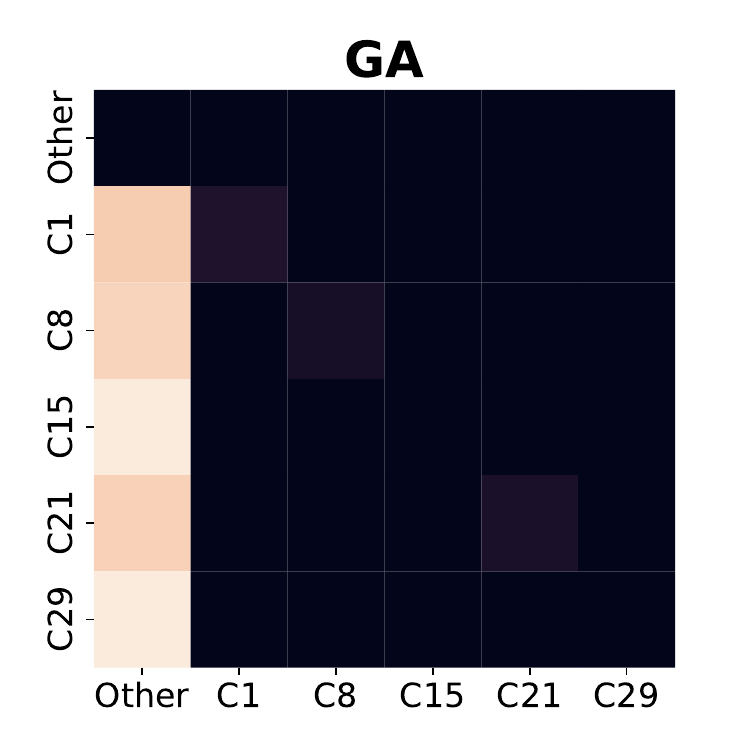}
		\includegraphics[width=0.24\linewidth]{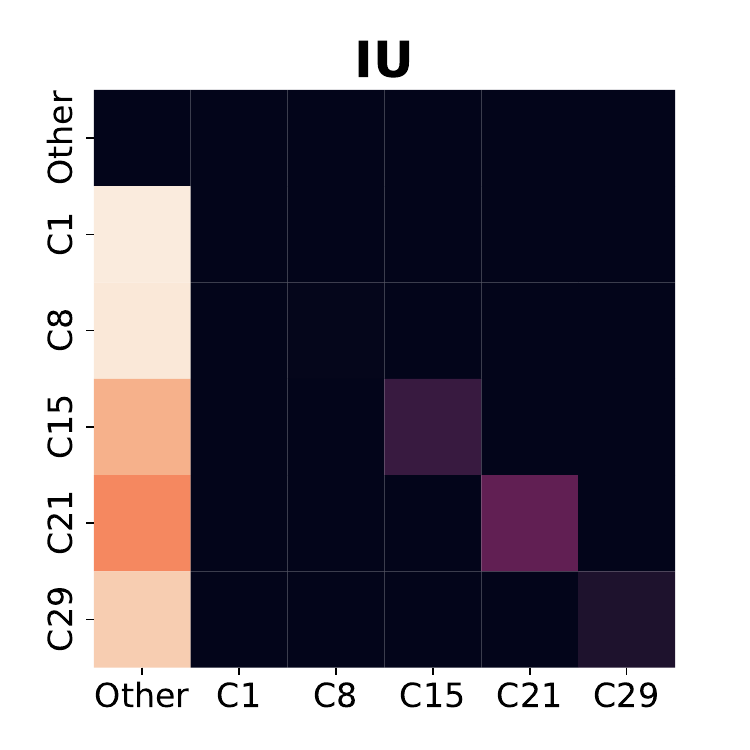}
		\includegraphics[width=0.24\linewidth]{figures/results/pet-37/fr_0.5_ijcai/BU_resnet18_ul_cmt.pdf}
		\includegraphics[width=0.24\linewidth]{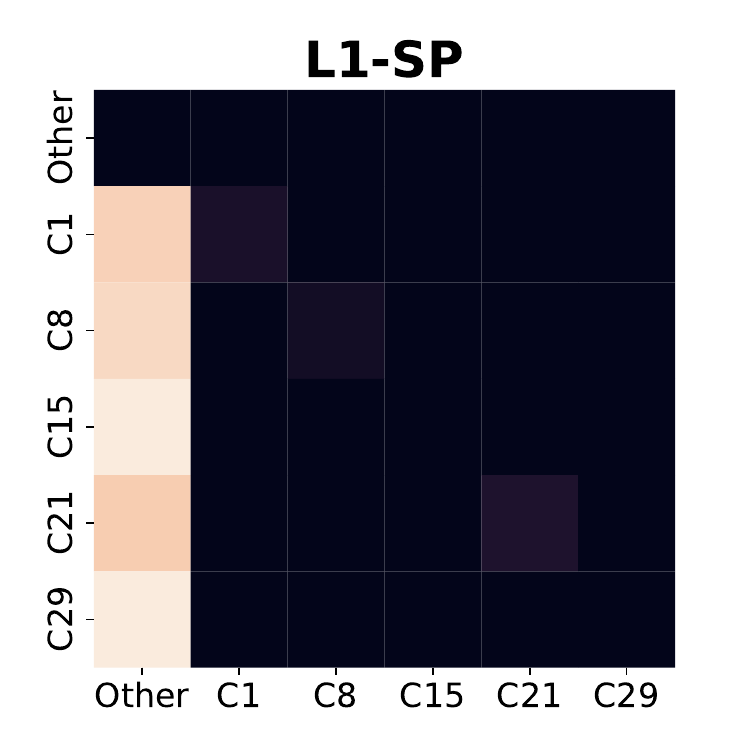}
		\includegraphics[width=0.24\linewidth]{figures/results/pet-37/fr_0.5_ijcai/SalUn_resnet18_ul_cmt.pdf}
		\includegraphics[width=0.24\linewidth]{figures/results/pet-37/fr_0.5_ijcai/UNSC_resnet18_ul_cmt.pdf}
		\caption{Confusion matrices of the \textbf{UML} w.r.t. different MU methods}
	\end{minipage}
\end{figure*}

\begin{figure*}[ht!]
	\begin{minipage}[c]{\linewidth}
		\flushleft
		\includegraphics[width=0.24\linewidth]{figures/results/pet-37/fr_0.5_ijcai/fisher_resnet18_distill_cmt.pdf}
		\includegraphics[width=0.24\linewidth]{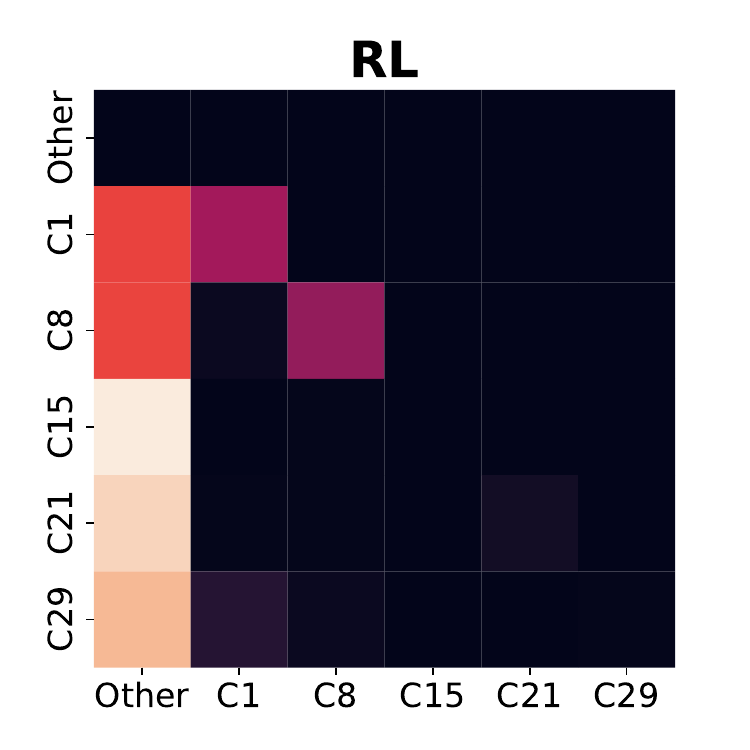}
		\includegraphics[width=0.24\linewidth]{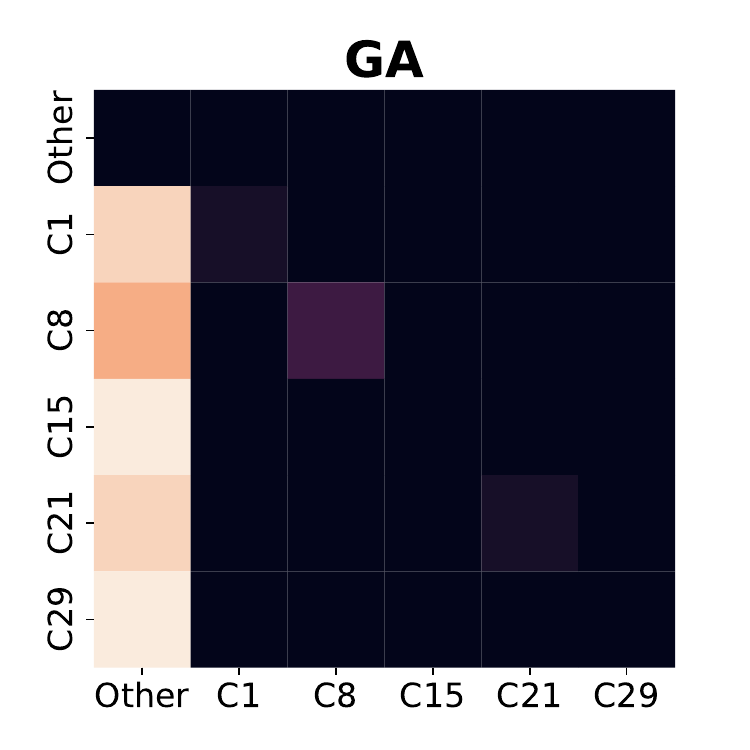}
		\includegraphics[width=0.24\linewidth]{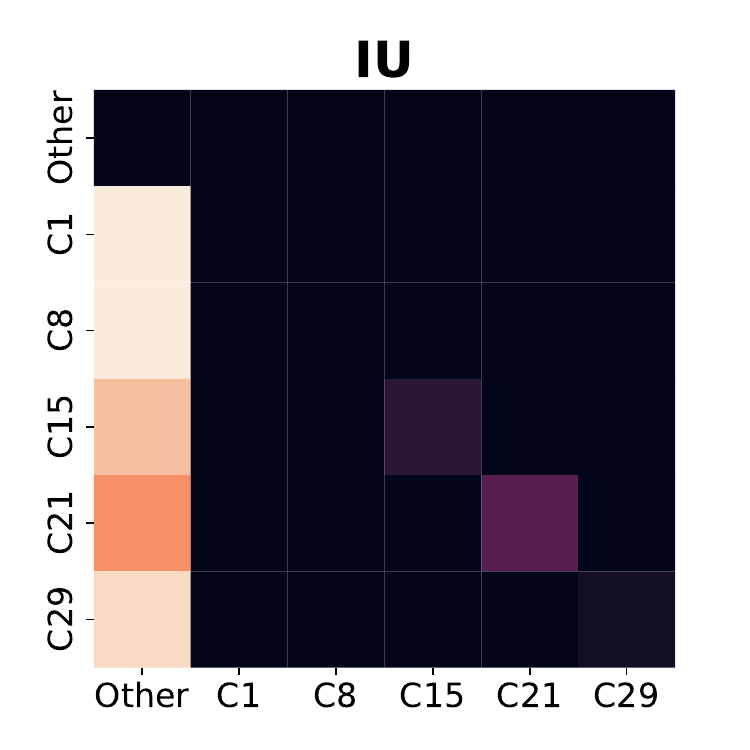}
		\includegraphics[width=0.24\linewidth]{figures/results/pet-37/fr_0.5_ijcai/BU_resnet18_distill_cmt.pdf}
		\includegraphics[width=0.24\linewidth]{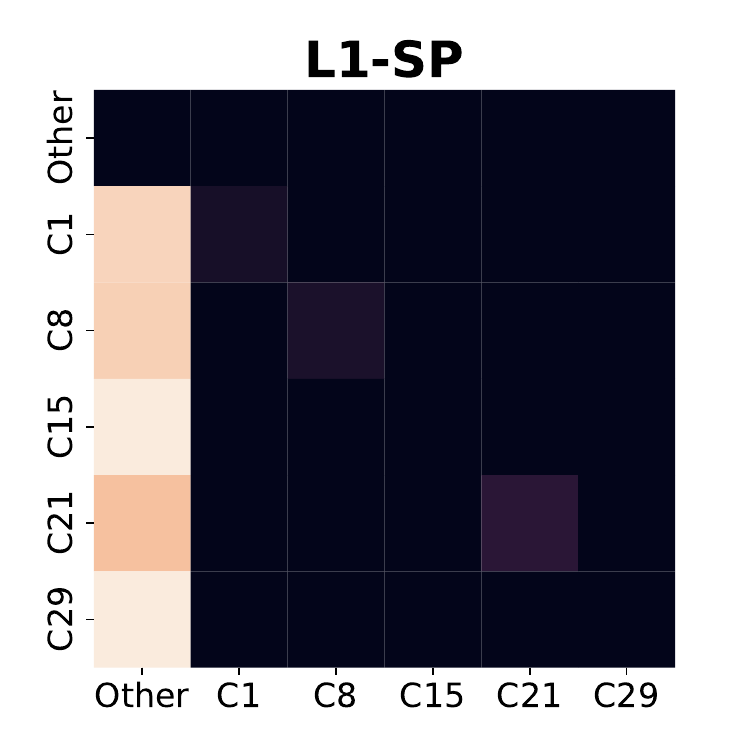}
		\includegraphics[width=0.24\linewidth]{figures/results/pet-37/fr_0.5_ijcai/SalUn_resnet18_distill_cmt.pdf}
		\includegraphics[width=0.24\linewidth]{figures/results/pet-37/fr_0.5_ijcai/UNSC_resnet18_distill_cmt.pdf}
		\caption{Confusion matrices of the \textbf{DST} w.r.t. different MU methods}
	\end{minipage}
\end{figure*}

\begin{figure*}[ht!]
	\begin{minipage}[c]{\linewidth}
		\flushleft
		\includegraphics[width=0.24\linewidth]{figures/results/pet-37/fr_0.5_ijcai/fisher_resnet18_student_only_cmt.pdf}
		\includegraphics[width=0.24\linewidth]{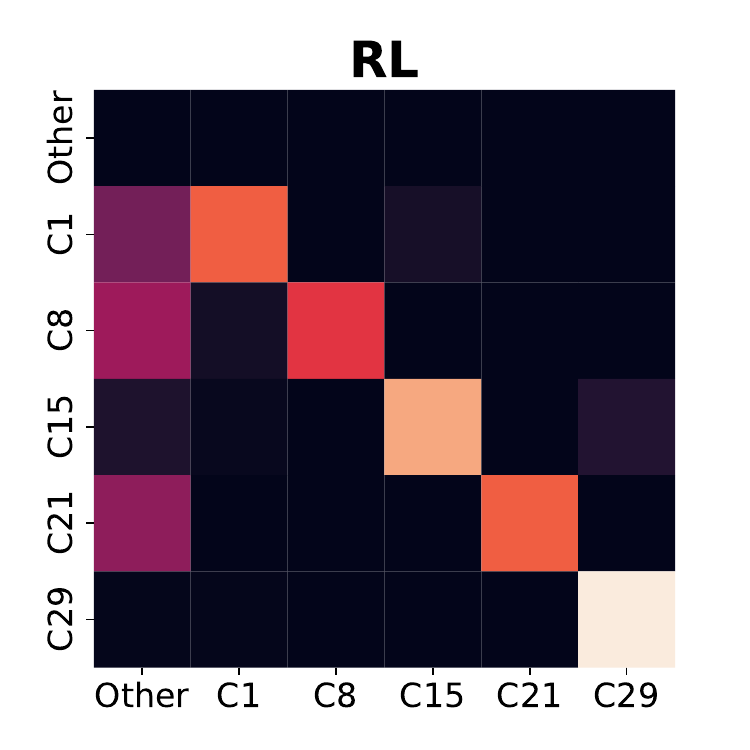}
		\includegraphics[width=0.24\linewidth]{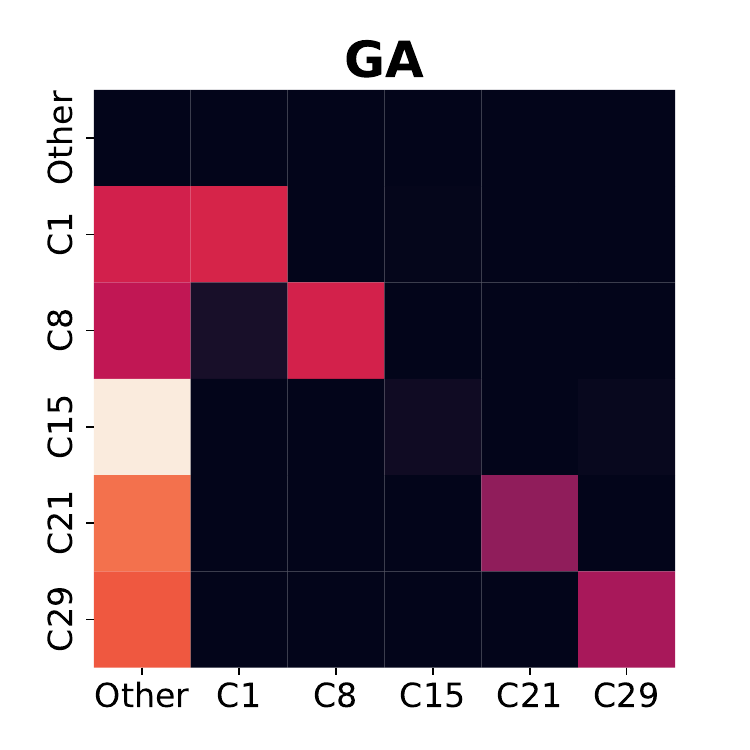}
		\includegraphics[width=0.24\linewidth]{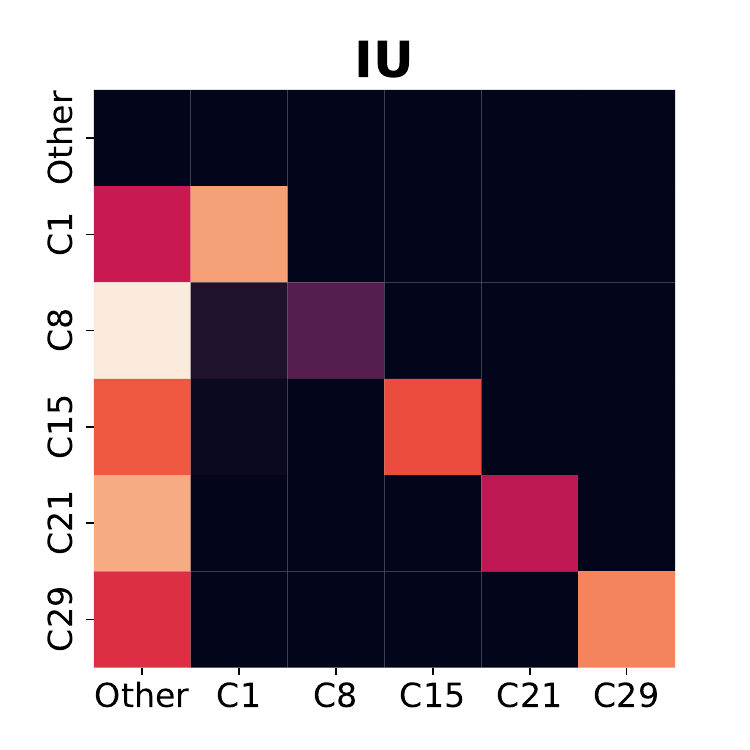}
		\includegraphics[width=0.24\linewidth]{figures/results/pet-37/fr_0.5_ijcai/BU_resnet18_student_only_cmt.pdf}
		\includegraphics[width=0.24\linewidth]{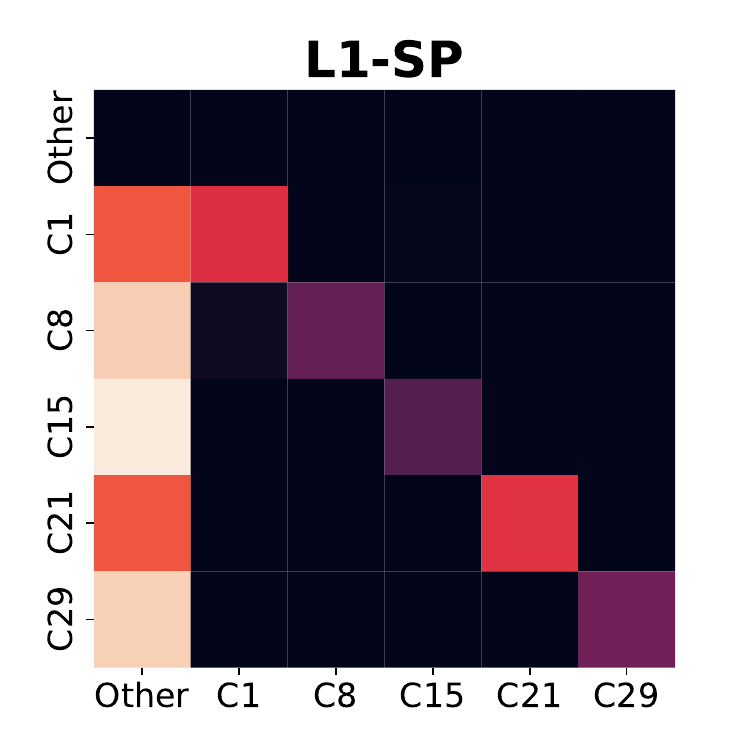}
		\includegraphics[width=0.24\linewidth]{figures/results/pet-37/fr_0.5_ijcai/SalUn_resnet18_student_only_cmt.pdf}
		\includegraphics[width=0.24\linewidth]{figures/results/pet-37/fr_0.5_ijcai/UNSC_resnet18_student_only_cmt.pdf}
		\caption{Confusion matrices of the \textbf{DST+STU} w.r.t. different MU methods}
	\end{minipage}
\end{figure*}

\begin{figure*}[ht!]
	\begin{minipage}[c]{\linewidth}
		\flushleft
		\includegraphics[width=0.24\linewidth]{figures/results/pet-37/fr_0.5_ijcai/fisher_resnet18_restore_cmt.pdf}
		\includegraphics[width=0.24\linewidth]{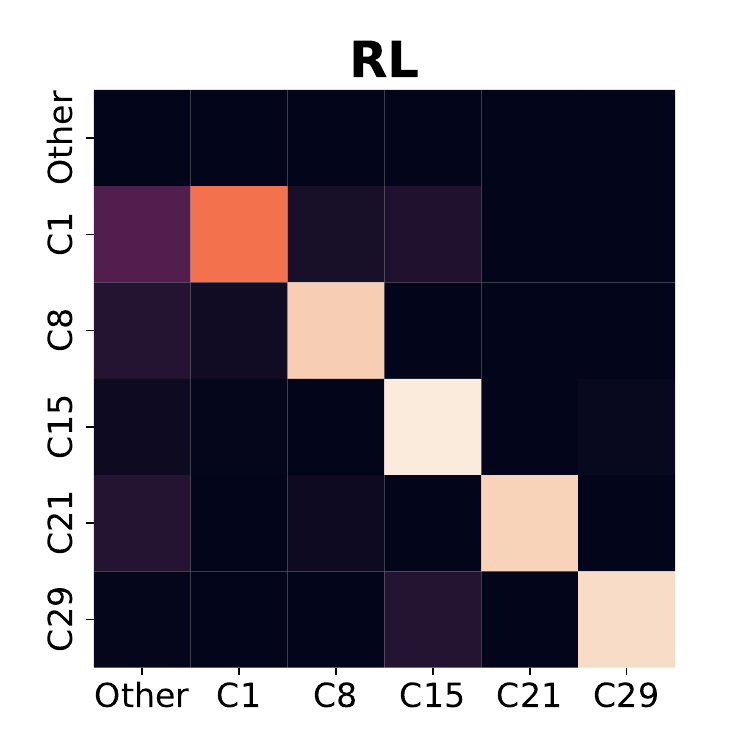}
		\includegraphics[width=0.24\linewidth]{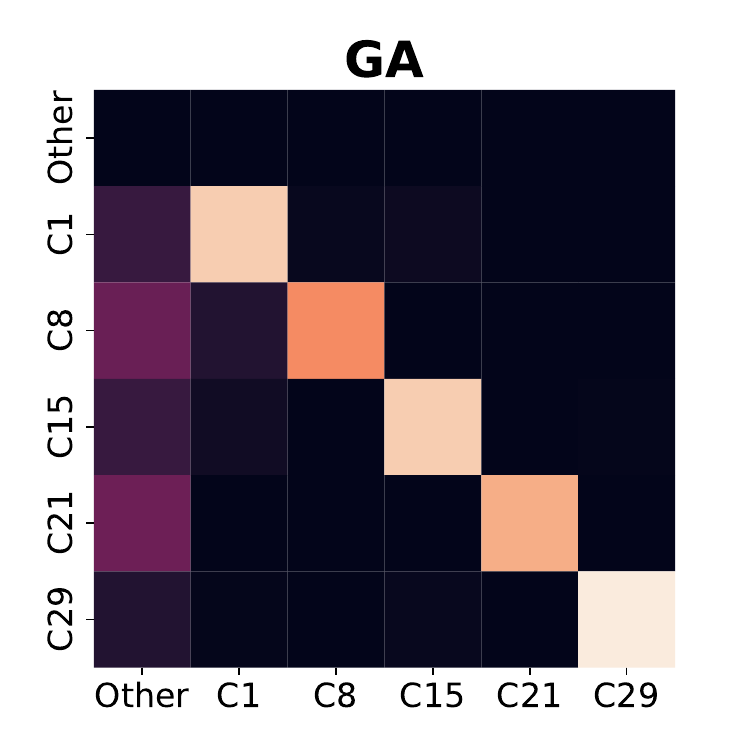}
		\includegraphics[width=0.24\linewidth]{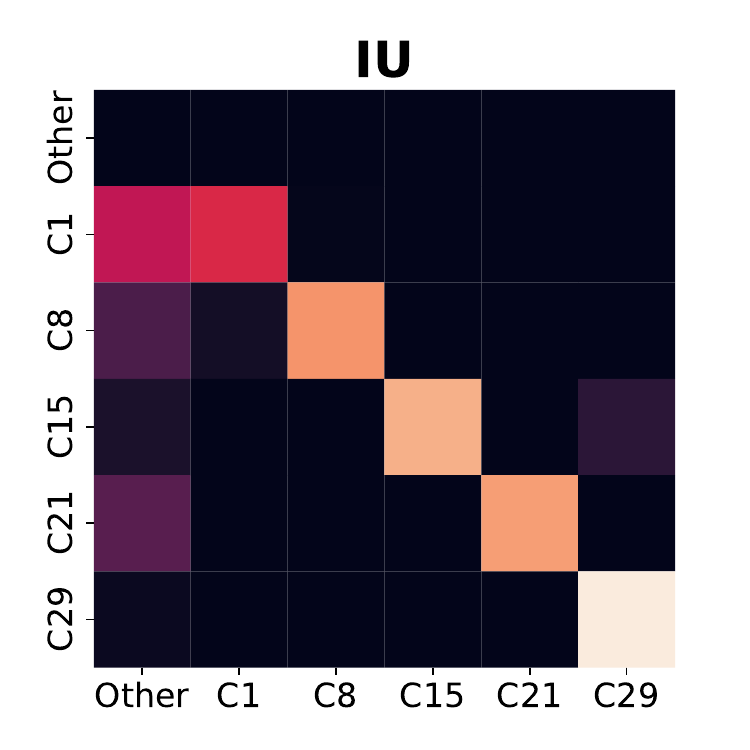}
		\includegraphics[width=0.24\linewidth]{figures/results/pet-37/fr_0.5_ijcai/BU_resnet18_restore_cmt.pdf}
		\includegraphics[width=0.24\linewidth]{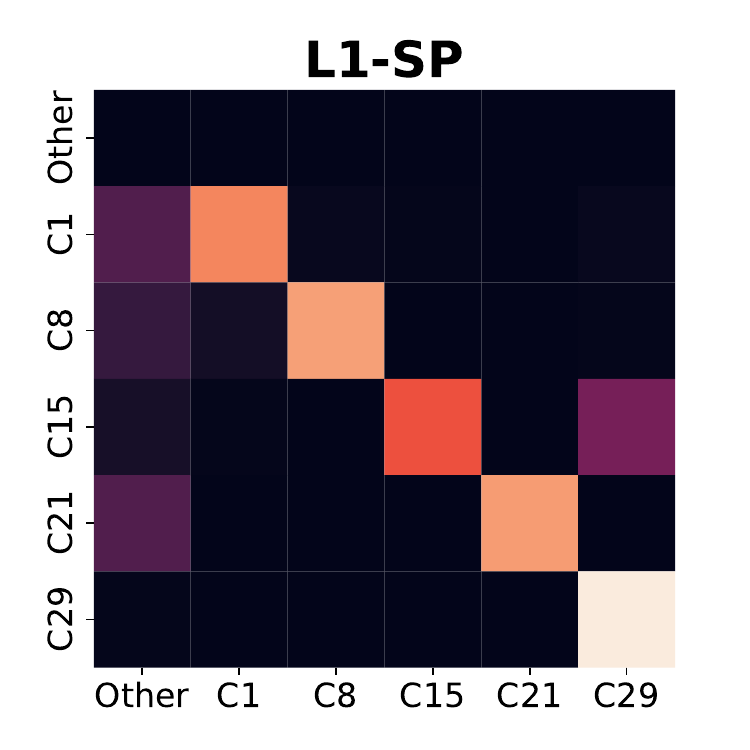}
		\includegraphics[width=0.24\linewidth]{figures/results/pet-37/fr_0.5_ijcai/SalUn_resnet18_restore_cmt.pdf}
		\includegraphics[width=0.24\linewidth]{figures/results/pet-37/fr_0.5_ijcai/UNSC_resnet18_restore_cmt.pdf}
		\caption{Confusion matrices of the \textbf{DST+STU+TCH} w.r.t. different MU methods}
	\end{minipage}
	\label{fig:open_source_pet37}
\end{figure*}

\subsection{Ablation Study on FLower-102}
\begin{table*}[ht!]
	\centering
	\resizebox{\linewidth}{!}
	{
		\begin{tabular}{c|c|c||c|c|c|c|c|c|c|c|c|c|c|c|c|c|c|c}
			\toprule
			\multicolumn{3}{c||}{\textbf{Component}} & \multicolumn{2}{c|}{\textbf{FF}} & \multicolumn{2}{c|}{\textbf{RL}} & \multicolumn{2}{c|}{\textbf{GA}} & \multicolumn{2}{c|}{\textbf{IU}} & \multicolumn{2}{c|}{\textbf{BU}} & \multicolumn{2}{c|}{\textbf{L1-SP}} & \multicolumn{2}{c|}{\textbf{SalUn}} & \multicolumn{2}{c}{\textbf{UNSC}} \\
			\midrule
			\textbf{DST} & \textbf{STU} & \textbf{TCH} & $\mathcal{D}_{ts}$ & $\mathcal{D}_f$  & $\mathcal{D}_{ts}$ & $\mathcal{D}_f$  & $\mathcal{D}_{ts}$ & $\mathcal{D}_f$  & $\mathcal{D}_{ts}$ & $\mathcal{D}_f$  & $\mathcal{D}_{ts}$ & $\mathcal{D}_f$  & $\mathcal{D}_{ts}$ & $\mathcal{D}_f$  & $\mathcal{D}_{ts}$ & $\mathcal{D}_f$  & $\mathcal{D}_{ts}$ & $\mathcal{D}_f$ \\
			\midrule
			\checkmark &       &       & \multicolumn{1}{c}{0.278 } & \multicolumn{1}{c}{0.198 } & \multicolumn{1}{c}{0.741 } & \multicolumn{1}{c}{0.170 } & \multicolumn{1}{c}{0.319 } & \multicolumn{1}{c}{0.231 } & \multicolumn{1}{c}{0.519 } & \multicolumn{1}{c}{0.320 } & \multicolumn{1}{c}{0.714 } & \multicolumn{1}{c}{0.377 } & \multicolumn{1}{c}{0.511 } & \multicolumn{1}{c}{0.308 } & \multicolumn{1}{c}{0.505 } & 0.247  & NA    & NA \\
			\midrule
			\checkmark & \checkmark &       & 0.376  & 0.312  & 0.831  & 0.567  & 0.510  & 0.352  & 0.589  & 0.364  & 0.758  & 0.478  & 0.655  & 0.538  & 0.599  & 0.429  & NA    & NA \\
			\midrule
			\checkmark & \checkmark & \checkmark & 0.607  & 0.482  & 0.915  & 0.988  & 0.772  & 0.709  & 0.829  & 0.725  & 0.889  & 0.972  & 0.856  & 0.935  & 0.857  & 0.915  & NA    & NA \\
			\bottomrule
		\end{tabular}%
	}
	\caption{Ablation results (\textit{Acc}) of MRA on FLower-102 dataset w.r.t. different MU methods}
	\label{tab:addlabel}%
\end{table*}%

\begin{figure*}[ht!]
	\begin{minipage}[c]{\linewidth}
		\flushleft
		\includegraphics[width=0.24\linewidth]{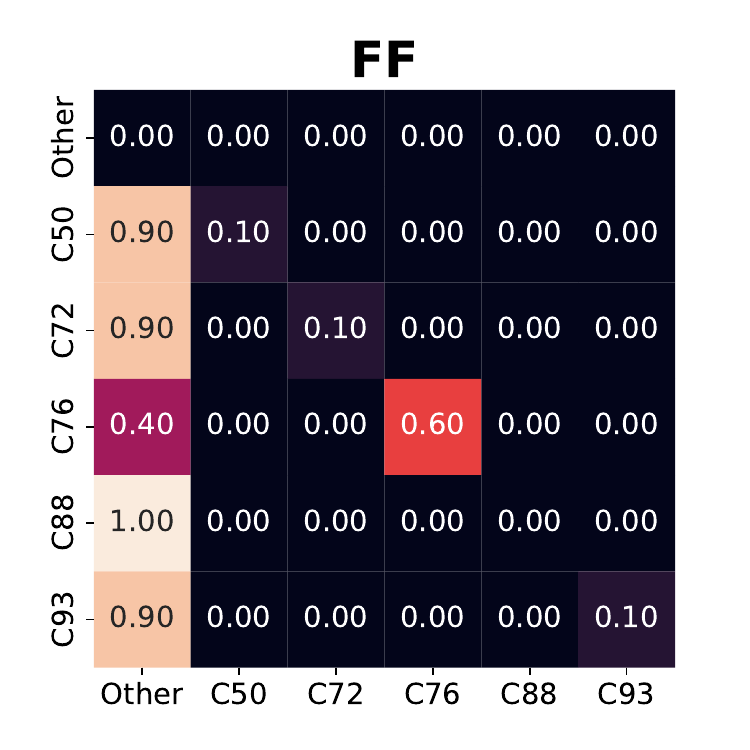}
		\includegraphics[width=0.24\linewidth]{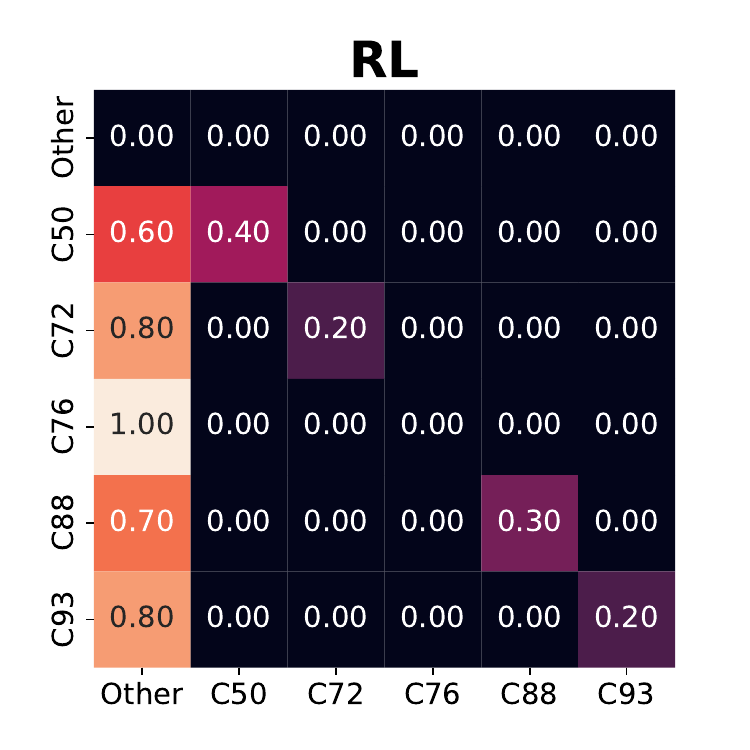}
		\includegraphics[width=0.24\linewidth]{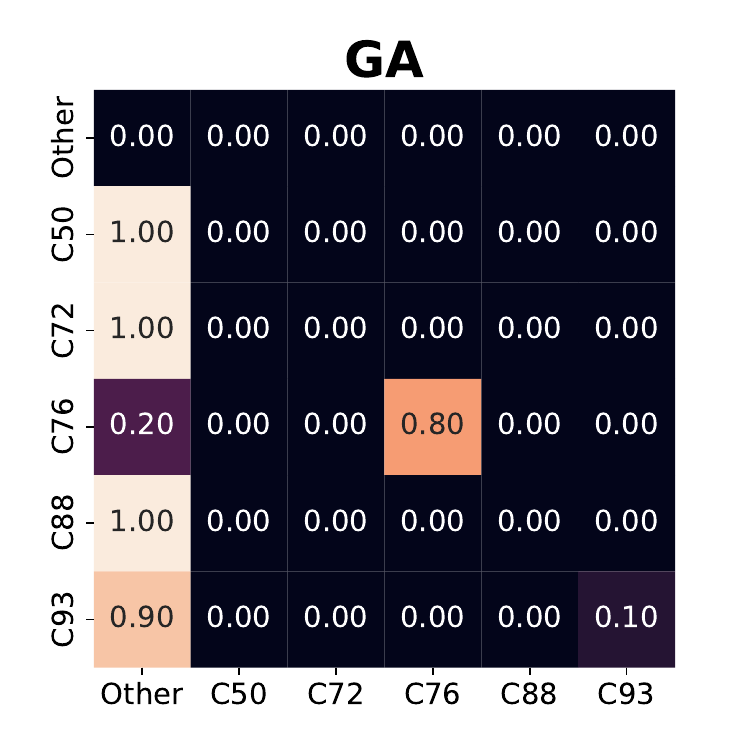}
		\includegraphics[width=0.24\linewidth]{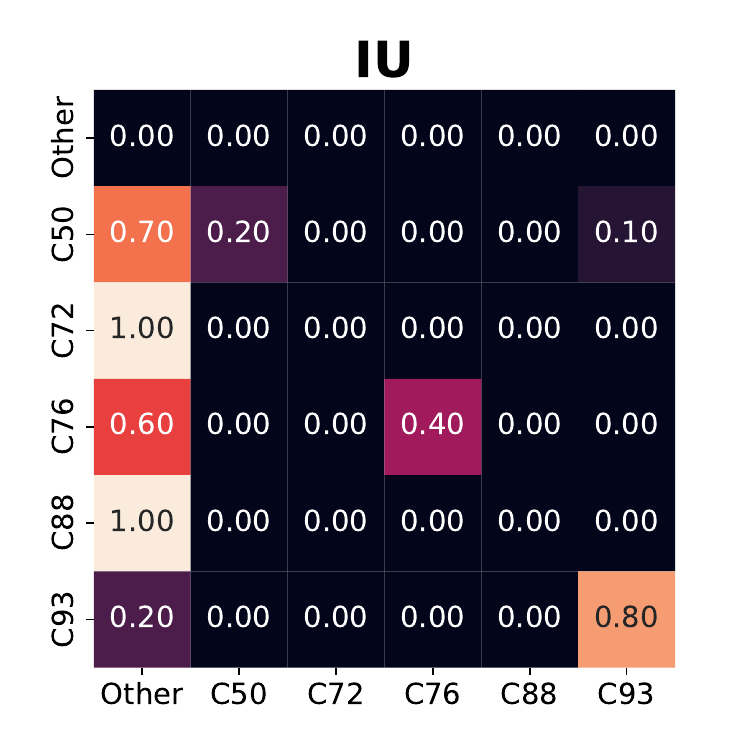}
		\includegraphics[width=0.24\linewidth]{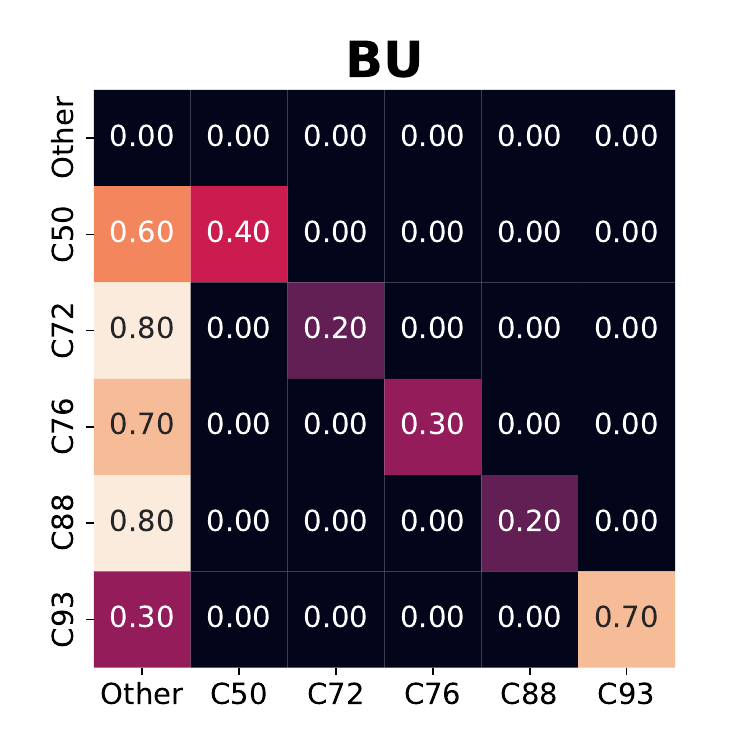}
		\includegraphics[width=0.24\linewidth]{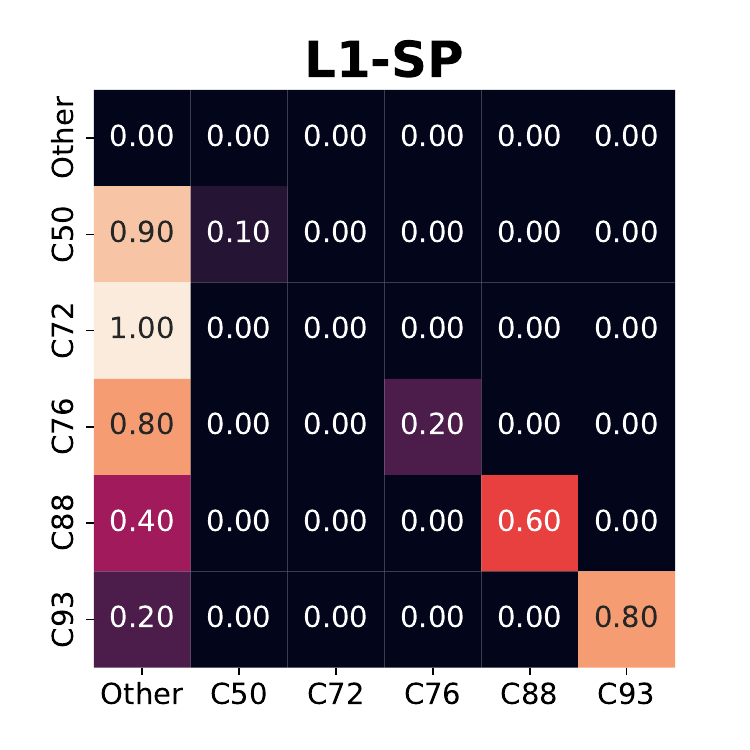}
		\includegraphics[width=0.24\linewidth]{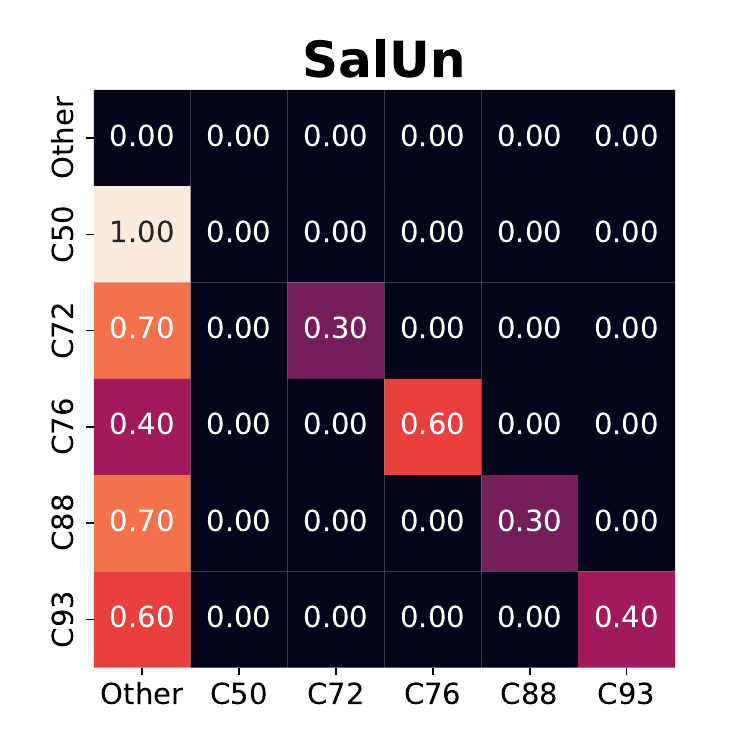}
		\caption{Confusion matrices of the \textbf{ULM} w.r.t. different MU methods}
	\end{minipage}
\end{figure*}

\begin{figure*}[ht!]
	\begin{minipage}[c]{\linewidth}
		\flushleft
		\includegraphics[width=0.24\linewidth]{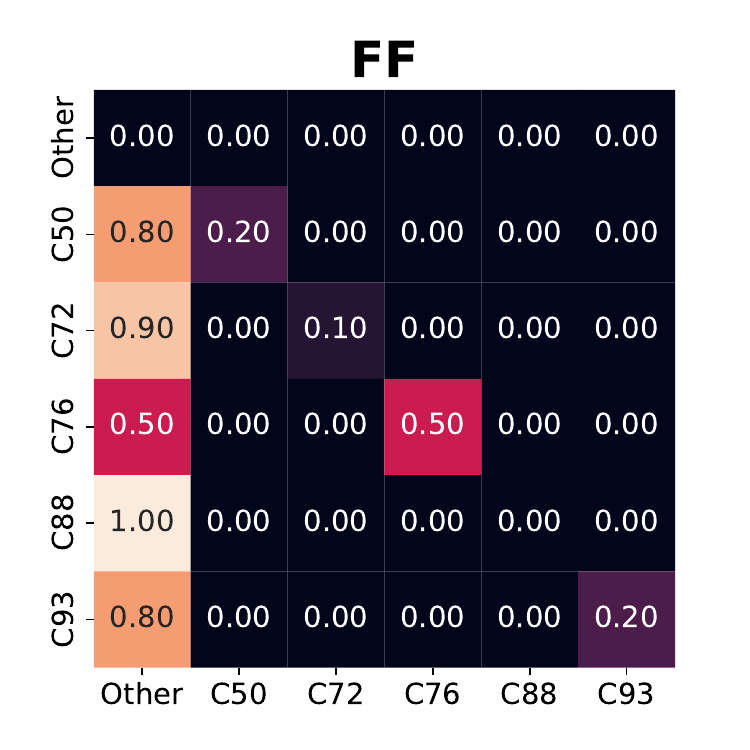}
		\includegraphics[width=0.24\linewidth]{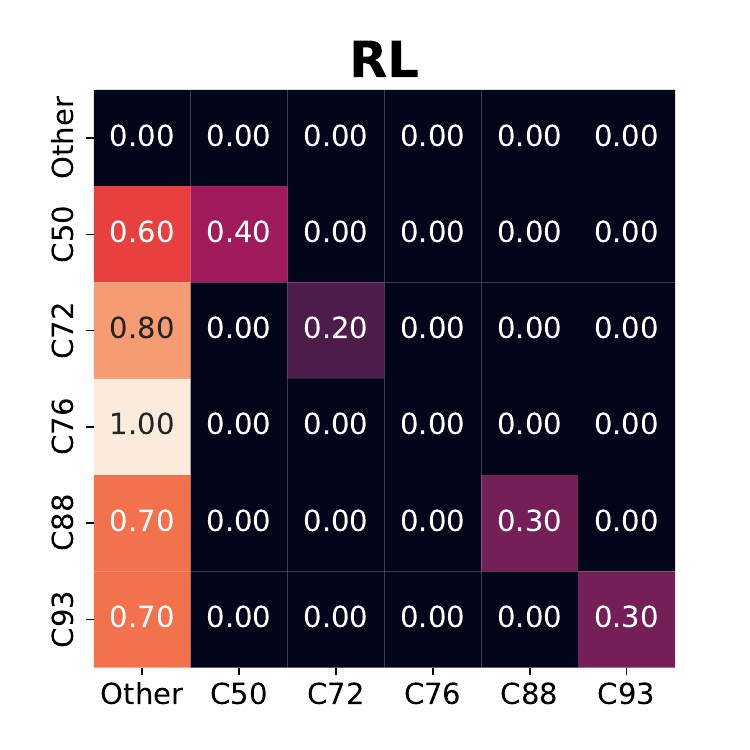}
		\includegraphics[width=0.24\linewidth]{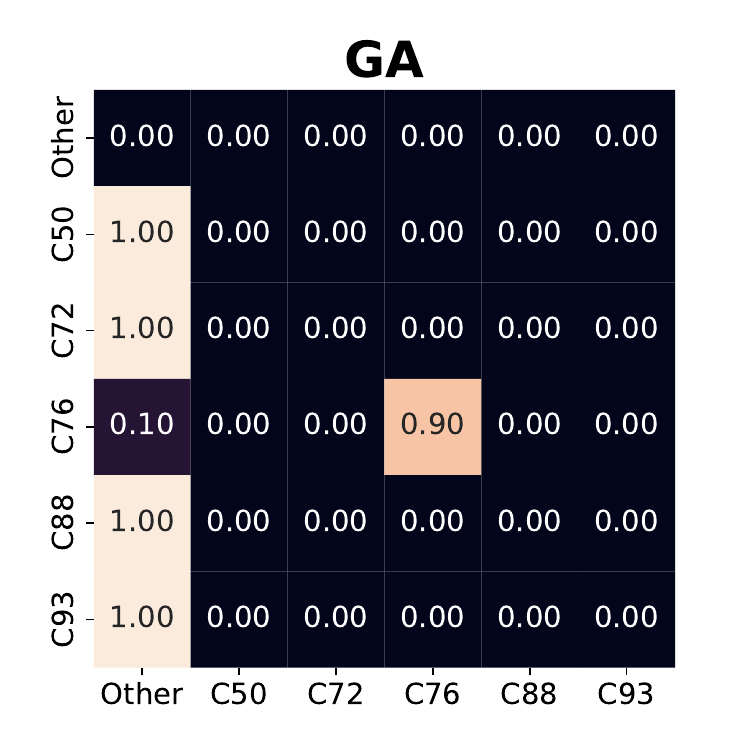}
		\includegraphics[width=0.24\linewidth]{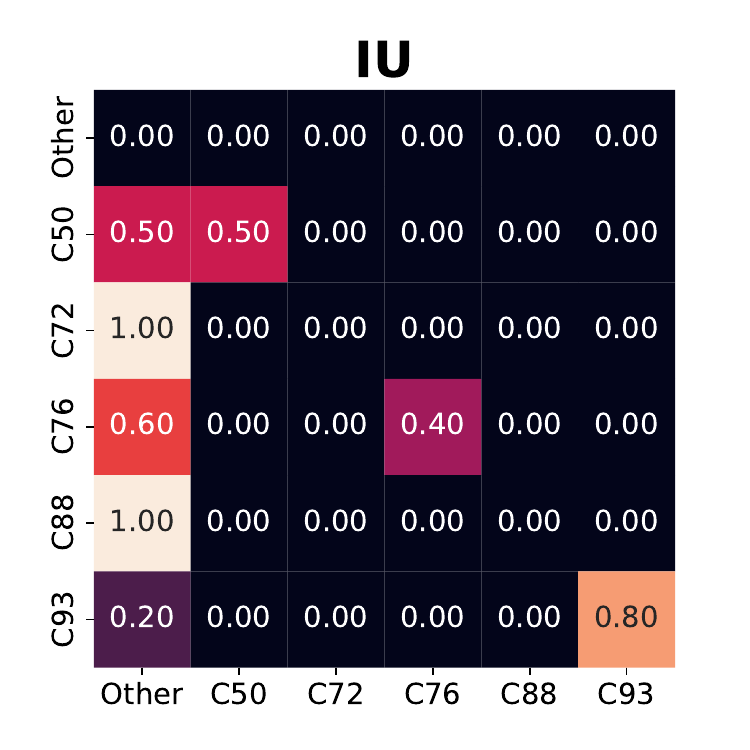}
		\includegraphics[width=0.24\linewidth]{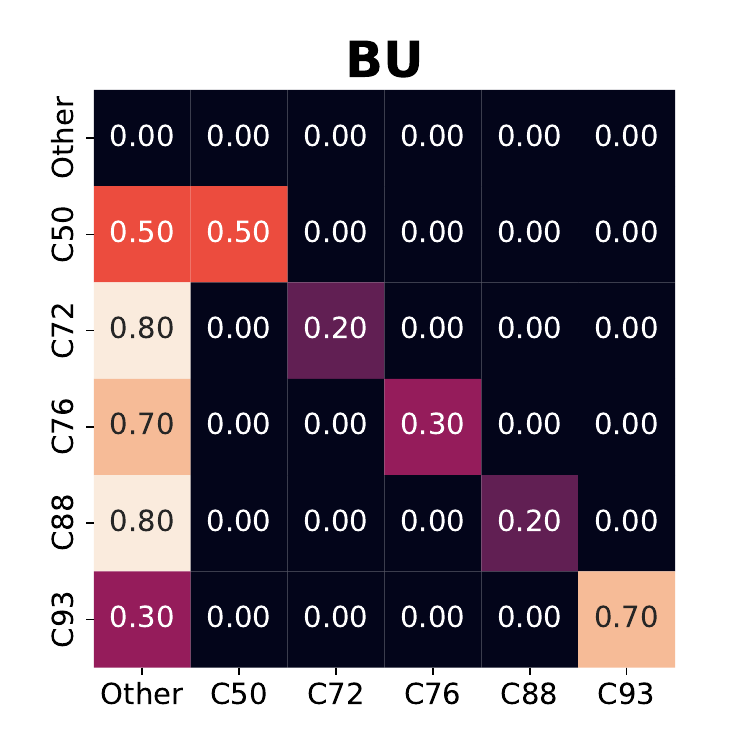}
		\includegraphics[width=0.24\linewidth]{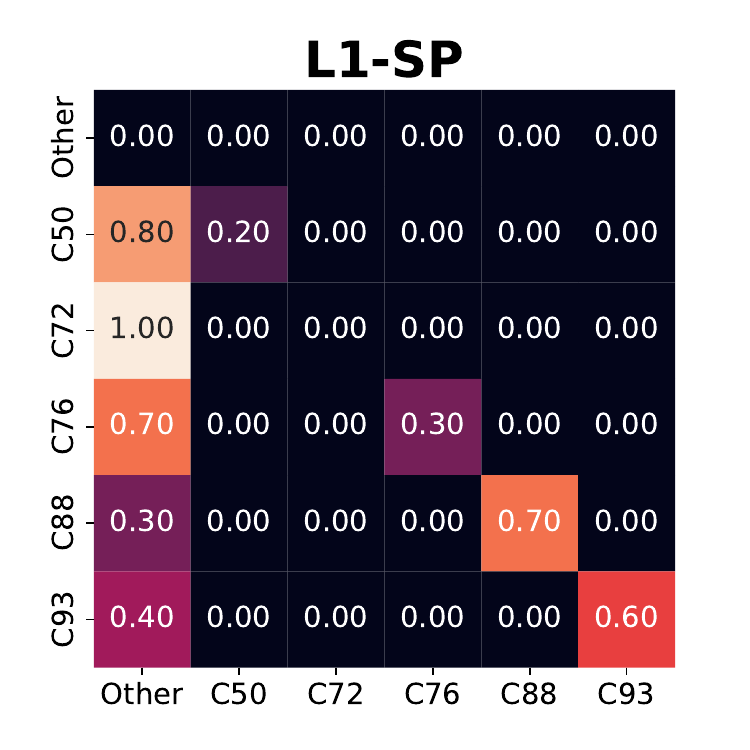}
		\includegraphics[width=0.24\linewidth]{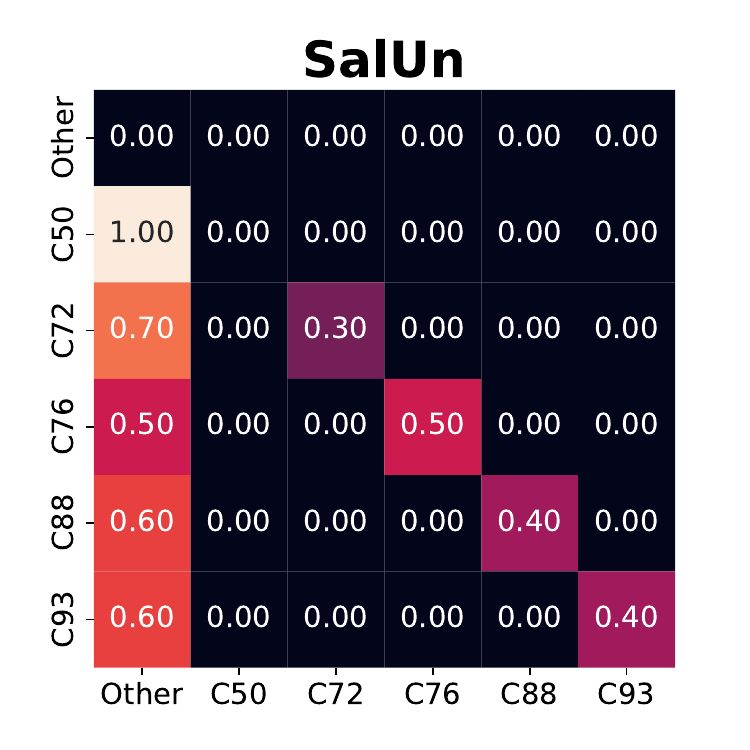}
		\caption{Confusion matrices of the \textbf{DST} w.r.t. different MU methods}
	\end{minipage}
\end{figure*}

\begin{figure*}[ht!]
	\begin{minipage}[c]{\linewidth}
		\flushleft
		\includegraphics[width=0.24\linewidth]{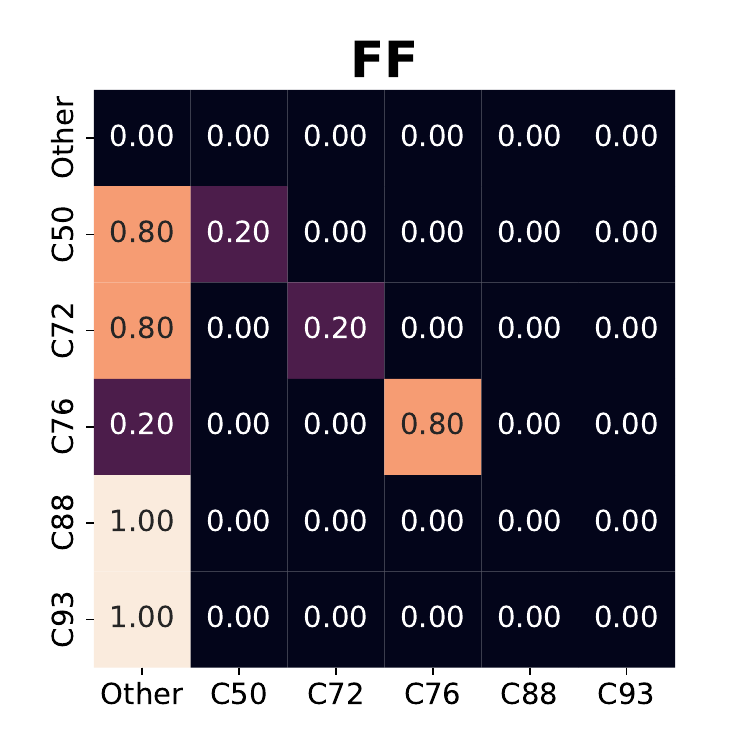}
		\includegraphics[width=0.24\linewidth]{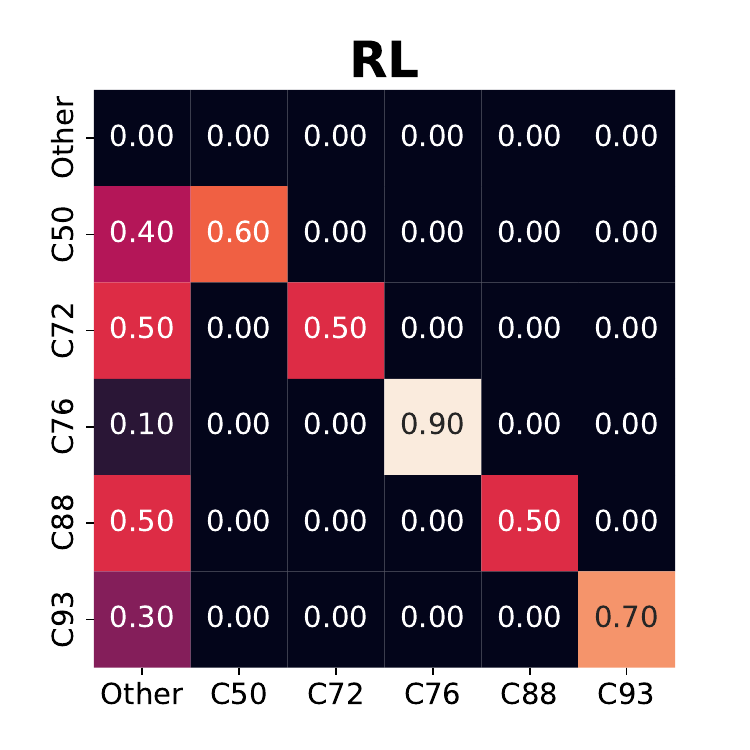}
		\includegraphics[width=0.24\linewidth]{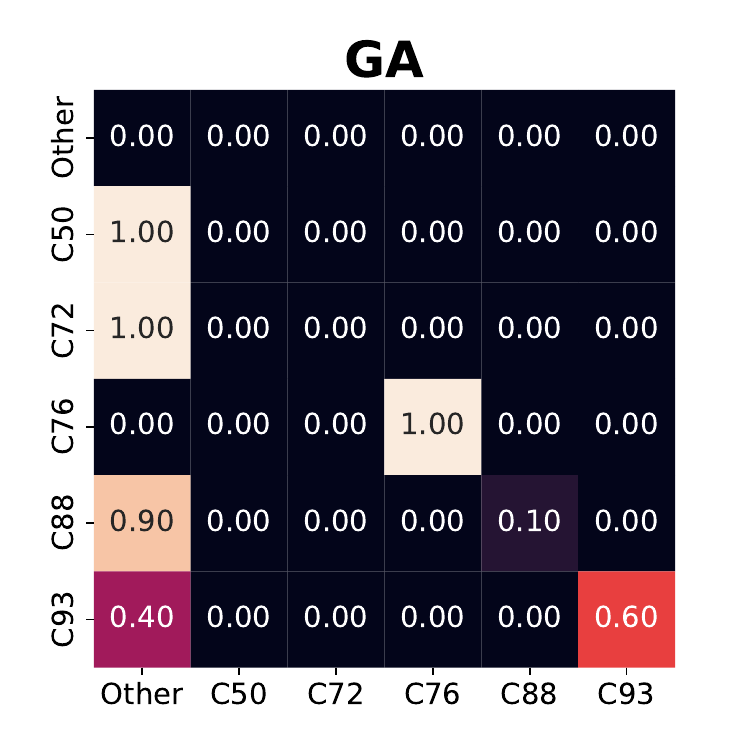}
		\includegraphics[width=0.24\linewidth]{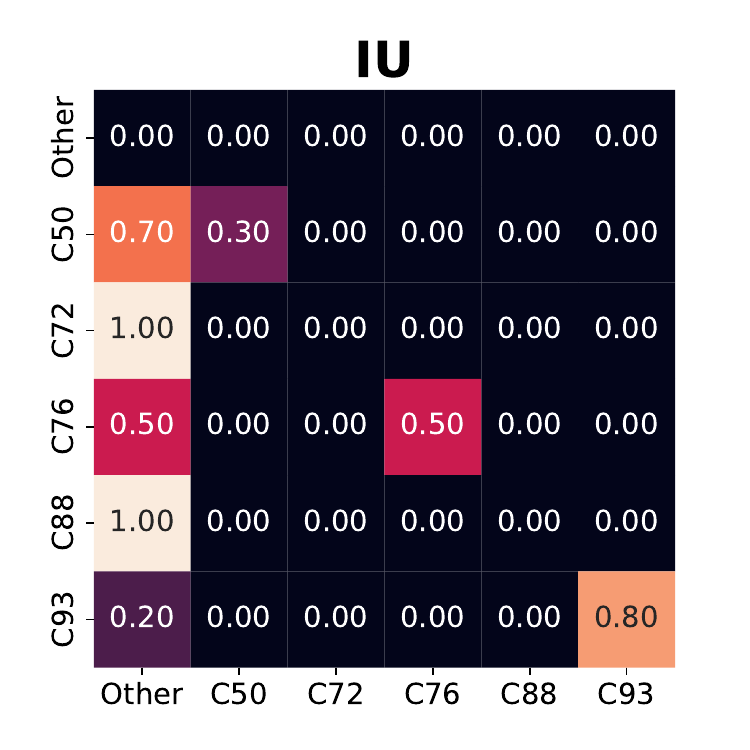}
		\includegraphics[width=0.24\linewidth]{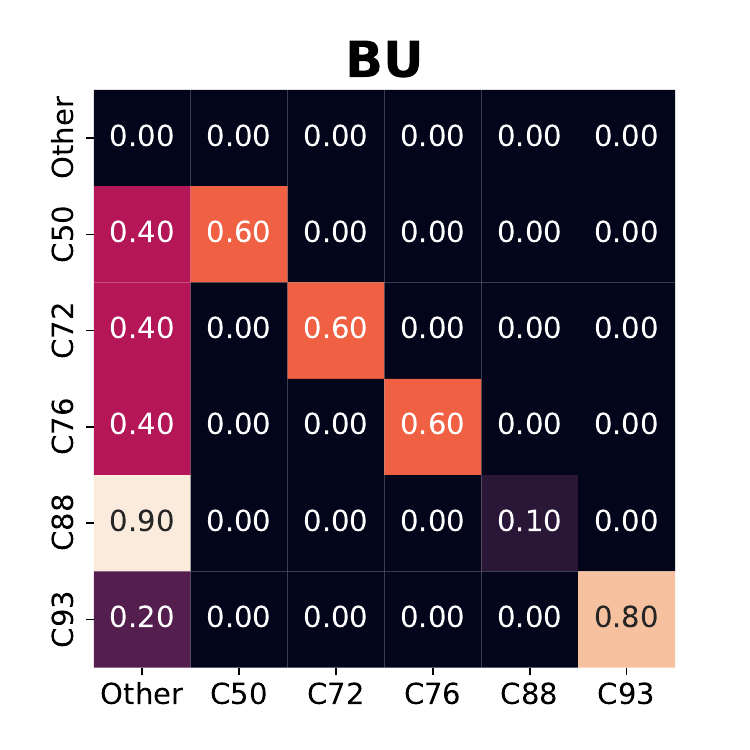}
		\includegraphics[width=0.24\linewidth]{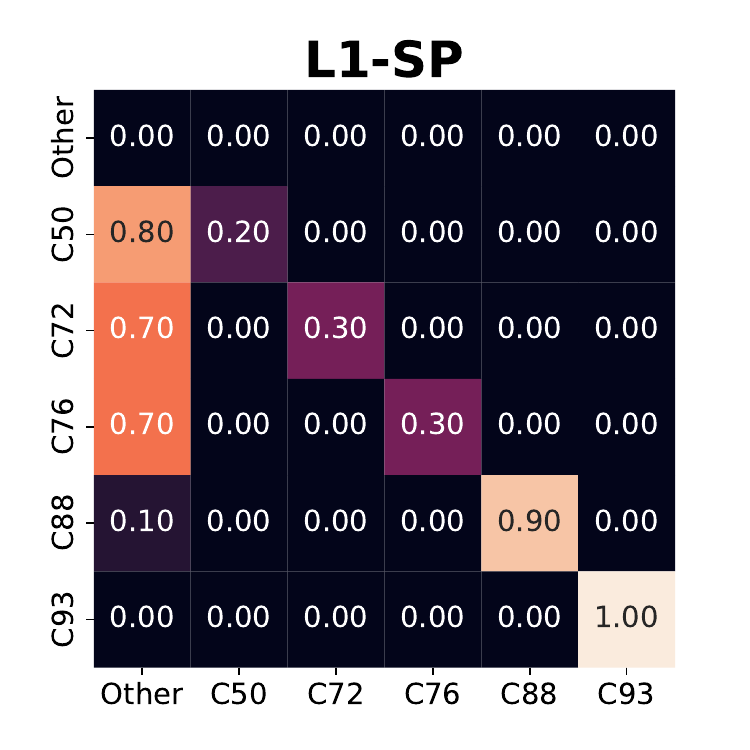}
		\includegraphics[width=0.24\linewidth]{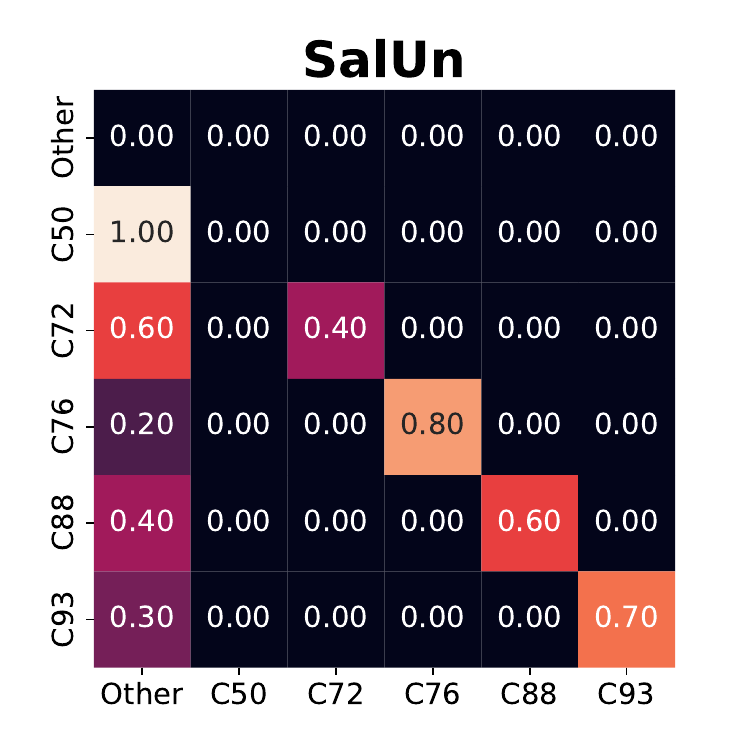}
		\caption{Confusion matrices of the \textbf{DST+STU} w.r.t. different MU methods}
	\end{minipage}
\end{figure*}

\begin{figure*}[ht!]
	\begin{minipage}[c]{\linewidth}
		\flushleft
		\includegraphics[width=0.24\linewidth]{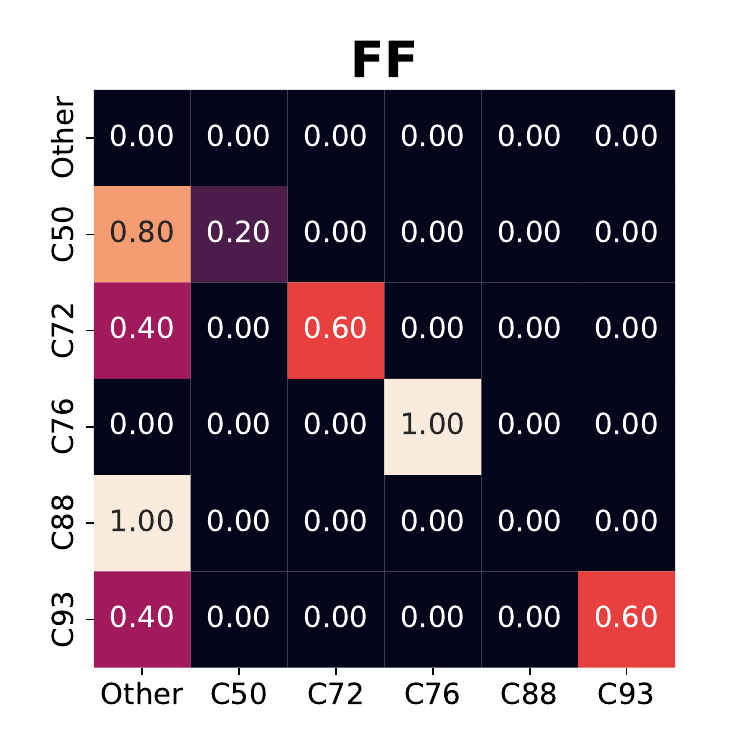}
		\includegraphics[width=0.24\linewidth]{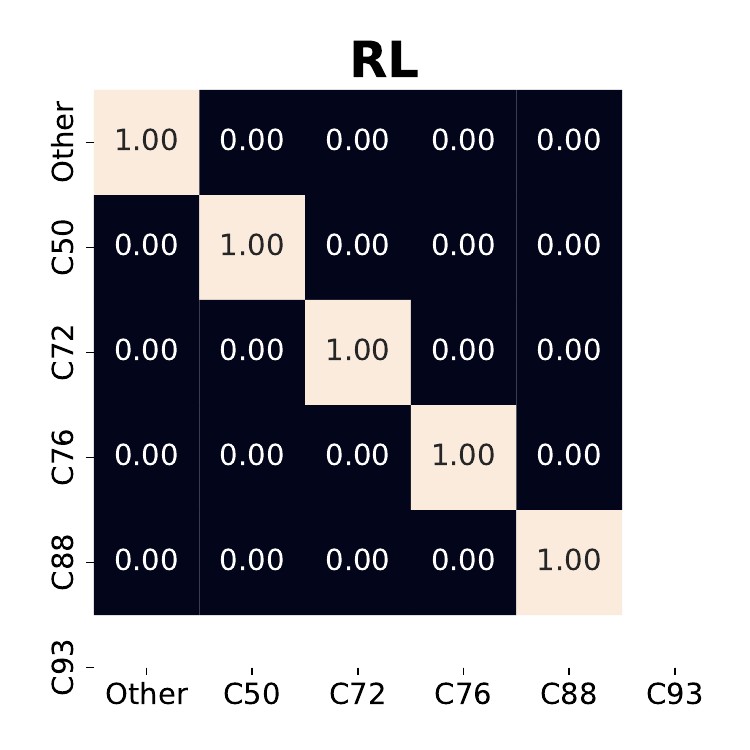}
		\includegraphics[width=0.24\linewidth]{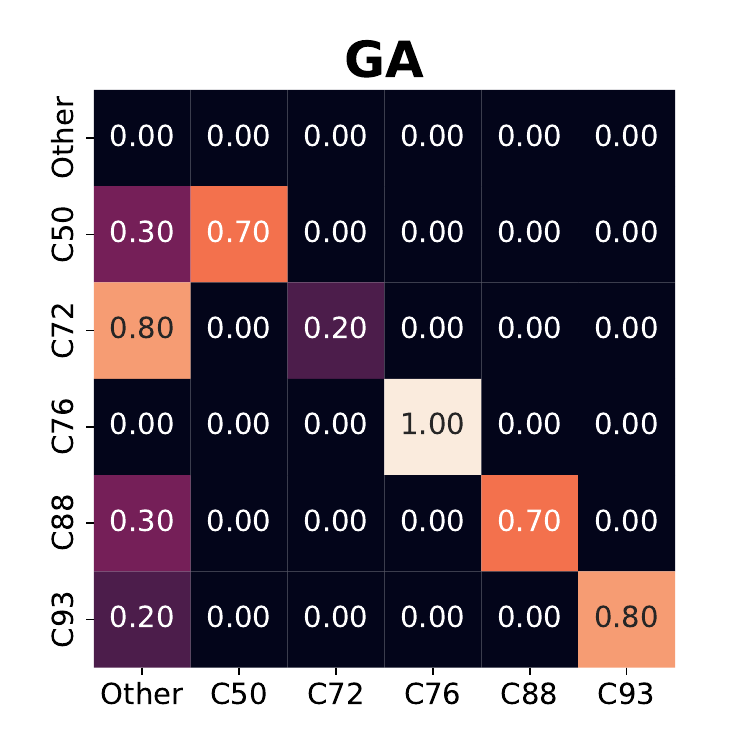}
		\includegraphics[width=0.24\linewidth]{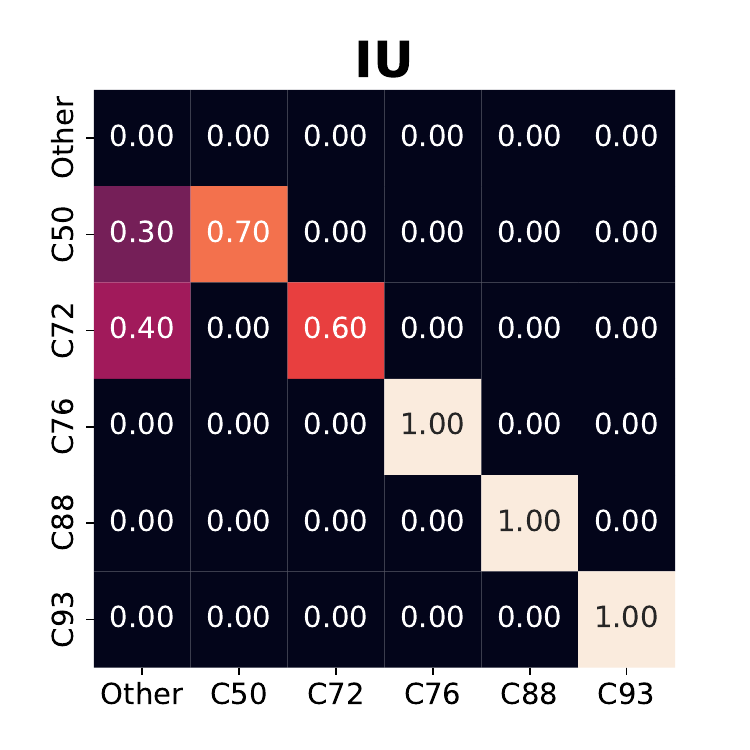}
		\includegraphics[width=0.24\linewidth]{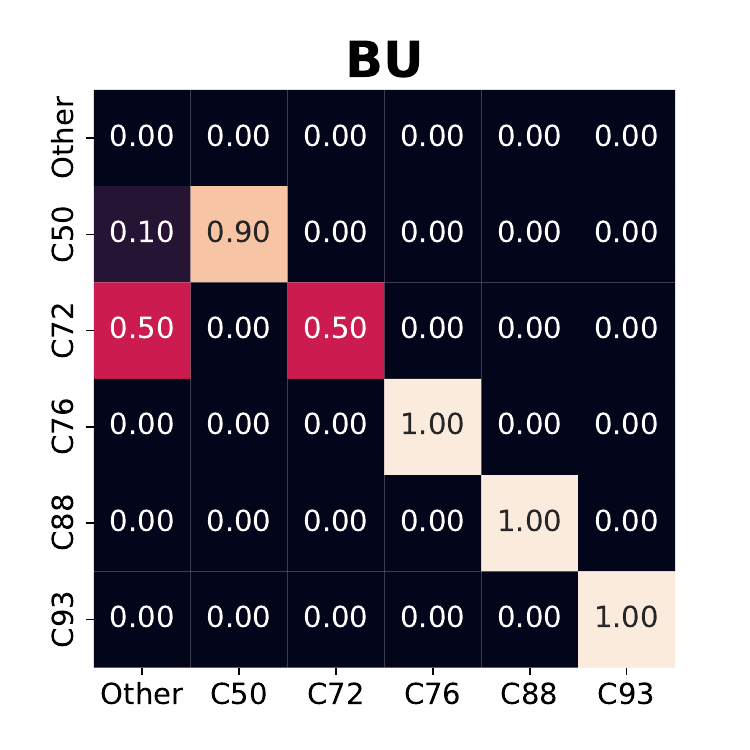}
		\includegraphics[width=0.24\linewidth]{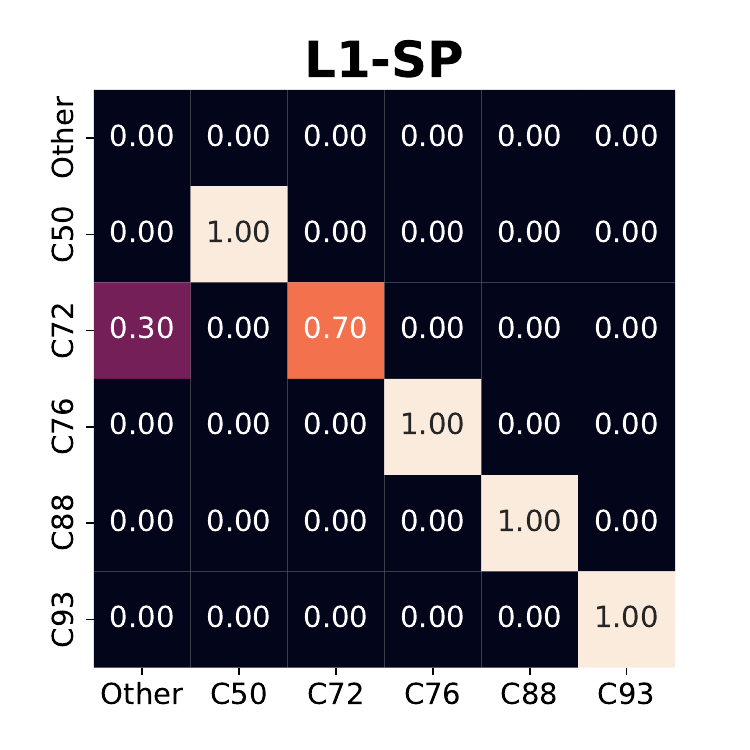}
		\includegraphics[width=0.24\linewidth]{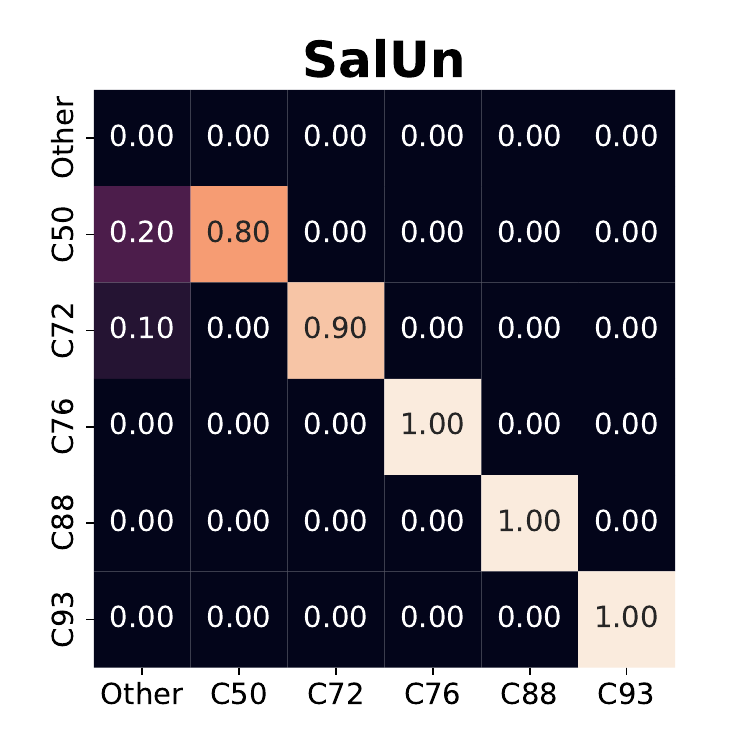}
		\caption{Confusion matrices of the \textbf{DST+STU+TCH} w.r.t. different MU methods}
	\end{minipage}
	\label{fig:open_source_pet37}
\end{figure*}

\end{document}